\documentclass[letterpaper, 10pt, conference]{ieeeconf}  

\IEEEoverridecommandlockouts                              

\usepackage{xr}

\newcommand{\omitted}[1]{}

\renewcommand{\baselinestretch}{0.96}
\title{\LARGE \bf
Outlier-Robust Spatial Perception: \\ Hardness, General-Purpose Algorithms, and Guarantees
}

\author{Vasileios Tzoumas,$^*$ Pasquale Antonante,$^*$ Luca Carlone 
\thanks{$^*$The authors contributed equally to this work.}
\thanks{V. Tzoumas, P. Antonante, L. Carlone are with the Laboratory for Information and Decision Systems, Massachusetts Institute of Technology, Cambridge, MA 02139, USA. \hfill \mbox{ }	{\tt\footnotesize \{vtzoumas,antonap,lcarlone\}@mit.edu}}%
\thanks{This work was partially funded by ARL DCIST CRA W911NF-17-2-0181, ONR RAIDER N00014-18-1-2828, Lincoln Laboratory ``Resilient Perception in Degraded Environments'' program, and the Google Daydream Research Program.}
}

\usepackage{comment}
\usepackage{siunitx}
\usepackage{relsize}
\usepackage{ifthen}
\usepackage[colorinlistoftodos]{todonotes}

\usepackage[caption=false]{subfig}

\usepackage[vlined,ruled,linesnumbered]{algorithm2e}
\usepackage{graphics} 
\usepackage{rotating}
\usepackage{color}
\usepackage{enumerate}
\usepackage[T1]{fontenc}
\usepackage{psfrag}
\usepackage{epsfig} 
\usepackage{booktabs}
\usepackage{url}
\usepackage{graphicx}
\usepackage{multirow}
\usepackage{array}
\usepackage{latexsym}
\usepackage{amsfonts}
\usepackage{amsmath}
\usepackage{amssymb}
\usepackage{xstring}
\usepackage[noend]{algorithmic}
\usepackage{multirow}
\usepackage{xcolor}
\usepackage{prettyref}
\usepackage{flexisym}
\usepackage{bigdelim}
\usepackage{breqn} 
\usepackage{listings}
\let\labelindent\relax
\usepackage{enumitem}
\usepackage{xspace}
\usepackage{bm}
\usepackage{tikz}
\usetikzlibrary{matrix,calc}

\usepackage{mdwlist}

\let\itemize\undefined
\makecompactlist{itemize}{stditemize}

\newrefformat{prob}{Problem\,\ref{#1}}
\newrefformat{def}{Definition\,\ref{#1}}
\newrefformat{sec}{Section\,\ref{#1}}
\newrefformat{sub}{Section\,\ref{#1}}
\newrefformat{prop}{Proposition\,\ref{#1}}
\newrefformat{app}{Appendix\,\ref{#1}}
\newrefformat{alg}{Algorithm\,\ref{#1}}
\newrefformat{cor}{Corollary\,\ref{#1}}
\newrefformat{thm}{Theorem\,\ref{#1}}
\newrefformat{lem}{Lemma\,\ref{#1}}
\newrefformat{fig}{Fig.\,\ref{#1}}
\newrefformat{tab}{Table\,\ref{#1}}

\newtheorem{theorem}{Theorem}
\newtheorem{problem}{Problem}

\newtheorem{lemma}[theorem]{Lemma}
\newtheorem{assumption}[theorem]{Assumption}
\newtheorem{definition}[theorem]{Definition}

\newtheorem{remark}[theorem]{Remark}
\newtheorem{example}[theorem]{Example}

\newcommand{\bdmath}{\begin{dmath}}
\newcommand{\edmath}{\end{dmath}}
\newcommand{\beq}{\begin{equation}}
\newcommand{\eeq}{\end{equation}}
\newcommand{\bdm}{\begin{displaymath}}
\newcommand{\edm}{\end{displaymath}}
\newcommand{\bea}{\begin{eqnarray}}
\newcommand{\eea}{\end{eqnarray}}
\newcommand{\beal}{\beq \begin{array}{ll}}
\newcommand{\eeal}{\end{array} \eeq}
\newcommand{\beas}{\begin{eqnarray*}}
\newcommand{\eeas}{\end{eqnarray*}}
\newcommand{\ba}{\begin{array}}
\newcommand{\ea}{\end{array}}
\newcommand{\bit}{\begin{itemize}}
\newcommand{\eit}{\end{itemize}}
\newcommand{\ben}{\begin{enumerate}}
\newcommand{\een}{\end{enumerate}}

\newcommand{\calE}{{\cal E}}
\newcommand{\calF}{{\cal F}}

\newcommand{\calI}{{\cal I}}

\newcommand{\calM}{{\cal M}}

\newcommand{\calO}{{\cal O}}
\newcommand{\calP}{{\cal P}}

\newcommand{\etal}{\emph{et~al.}\xspace}
\newcommand{\setal}{~\emph{et~al.}\xspace}

\newcommand{\hide}[1]{}

\newcommand{\hiddenText}{{\color{gray} hidden text.}}
\newcommand{\hideWithText}[1]{\hiddenText}

\newcommand{\Real}[1]{ { {\mathbb R}^{#1} } }

\newcommand{\SE}[1]{\ensuremath{\mathrm{SE}(#1)}\xspace}
\newcommand{\SEthree}{\ensuremath{\mathrm{SE}(3)}\xspace}
\newcommand{\SOtwo}{\ensuremath{\mathrm{SO}(2)}\xspace}
\newcommand{\SOthree}{\ensuremath{\mathrm{SO}(3)}\xspace}

\newcommand{\scenario}[1]{{\smaller \sf#1}\xspace}

\newcommand{\blue}[1]{{\color{blue}#1}}

\newcommand{\linkToPdf}[1]{\href{#1}{\blue{(pdf)}}}
\newcommand{\linkToPpt}[1]{\href{#1}{\blue{(ppt)}}}
\newcommand{\linkToCode}[1]{\href{#1}{\blue{(code)}}}
\newcommand{\linkToWeb}[1]{\href{#1}{\blue{(web)}}}
\newcommand{\linkToVideo}[1]{\href{#1}{\blue{(video)}}}
\newcommand{\award}[1]{\xspace} 

\newcommand{\maxOutliersSLAM}{{50\%}\xspace}
\newcommand{\maxOutliersRegView}{{90\%}\xspace}

\newcommand{\maxNumber}{k}
\newcommand{\perfBudgMin}{\epsilon}

\newcommand{\grSet}{\calM}

\newcommand{\res}{r}
\newcommand{\selSet}{\calO}
\newcommand{\apost}{\chi}

\newcommand{\pone}{p}
\newcommand{\ptwo}{p'}
\newcommand{\Pone}{\calP}
\newcommand{\Ptwo}{\calP'}

\newcommand{\fone}{f}
\newcommand{\ftwo}{f'}
\newcommand{\Fone}{\calF}
\newcommand{\Ftwo}{\calF'}

\newcommand{\myParagraph}[1]{{\bf #1.}}

\newcommand{\pcm}{\scenario{PCM}}
\newcommand{\rrr}{\scenario{RRR}}
\newcommand{\dcs}{\scenario{DCS}}

\newcommand{\BnB}{BnB\xspace}
\newcommand{\SLAM}{SLAM\xspace}

\newcommand{\ransac}{\scenario{RANSAC}}
\newcommand{\FGR}{\scenario{FGR}}
\newcommand{\DCS}{\scenario{DCS}}

\newcommand{\name}{\scenario{ADAPT}}
\newcommand{\nameLong}{Adaptive Trimming\xspace}

\newcommand{\nameFast}{\name}

\newcommand{\MTS}{MTS\xspace}
\newcommand{\MTSlong}{Minimal Trimmed Squares\xspace}

\newcommand{\MATLAB}{\textsc{MATLAB}}

\PassOptionsToPackage{end}{algorithmic}

\begin{document}

\maketitle
\thispagestyle{empty}
\pagestyle{empty}

\begin{abstract}
Spatial perception is the backbone of many robotics applications, and spans a broad range of research problems, including localization and mapping, point cloud alignment, and relative pose estimation from camera images.  
Robust spatial perception is jeopardized by the presence of incorrect data association, and in general, outliers. Although techniques to handle outliers do exist, they can fail in unpredictable manners (e.g., \ransac, robust estimators), or can have exponential runtime (e.g., branch-and-bound).
In this paper, we advance the state of the art in outlier rejection by making three contributions. 
First, we show that even a simple linear instance of outlier rejection is \emph{inapproximable}: in the worst-case one cannot design a quasi-polynomial time algorithm that computes an approximate solution efficiently.
Our second contribution is to provide the first per-instance sub-optimality bounds to assess the approximation quality of a given outlier rejection outcome. 
Our third contribution is to propose a simple general-purpose algorithm, named \emph{adaptive trimming}, to remove outliers. Our algorithm leverages recently-proposed global solvers that are able to solve outlier-free problems, and iteratively removes measurements with large errors. 
We demonstrate the proposed algorithm on three spatial perception problems: 3D registration, two-view geometry, and \SLAM. The results show that our algorithm outperforms several
 state-of-the-art methods across applications while being a general-purpose method.
\omitted{Also, it can tolerate up to \maxOutliersRegView outliers in 3D registration and two-view geometry, and up to \maxOutliersSLAM outliers in SLAM. 
Finally, the results show that the proposed sub-optimality bound is empirically tight, and effective in assessing the outlier rejection outcomes.}
\end{abstract}
\section{Introduction}
\label{sec:intro}

\emph{Spatial perception}
is concerned with the estimation of a geometric model that describes the state  of the robot, and/or 
the environment the robot is deployed in. 
As such, spatial perception includes a broad set of robotics problems, 
including motion estimation~\cite{Scaramuzza11ram}, object detection, localization and tracking~\cite{Tam13tvcg-registrationSurvey}, 
multi-robot localization~\cite{Roumeliotis02tra}, dense reconstruction~\cite{Whelan15rss}, and Simultaneous Localization and Mapping (\SLAM)~\cite{Cadena16tro-SLAMsurvey}. 
Spatial perception algorithms find applications beyond robotics, including virtual and augmented reality, and medical imaging~\cite{Tam13tvcg-registrationSurvey}, to mention a few.  

\newcommand{\myhspace}{\hspace{-3mm}}

\newcommand{\mpw}{4.5cm}
\begin{figure}[t]
	\begin{center}
	\begin{minipage}{\textwidth}
	\hspace{-0.2cm}
	\begin{tabular}{cc}%
		\myhspace
			\begin{minipage}{\mpw}%
			\centering%
			\includegraphics[width=0.65\columnwidth]{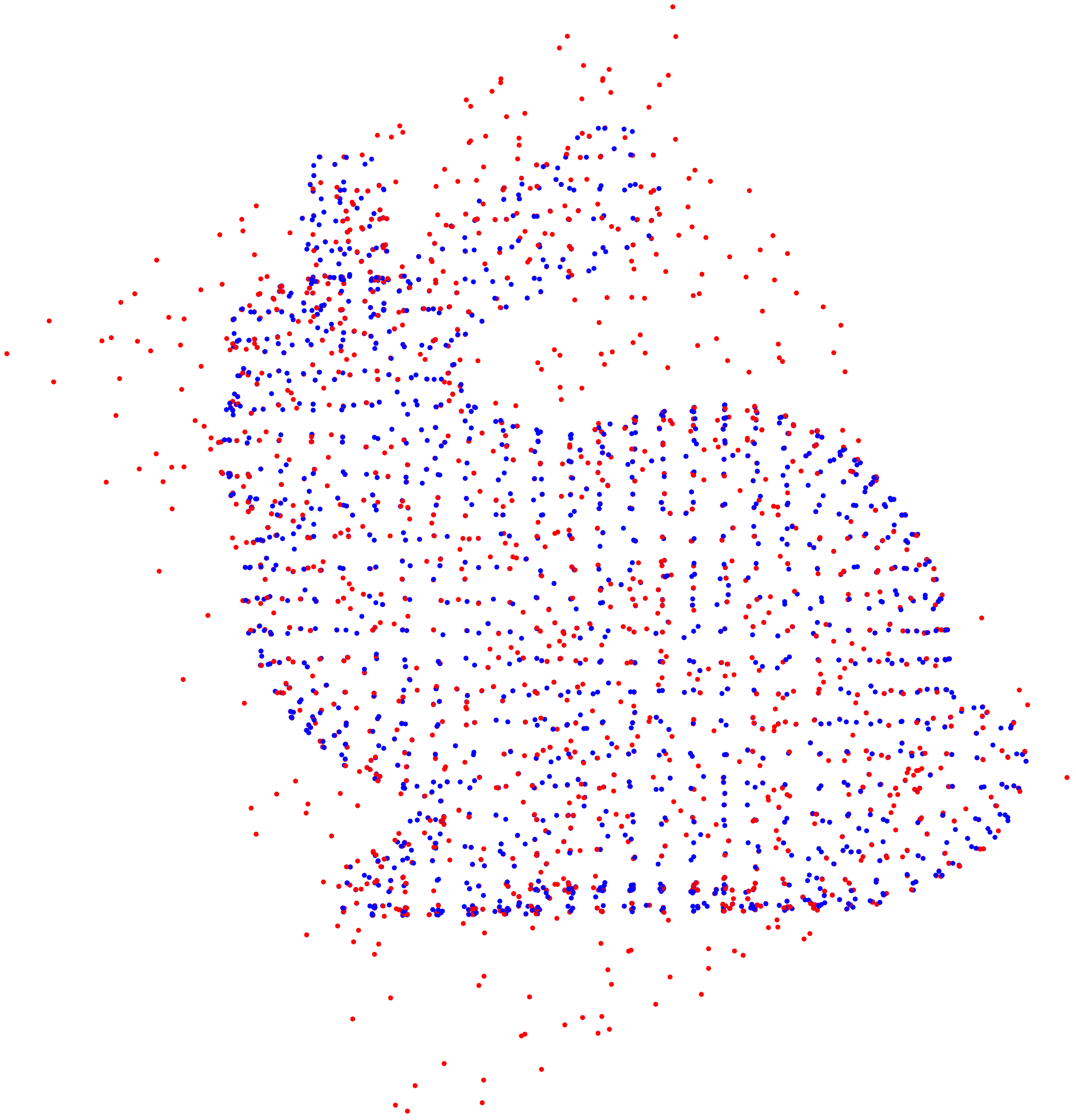} \\
			(a) 3D registration 
			\end{minipage}
		& \myhspace
			\begin{minipage}{\mpw}%
			\centering%
      \includegraphics[width=0.675\columnwidth]{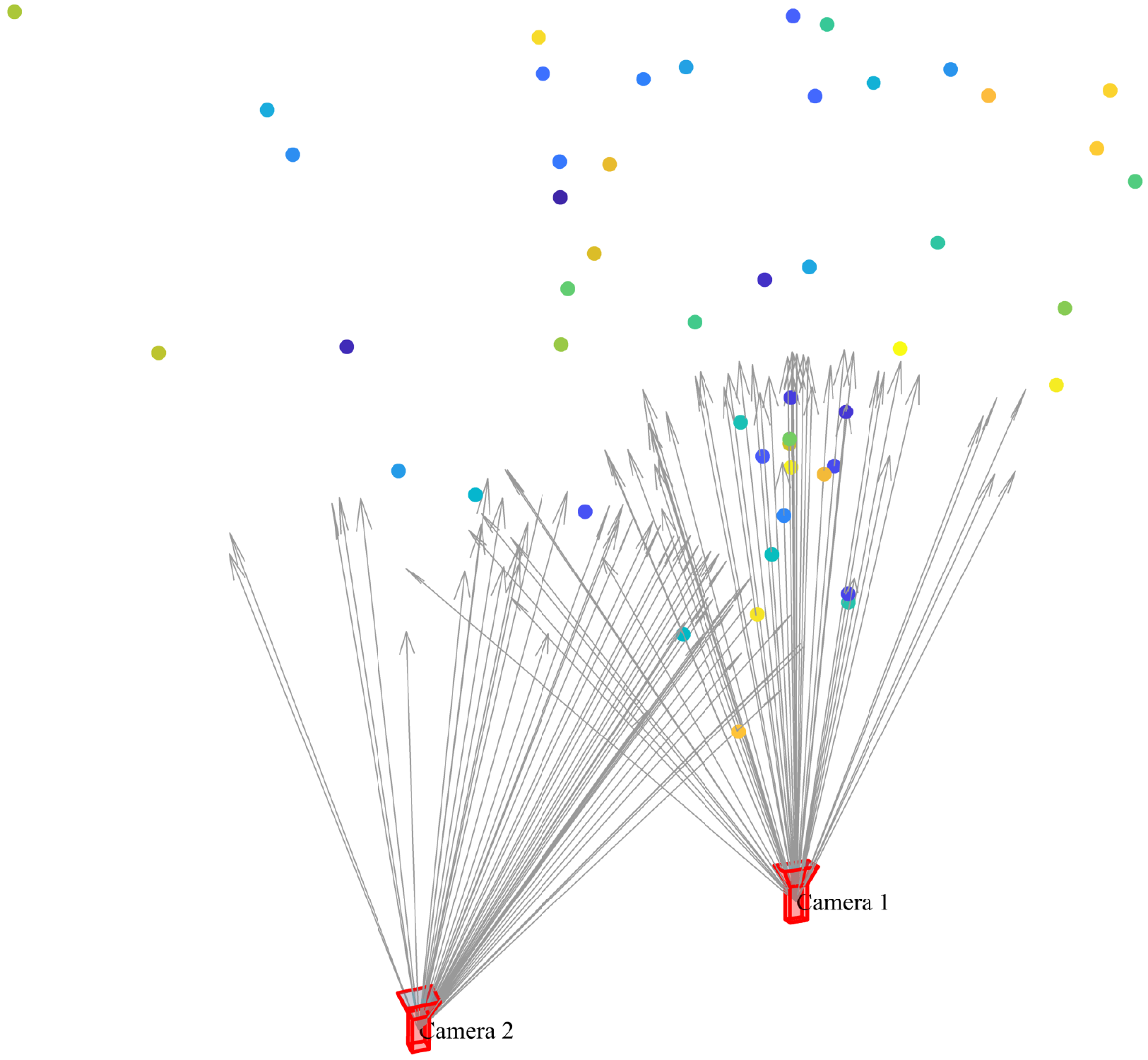} \\
      \vspace{2mm}
			(b) Two-view geometry
			\end{minipage}
		\\ \myhspace
      \begin{minipage}{\mpw}%
      \vspace{2mm}
			\centering%
			\includegraphics[width=0.7\columnwidth]{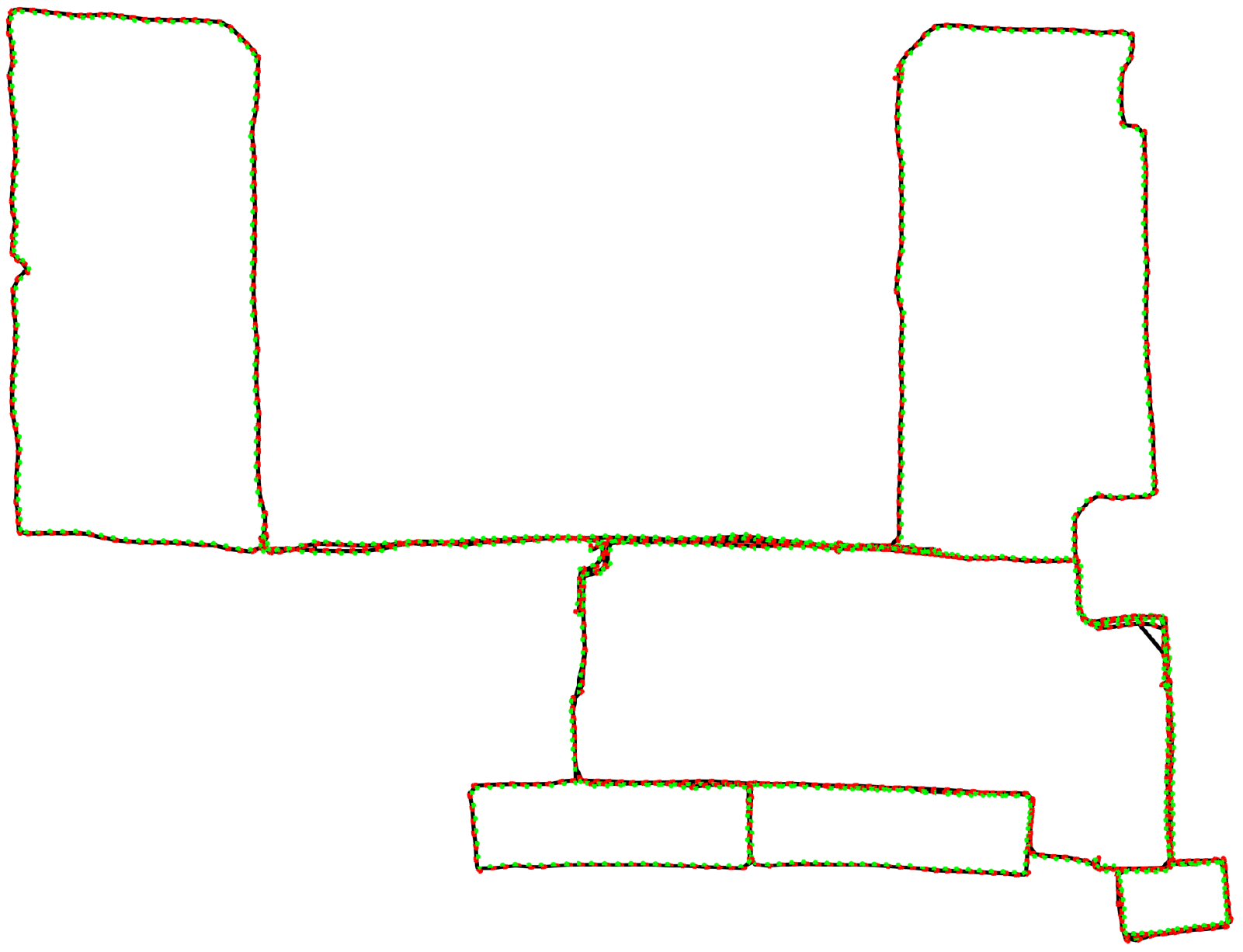} \\
			(c) 2D SLAM
			\end{minipage}
		& \myhspace
      \begin{minipage}{\mpw}%
      \vspace{2mm}
			\centering%
			\includegraphics[width=0.533\columnwidth]{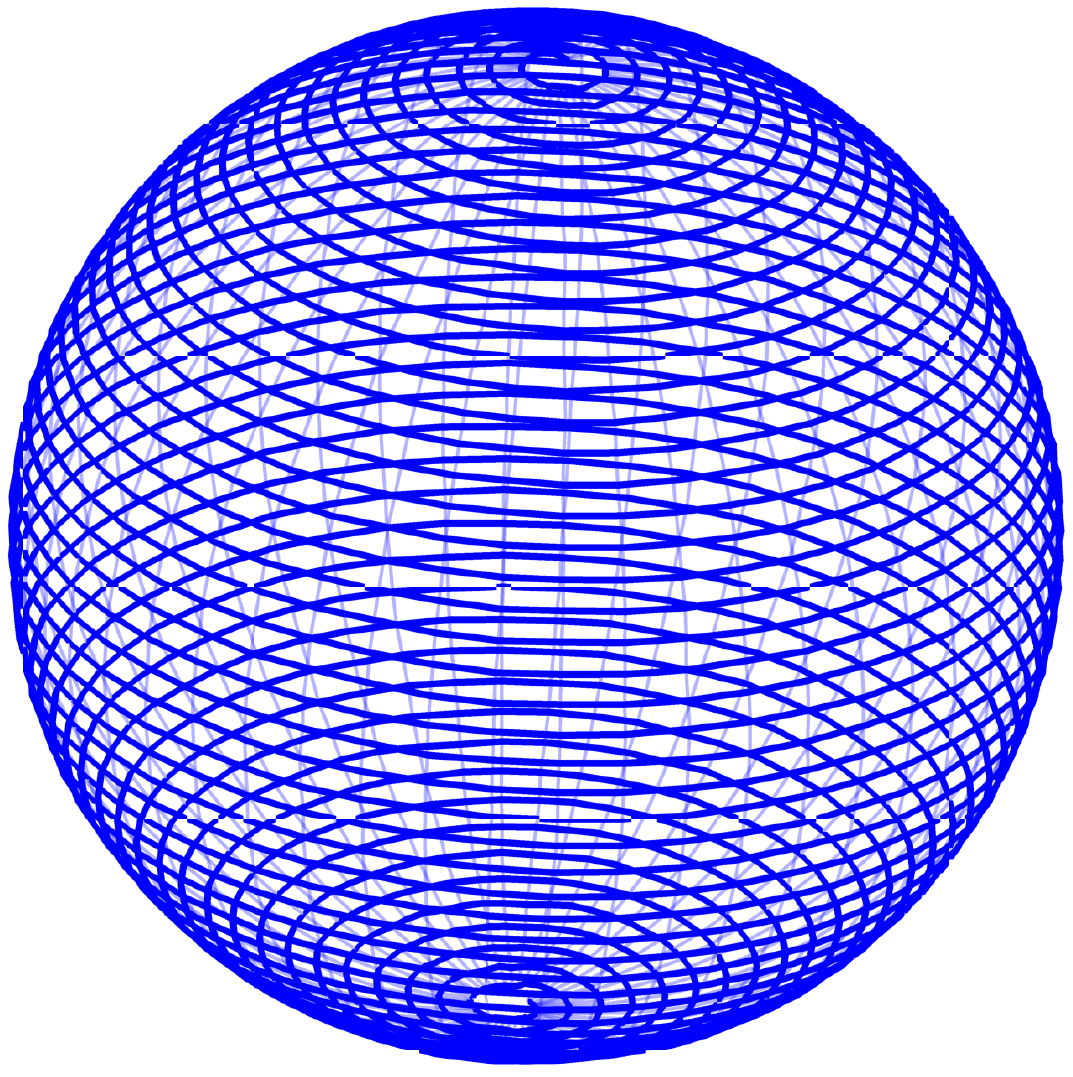} \\
			(d) 3D SLAM
			\end{minipage}
		\end{tabular}
	\end{minipage}
	\begin{minipage}{\textwidth}
	\end{minipage}
	\vspace{-3mm}
	\caption{
We investigate outlier rejection across multiple spatial perception problems, including 
(a) 3D registration, (b) two-view geometry, and (c-d) SLAM. 
We provide inapproximability results and performance bounds. 
We also propose an algorithm, \name, that outperforms \ransac and other specialized methods.
\name tolerates up to \maxOutliersRegView outliers in 3D registration, and up to 
\maxOutliersSLAM outliers in two-view geometry and most SLAM datasets. 
	 \label{fig:overview}}
	\vspace{-8mm} 
	\end{center}
\end{figure}

Safety-critical applications, including self-driving cars, demand robust spatial perception algorithms that can 
estimate correct models (and assess their performance) in the presence of measurement noise and outliers. 
While we currently have several approaches that can tolerate large measurement noise 
(e.g.,~\cite{Rosen18ijrr-sesync,Briales18cvpr-global2view,Horn87josa}), these algorithm tend to catastrophically fail in the presence of outliers resulting from incorrect data association, sensor malfunction, or even adversarial attacks.

In this paper, we focus on the analysis and design of \emph{outlier-robust} \emph{general-purpose} algorithms for 
robust estimation applied to spatial perception. 
Our proposal is motivated by three observations. 
First, recent years have seen a convergence of the robotics community towards optimization-based approaches for spatial perception.
Therefore, despite the apparent heterogeneity of the perception landscape, it is possible to develop \emph{general-purpose} methods to 
reject outliers (e.g., M-estimators~\cite{Bosse17fnt} and \emph{consensus maximization}~\cite{Speciale17cvpr-consensusMaximization} can be thought as general estimation tools).
Second, the research community has developed global solutions to many perception problems \emph{without} outliers, 
from well-established techniques for point cloud registration~\cite{Horn87josa}, to very recent solvers for SLAM~\cite{Rosen18ijrr-sesync} and two-view geometry~\cite{Briales18cvpr-global2view}.
These global solvers offer unprecedented opportunities to tackle robust estimation \emph{with} outliers. 
Third, the literature still lacks a {satisfactory answer to provably-robust spatial perception.}

The literature on outlier-robust spatial perception is currently divided between \emph{fast} approaches (that mainly work in the low-outlier regime, without performance guarantees) and \emph{provably-robust} approaches (that can tolerate many outliers, but have exponential run-time). While we postpone 
a comprehensive  literature review to Section~\ref{sec:relatedWork}, 
it is instructive to briefly review this dichotomy.
\emph{Fast} approaches include \ransac~\cite{Chen99pami-ransac}, M-estimators~\cite{Bosse17fnt}, and 
measurement-consistency checking~\cite{Latif12rss,Mangelson18icra}. These methods fall short of providing performance guarantees.  In particular, \ransac is known to become slow and brittle with high outlier rates ($>50\%$)~\cite{Speciale17cvpr-consensusMaximization}, and 
does not scale to high-dimensional problems, while M-estimators have a breakdown point of zero, meaning that a single ``bad'' outlier can compromise the results. 
On the other hand, provably-robust methods, typically based on \emph{branch-and-bound}~\cite{Bazin14eccv-robustRelRot,hartley2009ijcv-global,Zheng11cvpr-robustFitting,Li09cvpr-robustFitting,Speciale17cvpr-consensusMaximization}, can tolerate more than $50\%$ of outliers~\cite{Bustos18pami-GORE}, but do not scale to large problems and are relatively slow for robotics applications. 
Overall, the first goal of this paper is to understand whether we can resolve this divide, and design algorithms that are both efficient and provably robust.
\omitted{The second and final goal of this paper is to understand whether we can resolve another divide in outlier-robust optimization: that of \emph{worst-case-}instance inapproximability versus \emph{per-}instance approximability.  In particular, in this paper we provide worst-case instances of outlier rejection that are impossible to be approximated within any approximation factor.  However, despite the theoretical importance of the result, a question of crucial practical importance remains open:  Given any other instance of an outlier-rejection problem, and an algorithm for its solution, how well have we approximated the optimal estimation error after running the algorithm? By answering this question, one enables an \emph{a posteriori (post-run)} performance assessment of the used algorithm, and the detection of any potential algorithmic failures.  Thus, one enables fail-safe procedures.  We give the first answer to the previous question as described in our contributions below.  And we postpone the relevant literature review to the Appendix.  Here we only note that past literature relies on \emph{a-priori} sub-optimality bounds that are NP-hard to compute, or are even unverifiable, since they rely on assumptions such as the magnitute of the outliers~\cite{Candes05tit,Flores2015test-sharp,Liu2018tsp-GreedyRobustRegression,Das11icml-submodularity, sviridenko2017optimal,bian2017guarantees}.}  

\myParagraph{Contributions}  We propose a  \textit{\MTSlong (\MTS)} formulation for 
outlier-robust estimation.  \MTS encapsulate a wide spectrum of commonly-used outlier-robust formulations in the literature, such as the popular \emph{maximum consensus}~\cite{Chin18eccv-robustFitting}, \emph{Linear Trimmed Squares}~\cite{Rousseeuw87book}, and \emph{truncated least-squares}~\cite{Huber64ams}.
In particular, \MTS  aims to compute a ``good'' estimate by rejecting a minimal set of measurements.

Our first contribution (Section~\ref{sec:hardness}) is a negative result:
we show that outlier rejection is \emph{inapproximable}. In the worst-case, there exist no quasi-polynomial algorithm that can compute (even an approximate) solution to the outlier rejection problem.
We prove that this remains true, surprisingly, even if the algorithm knows the true number of outliers 
and even if we allow the algorithm to reject more measurements than necessary.
Our conclusions largely extend previously-known negative results~\cite{Chin18eccv-robustFitting}, which already 
ruled-out the possibility of designing polynomial-time approximation methods.

Our second contribution (Section~\ref{sec:guarantees}) is to derive the first per-instance sub-optimality bounds to assess the  
 quality of a given outlier rejection solution. 
 While in the worst case we expect efficient algorithms to perform poorly, we can still hope that 
 in typical problem instances a polynomial-time algorithm can compute good solutions, and we can use the proposed sub-optimality bounds to assess the performance of such an algorithm. 
 Our bounds are algorithm-agnostic (e.g., they also apply to \ransac) and can be computed efficiently.

Our third contribution (Section~\ref{sec:algorithms}) is a \emph{general-purpose} algorithm for outlier rejection, named \emph{\nameLong ({\name})}.  \name leverages recently-proposed global solvers that solve outlier-free problems and adaptively removes measurements with large residual errors.    
Despite its simplicity, our experiments show that it outperforms \ransac and even specialized state-of-the-art methods
 for robust estimation.

We conclude the paper by providing an experimental evaluation across multiple spatial perception problems (Section~\ref{sec:experiments}).
The experiments show that 
\name can tolerate up to \maxOutliersRegView outliers in 3D registration (with a run-time similar to existing methods), and up to \maxOutliersSLAM outliers in two-view geometry and most SLAM datasets. 
The experiments also show that the proposed sub-optimality bounds are effective in assessing the outlier rejection outcomes.  We report extra results and proofs in the Appendix.

\section{Outlier Rejection: a Minimally Trimmed Squares Formulation}
\label{sec:problem_formulation}

Many estimation problems in robotics and computer vision can be formulated as non-linear least squares problems:
\begin{equation}
\label{eq:nls}
\min_{x\in \mathbb{X}} \;\; \sum_{i \in \grSet} \|h_i(y_i,x)\|^2, 
\end{equation}
where we are given measurements $y_i$ of an unknown variable $x$, with $i \in \grSet$ ($\grSet$ is the measurement set), 
and we want to estimate $x$, potentially restricted to a given domain $\mathbb{X}$
(e.g., $x$ is a pose, and $\mathbb{X}$ is the set of 3D poses). The least squares problem in eq.~\eqref{eq:nls} looks for the $x$ 
that minimizes the (squares of) the \emph{residual errors} $h_i(y_i,x)$, where the $i$-th residual error captures how well 
$x$ explains the measurement~$y_i$. The problem in eq.~\eqref{eq:nls} typically results from maximum likelihood and maximum a posteriori estimation~\cite{Cadena16tro-SLAMsurvey,Dellaert17fnt-factorGraph}, 
under the assumption that the measurement noise is Gaussian.

Both researchers and practitioners are well-aware that least squares formulations are sensitive to outliers, and that the estimator in eq.~\eqref{eq:nls} fails to produce a meaningful estimate of $x$ in the presence of gross outliers $y_i$. 
Therefore, in this paper we address the following question:
\begin{center}
\textit{Can we compute an accurate estimate of $x$ that is insensitive to the presence 
of outlying measurements?}
\end{center}

We formulate the resulting robust estimation problem as the problem of selecting a small 
number of outliers, such that the remaining measurements (the inliers) can be explained with small error. In other words, a good estimate (in the presence of outliers) is one that explains as many measurements as possible while disregarding outliers. This intuition leads to the following formulation.

\begin{problem}[Minimally \hspace{-0.1mm}Trimmed \hspace{-0.1mm}Squares \hspace{-0.1mm}(\MTS)] 
\label{pr:dual}
Let $\grSet$ denote a set of measurements of an unknown variable~$x$, and let $y_i$ denote the $i$-th measurement. Also denote with $h_i(y_i,x)$ the residual error that quantifies how well $x$ fits the measurement $y_i$.
Then, the \emph{minimally trimmed squares} problem consists in estimating 
the unknown variable $x$ by solving the following optimization problem:
\begin{equation}
\label{eq:MTS}
\min_{\selSet\subseteq \grSet}
\;\; 
\min_{x\in \mathbb{X}}  \;\;\;|\selSet|,\;\;\text{s.t.}\;\;\;\sum_{i \in \grSet\setminus\selSet} \|h_i(y_i,x)\|^2 \leq \perfBudgMin_{\grSet\setminus \selSet},
\end{equation}
where one searches for the smallest set of outliers $\selSet$ ($|\hspace{.8mm}\cdot\hspace{.8mm}|$ is the cardinality of a set) among the given measurements $\grSet$, such that the remaining measurements $\grSet\setminus \selSet$ (i.e., the inliers) can be explained with small error, i.e., $\sum_{i \in \grSet\setminus\selSet} \|h_i(y_i,x)\|^2 \leq \perfBudgMin_{\grSet\setminus \selSet}$ for 
some $x \in \mathbb{X}$, and where $\perfBudgMin_{\grSet\setminus \selSet}$ is a given \emph{outlier-free} bound.
\hfill $\lrcorner$
\end{problem}

\begin{example}[Robust linear estimation and bound $\perfBudgMin_{\grSet\setminus \selSet}$]\label{ex:linear_dec} 
In linear estimation one wishes to recover a parameter $x \in \mathbb{R}^n$ from a set of noisy measurements $y_i=a_i^\top x+d_i$, $i\in \grSet$, where $a_i$ is a known vector, and 
 $d_i \in \Real{}$ models the unknown measurement noise.
 Some of the measurements (the inliers) are such that the corresponding noise $d_i$ can be assumed to follow a Gaussian distribution, while others (the outliers) may be affected by large noise. 
 Therefore, our \MTS estimator can be written as:  
\begin{equation}\label{eq:dual_linear_regr}
\min_{x\in \mathbb{R}^n} \;\; \min_{\selSet\subseteq \grSet} \;\;\;|\selSet|,\;\;\text{s.t.}\;\;\;\sum_{i \in \grSet\setminus\selSet} \|y_i-a_i^\top x\|^2\;\leq \perfBudgMin_{\grSet\setminus \selSet}.
\end{equation}
Evidently, $\perfBudgMin_{\grSet\setminus \selSet}$ 
must increase with the number of inliers, since each inlier adds a positive summand $\|y_i-a_i^\top x\|^2$ due to the presence of noise. 
Moreover, since the sum is restricted to the inliers, for which the noise is assumed to be Gaussian, we can compute the desired outlier-free bound explicitly: 
if 
$d_i$ follows a Gaussian distribution, then each $\|y_i-a_i^\top x\|^2_{2}$ follows a $\chi^2$ distribution with $1$ degree of freedom. 
 Thus, with desired probability $p_\epsilon$ (e.g., $0.99$),  $\|y_i-a_i^\top x\|^2_{2}\leq \perfBudgMin$ where $\perfBudgMin$ is the $p_\epsilon$-quantile of the $\chi^2$ distribution, 
 and  the outlier-free bound is $\perfBudgMin_{\grSet\setminus \selSet}=|\grSet\setminus \selSet| \perfBudgMin$.
\hfill $\lrcorner$
\end{example}

\begin{remark}[Generality \!and \!applicability]
In this paper we address robustness in non-linear and non-convex estimation problems 
as the ones arising in robotics and computer vision. 
Therefore, while the linear estimation Example~\ref{ex:linear_dec} is instructive 
 (and indeed we will prove in Section~\ref{sec:hardness} that even in such a simple case, it is not possible to even approximate the \MTS 
estimator in polynomial time), the algorithms and bounds presented in this paper hold for any 
function $h_i(y_i,x)$ and  any domain $\mathbb{X}$. 
In contrast with related work~\cite{Rousseeuw11dmkd, Yang16pami-goicp}, we do not assume the number of outliers to be known in advance (an unrealistic assumption in perception problems). 
Indeed, our \MTS formulation looks for the smallest set of outliers. 
Finally, while the formulation~\eqref{eq:MTS} requires to set an outlier-free threshold, we will propose an algorithm (Section~\ref{sec:algorithms}) that will automatically compute a suitable threshold without any prior knowledge about the measurement noise. 
\hfill $\lrcorner$
\end{remark}

In summary, \MTS is a general non-linear and non-convex outlier rejection framework.
We exemplify its generality by discussing its application to three core perception problems:
 3D registration, two-view geometry, and SLAM.
\subsection{Outlier rejection for robust spatial perception: \\   3D registration, two-view geometry, and SLAM} 
\label{sec:examples}

Here we review three core problems in spatial perception, 
and show how to tailor the framework of Section~\ref{sec:problem_formulation} to these examples. 
The expert reader can safely skip this section.

\myParagraph{Outlier rejection for 3D registration}\label{subsec:registration}
Point cloud registration 
consists in finding the rigid transformation that aligns two point clouds. 
Formally, 
we are given two sets of points 
$\Pone \doteq \{\pone_1,\ldots, \pone_n\}$
and $\Ptwo \doteq \{\ptwo_1,\ldots, \ptwo_n\}$ (with $\pone_i,\ptwo_i \in \Real{3}$, for $i=1,\ldots,n$),
as well as a set $\grSet$ of \emph{putative correspondences} 
$(i,j)$, such that the point $\pone_i\in\Pone$ and the point $\ptwo_j \in \Ptwo$ are (putatively) related by
a rigid transformation, for all $(i,j) \in \grSet$. 
Point correspondences are typically obtained by descriptor matching~\cite{Bustos18pami-GORE}.
 
Given the points and the putative correspondences, 
3D registration looks for a rotation $R \in SO(3)$, and a translation $t\in\mathbb{R}^3$, 
that align (i.e., minimize the sum of the squared distances between) corresponding points:
\begin{equation}\label{eq:3D-reg}
\min_{ \substack{R \in SO(3) \\ \;t\in\mathbb{R}^3} }\;\sum_{(i,j)\in \grSet} \| R\,\pone_i + t - \ptwo_j\|^2.
\end{equation} 
The problem in eq.~\eqref{eq:3D-reg} can be solved in closed form~\cite{Horn87josa}.
However, eq.~\eqref{eq:3D-reg} fails to produce a reasonable pose (rotation and translation) estimate when some of the correspondences are outliers~\cite{Bustos18pami-GORE,Zhou16eccv-fastGlobalRegistration}, and related work resorts to robust estimators {(reviewed in Section~\ref{sec:relatedWork}). 
Here we rephrase robust registration as an \MTS problem:
\begin{equation}\label{eq:rob_3D-reg}
\min_{ \substack{R \in SO(3) \\ \;t\in\mathbb{R}^3}  } \;\min_{\selSet\subseteq \grSet} \;\;\;|\selSet|,\;\text{s.t.}\!\!\!
\sum_{(i,j)\in \grSet\setminus \selSet} \hspace{-3mm} \|R\,\pone_i+t-\ptwo_j\|^2\leq \perfBudgMin_{\grSet\setminus \selSet}.
\end{equation}
\omitted{that looks for the smallest set of outliers $\selSet$ within the given putative correspondences $\grSet$, such that the remaining correspondences (i.e., the inliers $\grSet\setminus \selSet$) can be aligned with a cumulative error smaller than a given threshold $\perfBudgMin_{\grSet\setminus \selSet}$.}

\myParagraph{Outlier rejection for two-view geometry}\label{subsec:geometry}
Two-view geometry estimation consists in finding the relative pose (up to scale) 
 between two camera images picturing a static scene, and it is crucial for motion estimation~\cite{Nister04pami}, object localization~\cite{Nister04pami}, and reconstruction~\cite[Chapter~1]{Hartley04book}.
We consider a feature-based calibrated setup where 
the camera calibration is known and 
one extracts features (keypoints)
 $\Fone = \{\fone_1,\ldots,\fone_n\}$ and
 $\Ftwo = \{\ftwo_1,\ldots,\ftwo_n\}$ 
 from the first and second image, respectively.
We are also given a set of putative correspondences $\grSet$ between pairs of features 
$(i,j)$, such that features $\fone_i$ and $\ftwo_j$ (putatively) picture the same 3D point observed in both images.

Given the features and the putative correspondences, two-view geometry
looks for the rotation $R \in SO(3)$ and the translation $t \in \Real{3}$ (up to scale)
 that minimizes the violation of the epipolar constraint:
\begin{equation}\label{eq:2v-geo}
\min_{ \substack{R \in SO(3) \\ t\in\mathbb{S}^2} }\;\;\;\sum_{(i,j)\in \grSet}\left[f_i^\top \left(t\times (Rf_j')\right)\right]^2,
\end{equation} 
where $t$ is restricted to the unit sphere $\mathbb{S}^2$ to remove the scale ambiguity.
In the absence of outliers, problem~\eqref{eq:2v-geo} can be solved globally using convex relaxations~\cite{Briales18cvpr-global2view}.

In the presence of outliers, the non-robust formulation in eq.~\eqref{eq:2v-geo} fails to compute accurate pose estimates, hence we rephrase two-view estimation as an \MTS problem:

\begin{equation}\label{eq:rob_2v-geo}
\min_{ \substack{R \in SO(3) \\ t\in\mathbb{S}^2} } \;\min_{\selSet\subseteq \grSet} \;\;|\selSet|,\;\text{s.t.} 
\;\sum_{(i,j)\in \grSet} \!\!\!\left[t^\top \left(\fone_i\times (R\ftwo_j)\right)\right]^2\leq \perfBudgMin_{\grSet\setminus \selSet}.
\end{equation}


\myParagraph{Outlier rejection for SLAM}\label{subsec:pgo}
Here we consider one of the most popular SLAM formulations: \emph{Pose graph optimization} (PGO).
PGO estimates a set of robot poses $T_i \in \SEthree$ ($i=1,\ldots,n$)
from pairwise relative pose measurements $\bar{T}_{ij} \in \SEthree$ between pairs of poses $(i,j) \in \grSet$. The measurement set $\grSet$ includes odometry (ego-motion) measurements as well as loop closures. In the absence of outliers, one can compute the pose estimates as:
\begin{equation}
\min_{ \substack{T_i\in\SE{3} \\ i=1,\ldots,n} } 
\sum_{(i,j) \in \grSet} \| T_j - T_i \bar{T}_{ij} \|_F^2,
\end{equation}
where $\|\hspace{.8mm}\cdot\hspace{.8mm}\|_F^2$ denotes the Frobenius norm.  

In practice, many loop closure measurements  are outliers (e.g., due to failures in place recognition). Therefore, we rephrase PGO as an MTS problem over the loop closures:
\begin{align}
&\min_{ \substack{T_i\in\SE{3} \\ i=1,\ldots,n} } \;\; \min_{\selSet \subseteq \calE_{lc}}
\;\;|\selSet|,\;\;\text{s.t.}\;\;\nonumber\\
&\!\sum_{(i,j) \in \calE_o} \| T_j - T_i \bar{T}_{ij} \|_F^2+\hspace{-3mm}\sum_{(i,j) \in \calE_{lc}\setminus \selSet} \hspace{-3mm}\| T_{j} - T_{i} \bar{T}_{ij} \|_F^2 \leq \perfBudgMin_{\grSet\setminus \selSet},
\end{align}
where we split the measurement set $\grSet$ into odometric edges $\calE_o$ (these can be typically trusted), and loop closures $\calE_{lc}$ (typically containing outliers).

\section{Outlier Rejection is Inapproximable} 
\label{sec:hardness}

We show that \MTS is \emph{inapproximable} even by quasi-polynomial-time algorithms.  To this end, we find worst-case instances for which there is no algorithm that can reject a few measurements to achieve a prescribed residual error $\epsilon$ (subject to a widely believed conjecture in complexity theory, similar to $NP\neq P$).  We start with some definitions and present our key result in Theorem~\ref{th:hardness}. 

\begin{definition}[Approximability]\label{def:approx}
	Consider the \MTS Problem~\ref{pr:dual}.
	Let $\selSet^\star$ be an optimal solution, let $\maxNumber^\star\doteq|\selSet^\star|$, 
	and $\epsilon\doteq \epsilon_{\grSet\setminus \selSet^\star}$, that is, $\epsilon$ is the outlier-free bound when the measurements $\selSet^\star$ are the rejected outliers.  Also, consider a number $\lambda>1$.
	We say that an algorithm makes \MTS  \emph{$(\lambda, \epsilon)$-approximable} if it returns a set $\selSet$, 
	and a parameter $x$,
	such that
	 $|\selSet|\leq\lambda\maxNumber^\star$ and
	 $\sum_{i \in\grSet\setminus \selSet}\|h_i(y_i,x)\|^2\leq \epsilon$. 
	\hfill $\lrcorner$
\end{definition}

The definition of $(\lambda, \epsilon)$-approximability allows some slack in the quality of the \MTS's solution: rather than solving Problem~\eqref{eq:MTS} exactly ($\lambda = 1$), 
Definition~\ref{def:approx} only requires, for \MTS to be approximable, to find an algorithm that computes an estimate  \emph{close} to the optimal solution. Indeed, Definition~\ref{def:approx} includes algorithms that can 
reject more outliers than necessary (since $\lambda\maxNumber^\star>\maxNumber^\star$).

\begin{definition}[Quasi-polynomial \hspace{-0.6mm}algorithm]\label{def:quasi}
An algorithm is said to be \emph{quasi-polynomial} if it runs in $2^{O\left[( \log m)^c \right]}$ time, where 
$m$ is the size of the input and $c$ is constant.\hfill $\lrcorner$
\end{definition}

Any polynomial algorithm is also quasi-polynomial, since $m^k = 2^{k \log m}$. Yet, a quasi-polynomial algorithm is asymptotically faster than an exponential-time algorithm, since exponential algorithms run in $O(2^{m^c})$ time, for some $c>0$.

\begin{theorem}[Inapproximability]\label{th:hardness}
	Consider the linear \MTS problem~\eqref{eq:dual_linear_regr}. Let $x^\star$ be the optimal value of the variable to be estimated, $m$ be the number of measurements ($m\doteq |\grSet|$), $\selSet^\star$ be the optimal solution, and set $\maxNumber^\star\!\doteq\!|\selSet^\star|$. 
	Then, for any $\delta\!\in\!(0,1)$, there exist a polynomial $\lambda_1(m)$ and a  $\lambda_2(m)=2^{\Omega(\log^{1-\delta} m )}$ and instances of \MTS (i.e., measurements $y_i$, 
	vectors $a_i$,  and outlier-free bound $\epsilon$) where $\epsilon=\lambda_2(m)$, such that unless ${\rm NP} {\in}{\rm BPTIME(\textit{m}^{\rm poly \log \textit{m}})}$,\footnote{The complexity hypothesis ${\rm NP} {\notin}{\rm BPTIME(\textit{m}^{\rm poly \log \textit{m}})}$ means there is no randomized algorithm which outputs solutions to problems in ${\rm NP}$ with probability $2/3$, after running for $O(m^{(\log m)^c})$ time, for a constant $c$~\cite{Arora09book-complexity}.} there is no quasi-polynomial algorithm making \MTS  $(\lambda_1(m), \lambda_2(m))$-approximable.  This holds true {even if the algorithm knows $\maxNumber^\star$\!, and that $x^\star$ \!exist.~\hfill $\lrcorner$}
\end{theorem}

Theorem~\ref{th:hardness} stresses the extreme hardness of \MTS.  Even if we inform the algorithms with the true number of outliers, it is impossible in the worst-case for even quasi-polynomial algorithms to find a good set of inliers. 
Surprisingly, this remains true even if we allow the algorithms to cheat by rejecting more measurements than $\maxNumber^\star$ (i.e., $\lambda_1 \maxNumber^\star$).


Thinking beyond the worst-case, our inapproximability result suggests that to obtain a good solution efficiently, our only hope is that nature (which picks the outliers) is not adversarial, thus fast algorithms can compute good solutions in practice. 
Hence, it becomes important to derive \emph{per-instance} bounds that, for a given \MTS problem 
(i.e., given $y_i, h_i$, and $\perfBudgMin_{\grSet\setminus \selSet}$ in~\eqref{pr:dual}), can 
evaluate how far an algorithm is from the optimal \MTS solution. 
In order words, since we cannot guarantee than any efficient algorithm will do well in the worst-case, we are happy with evaluating (a posteriori) if an algorithm computed a good solution for a given problem instance. 
For this reason,
in the next section we develop the first per-instance sub-optimality bound for Problem~\ref{pr:dual}.
\section{Performance Guarantees}
\label{sec:guarantees}

\begin{figure}[t]
	\begin{center}
	\begin{minipage}{\textwidth}
	\hspace{-0.2cm}
	\includegraphics[width=.5\columnwidth]{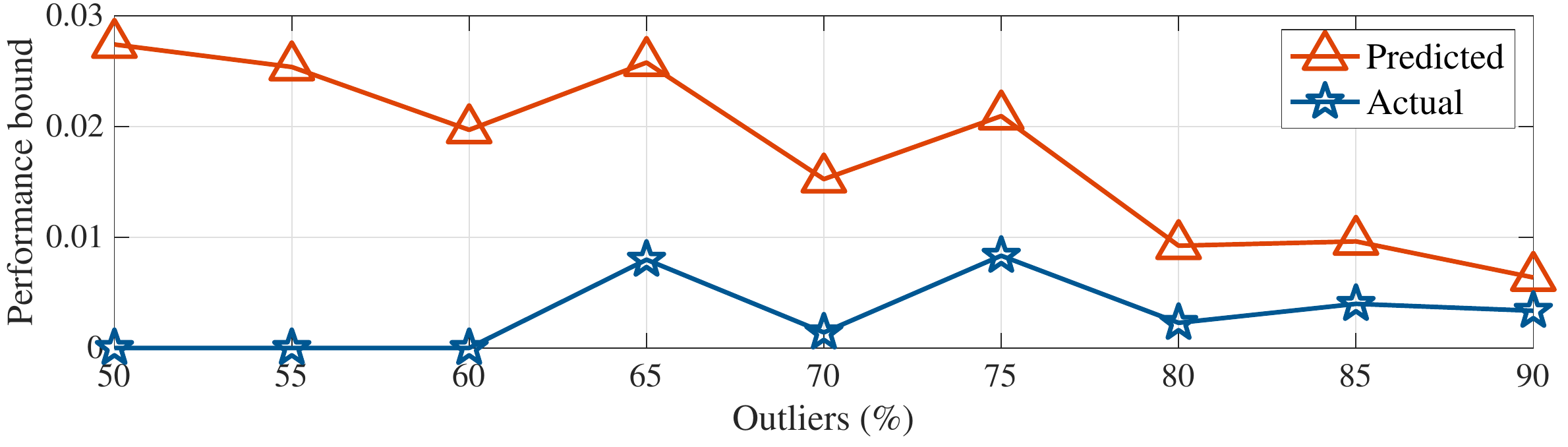} 
	\end{minipage}
\caption{Actual and predicted per-instance sub-optimality bound by Theorem~\ref{eq:bound} of heuristic algorithm (greedy in~\cite{nemhauser78analysis}), employed for small-scale instances of \MTS per the linear setup of Example~\ref{ex:linear_dec}.  
	 \label{fig:perf}}
	\vspace{-8mm} 
	\end{center}
\end{figure}

We present the first per-instance (i.e., a posteriori) sub-optimality bound for 
the \MTS Problem~\ref{pr:dual}. 
The bound is algorithm-agnostic (does not take assumption on the way ${\selSet}$ is generated), and is computable in $O(1)$ time. Also, we demonstrate its informativeness via simulations.  

\begin{theorem}[A posteriori sub-optimality bound]\label{th:guarantees} 
Consider the \MTS problem~\eqref{eq:MTS} and let 
$\selSet^\star$ be an optimal solution to~\eqref{eq:MTS}.
Also, for any candidate solution $\selSet$, let: 
\begin{itemize}
	\item $\res(\selSet)\doteq \min_{x\in\mathbb{X}} \;\;\sum_{i\in\grSet\setminus \selSet} \|h_i(y_i,x)\|^2$; i.e., $\res(\selSet)$ is the minimum residual error given the rejection $\selSet$;
	\item $\res^\star_\maxNumber\doteq \min_{\selSet\subseteq \grSet, |\selSet|\leq \maxNumber} \res(\selSet)$; i.e., $\res^\star_\maxNumber$ is the optimal residual error when at most~$\maxNumber$ measurements are rejected;
	\item $\res^\star\doteq \res(\selSet^\star)$; i.e., $\res^\star$ is the residual error for the optimal outlier rejection $\selSet^\star$.
\end{itemize}
Then, given a candidate solution $\selSet$, 
the following bound relates the residual error $\res(\selSet)$ of the candidate solution  with the 
residual error of an optimal solution rejecting the same number of outliers:
\begin{equation}\label{eq:guarantees}
\frac{\res(\selSet)-\res^\star_{|\selSet|}}{\res(\emptyset)-\res^\star_{|\selSet|}}\leq \apost_\selSet,
\end{equation}
where 
\begin{equation}\label{eq:bound}
\apost_\selSet\doteq\frac{\res(\selSet)}{\res(\emptyset)-\res(\selSet)}.
\end{equation}
Moreover, if it is also known that $|\selSet|\; \geq |\selSet^\star|$, then it holds:
\belowdisplayskip=-6pt
\begin{equation}\label{eq:bound_refined}
\frac{\res(\selSet)-\res^\star}{\res(\emptyset)-\res^\star}\leq \apost_\selSet.
\end{equation}
\hfill $\lrcorner$
\end{theorem}
Eq.~\eqref{eq:guarantees} quantifies the distance between the residual of the candidate solution and 
the residual of an optimal solution rejecting the same number of outliers $|\selSet|$. Intuitively, if we incorrectly pick outliers and obtain a residual error $\res(\selSet)$, there might exist a more clever selection that instead obtains $\res^\star_{|\selSet|} \ll \res(\selSet)$; 
on the other hand, we would like $\res(\selSet)$ and $\res^\star_{|\selSet|}$ to be as close as possible. For this reason,  the smaller $\apost_\selSet$, the closer the candidate selection is to the optimal selection.
For example, when $\apost_\selSet=0$, then $\res(\selSet)= \res^\star_{|\selSet|}$, 
i.e., we conclude that the algorithm returned a globally optimal solution (restricted to the ones rejecting $|\selSet|$ measurements).

Eq.~\eqref{eq:bound_refined} completes the picture by stating that 
 if the algorithm rejects at least as many measurements as the optimal solution ($|\selSet|\; \geq |\selSet^\star|$), 
 then the bound in eq.~\eqref{eq:bound_refined} compares the quality of $\selSet$ directly with the optimal residual error of $\selSet^\star$, the optimal solution of the \MTS Problem~\ref{pr:dual}. 

\begin{remark}[Quality of the bound]
We showcase the quality of the bound~\eqref{eq:bound} by considering the linear estimation Example~\ref{ex:linear_dec}. 
  We generate small instances for which we can compute the optimal solution and 
  evaluate the corresponding residual error $\res^\star$. In particular, we compute the optimal solution using CPLEX~\cite{CPLEXwebsite}, a popular library for mixed-integer linear programming, and we compare the optimal solution against a candidate solution $\selSet$. 
  We generate candidate solutions using a greedy algorithm~\cite{nemhauser78analysis}. 
  The greedy algorithm, at each iteration, rejects the measurement that induces the largest decrease in the residual error. Fig.~\ref{fig:perf} shows in blue the actual (true) approximation performance of the greedy algorithm ($\frac{\res(\selSet)-\res^\star}{\res(\emptyset)-\res^\star}$)
  and in red the bound $\apost_{\selSet}$ in eq.~\eqref{eq:guarantees}. The results are averaged over 10 Monte Carlo simulations.  The figure shows that the bound predicts well the actual sub-optimality ratio (note the scale of the $y$-axis) and its quality improves for increasing number of outliers. 
  \hfill $\lrcorner$
  \end{remark}

We remark that the bound~\eqref{eq:bound} can be also used to quantify the performance of existing algorithms, including \ransac. Having introduced our per-instance sub-optimality bounds, we step forward to a novel general-purpose algorithm for outlier rejection that empirically 
returns accurate solutions (and for which our bound $\apost_\selSet$ is typically close to zero).

\section{A General-Purpose Algorithm: \name}
\label{sec:algorithms}


\begin{algorithm}[t]
	\caption{\mbox{\nameLong (\name)}}
	\begin{algorithmic}[1]
		\REQUIRE \mbox{ } 
		\begin{enumerate}[leftmargin=-.5cm,label=\textbullet]
            \item $v$: minimum nr. of measurements required by global solver;
            \item $\gamma$: discount factor for outlier threshold (default $\gamma=0.99$);
            \item $\delta$: convergence threshold;
            \item $T$: nr. of iterations to decide convergence (default $T=2$);
            \item $\bar{g}$: maximum nr. of extra rejections per iteration.
	   \end{enumerate}

    \ENSURE outlier set $\selSet$.
    \STATE $t \gets 0$; \hspace{.5mm} $ \selSet_t \gets \emptyset$; \hspace{.5mm}$ g \gets \bar{g} $; \hspace{.5mm}$c \gets 0$; \hspace{.5mm}$\tau\gets \max_{i\in\grSet}r_i(\emptyset)$;\label{line:initialization}
    \WHILE {$\text{true}$}
      \STATE $t \gets t + 1$; \hspace{1mm}$\selSet_t \gets \selSet_{t-1}$;
      \WHILE[\textcolor{gray}{discount threshold \& update}] {$\selSet_t = \selSet_{t-1}$}\label{line:start_first_while}
        \STATE $\calI \gets $ indices of $g$~largest $r_i(\selSet_{t-1})$ across $i\in\grSet$;\label{line:start_update_set}
        \STATE $\selSet_t \gets \{i\in\calI  \text{ and } r_i(\selSet_{t-1})\geq \tau\}$;\label{line:finish_update_set}
        \IF[\textcolor{gray}{discount}] {$\selSet_t = \selSet_{t-1}$ or $\selSet_t = \emptyset$}\label{line:start_if_discount}
          \STATE $\tau \gets \gamma \min\left\{\tau,\max_{i \in\grSet\setminus\selSet_t} r_i(\selSet_t)\right\}$;\label{line:discount}
        \ENDIF
      \ENDWHILE
        \STATE $g \gets g + \bar{g}$;\label{line:update_group}
        \IF[\textcolor{gray}{terminate}] {$|\selSet_t|\; = |\grSet| - v$}\label{line:start_if_terminate_min_number}
          \RETURN $\selSet_{t}$.\label{line:end_if_terminate_min_number}
        \ENDIF
      \IF[\textcolor{gray}{check convergence}]{$|r(\selSet_{t})-r(\selSet_{t-1})|\; \leq \delta$}\label{line:start_if_possible_convergence}
        \STATE $c \gets c + 1$;
        \IF[\textcolor{gray}{terminate}] {$c = T$}\label{line:start_if_convergence}
          \RETURN $\selSet_{t}$. 
          \label{line:end_if_convergence}
        \ENDIF
      \ELSE[\textcolor{gray}{reset convergence counter}]
        \STATE $c \gets 0$.
      \ENDIF
    \ENDWHILE
	\end{algorithmic}\label{alg:ADAPT}
\end{algorithm}

We introduce a novel algorithm for outlier rejection that we name \textit{\nameLong (\name)}. 
The algorithm starts by processing all measurements and at each iteration it trims measurements with 
residuals larger than a threshold. It is \emph{adaptive} in that it dynamically decides the threshold at each iteration 
(hence relaxing the need for a threshold $\perfBudgMin_{\grSet\setminus \selSet}$). Moreover, it is not greedy in that it can reject multiple measurements at each iteration while it keeps revisiting the quality of previously rejected outliers.\footnote{In our tests we found that a greedy algorithm similar to~\cite{nemhauser78analysis} tends to converge to poor outlier rejection decisions and is typically slow for practical applications, since it has quadratic runtime in the number of measurements.} 

\begin{assumption}[Global solver]
\name assumes the availability of a black-box solver that can (even approximately) solve 
the outlier-free problem~\eqref{eq:nls} to optimality.
\end{assumption}

Luckily, for all problems in Section~\ref{sec:examples}, there exist (outlier-free) global solvers, including~\cite{Rosen18ijrr-sesync,Briales18cvpr-global2view,Horn87josa}.

\myParagraph{Description of \name} The preudo-code of \name is given in Algorithm~\ref{alg:ADAPT}.  Here, we use the additional notation:
\begin{itemize}
	\item Let $x^\star(\selSet) \in \arg\min_{x\in\mathbb{X}} \;\;\sum_{i\in\grSet\setminus \selSet} \|h_i(y_i,x)\|^2$; i.e., $x^\star(\selSet)$ is an estimator of $x$ given an outlier selection~$\selSet$. 
	\item Let $r_i(\selSet)\doteq \|h_i(y_i,x^\star(\selSet))\|^2$; i.e., $r_i(\selSet)$ is the residual of the measurement $i$, given an outlier selection $\selSet$. 
\end{itemize}
Per Algorithm~\ref{alg:ADAPT}, \name executes five distinctive operations:

\noindent\paragraph{Initialization (line~\ref{line:initialization})} 
\name initializes the iteration counter $t = 0$ and the current candidate outlier set $\selSet_t$ with the empty set.
 It also initializes $g$, the \emph{outlier group size}, which constrains the maximum number of measurements 
 that can be deemed as outliers in a single iteration of the algorithm. 
 Moreover, \name initializes the counter $c = 0$: this is used to decide whether convergence has been reached.  Finally, it  initializes the outlier threshold $\tau$ with the value of the largest residual across all measurements. 
 Note that computing $r_i(\selSet)$ (for any $\selSet \subseteq \grSet$) requires calling the global solver on the measurements $\grSet \setminus \selSet$, and then evaluating the residual errors for all measurements in $\grSet$.

\paragraph{Outlier set update (lines~\ref{line:start_update_set}-\ref{line:finish_update_set})}  Given the current threshold $\tau$ and group size $g$, the algorithm updates the outlier set in two steps: first (line~\ref{line:start_update_set}), it finds the set of measurements $\calI$ with the $g$ largest residuals among all measurements in $\grSet$;\footnote{Note that the selection is performed over all measurements $\grSet$, potentially revisiting 
measurements that were previously deemed to be outliers.} 
second (line~\ref{line:finish_update_set}), the algorithm updates the outlier set as the collection of all the measurements in $\calI$ whose residual exceeds the outlier threshold~$\tau$.    
 \paragraph{Outlier threshold update (lines~\ref{line:start_if_discount}-\ref{line:discount})}  
If the updated outlier set $\selSet_t$ remains the same as that in the previous iteration $\selSet_{t-1}$  (line~\ref{line:start_if_discount}), then the outlier threshold $\tau$ is not tight enough. As a result, the algorithm updates $\tau$ with a discounted value $\gamma < 1$ (line~\ref{line:discount}).  This process is repeated as long as necessary, as indicated by the ``while'' loop in line~\ref{line:start_first_while}. 
\paragraph{Outlier group size update (line~\ref{line:update_group})} After each iteration~$t$, \name increases the outlier group size $g$ by $\bar{g}$. This has the effect of increasing the 
maximum number of measurements that can be deemed as outliers in future iterations: intuitively, \name is conservative in rejecting measurements at the beginning (small initial $g = \bar{g}$) , while it gets more and more aggressive by gradually increasing $g$.  
 \paragraph{Termination}
\name terminates when one of the following two conditions is satisfied.  
First (lines~\ref{line:start_if_terminate_min_number}-\ref{line:end_if_terminate_min_number}), 
it may terminate when all but a number $v$ of measurements have been rejected, where $v$ is the minimum number of measurement that the global solver needs to solve the problem (for example, in 3D registration, $v=3$).  Second (lines~\ref{line:start_if_convergence}-\ref{line:end_if_convergence}), \name may terminate if convergence has been achieved.  In particular, if the algorithm observes for $T$ consecutive times that the absolute value of the residuals function changes 
by less than $\delta$, then it terminates ($c$ counts the number of consecutive times \mbox{a decrease smaller than $\delta$ is observed).}

\begin{remark}[Complexity and practicality]
The termination condition in line~\ref{line:start_if_terminate_min_number}  guarantees the termination of the algorithm with at most $|\grSet|-v$ calls of the global solver.
\name terminates faster as one increases the outlier group size $\bar{g}$, the convergence thresholds $\delta$, and/or as one decreases the  discount factor $\gamma$ and the number $T$ of iterations to decide convergence. Overall, the linear runtime (in the number of measurements) of \name  makes the algorithm practical in real-time applications where fast global solvers are available.
\end{remark}

\begin{remark}[vs. \ransac]
While \ransac builds an inlier set by sampling small (minimal) sets of measurements, 
\name iteratively prunes the overall set of measurements. 
Arguably, this gives \name a ``global vision'' of the measurement set as we showcase in the experimental section. 
\ransac assumes the availability of fast minimal solvers, while \name assumes the availability of fast 
global (non-minimal) solvers. 
Finally, \ransac is not suitable for high-dimensional problems where is becomes exponentially more difficult to 
sample an outlier-free set~\cite{Bustos18pami-GORE}. 
On the other hand, \name is deterministic and guaranteed to terminate in a number of iterations bounded by the 
number of measurements.
\end{remark}
\section{Experiments and Applications}
\label{sec:experiments}

We evaluate \name against the state of the art in three spatial perception problems: 3D registration (Section~\ref{sec:exp-registration}), two-view geometry (Section~\ref{sec:exp-2view}), and SLAM (Section~\ref{sec:exp-slam}). 
The results show that \name outperforms \ransac in terms of accuracy and scalability, and often outperforms specialized outlier rejection methods (in particular for SLAM) while being a general-purpose algorithm. 
Finally, the tests show that the performance bounds of Section~\ref{sec:guarantees} are informative and can be used to assess the outlier rejection outcomes. All results are averaged over 10 Monte Carlo runs.
	\subsection{Robust Registration}
\label{sec:exp-registration}

\renewcommand{\myhspace}{\hspace{-3mm}}
\renewcommand{\mpw}{4.5cm}


\begin{figure}[t]
	\begin{center}
	\begin{minipage}{\textwidth}
	\hspace{-0.5cm}
	\begin{tabular}{c}%
			\begin{minipage}{.5\columnwidth}%
			\centering%
			\includegraphics[width=1.0\columnwidth]{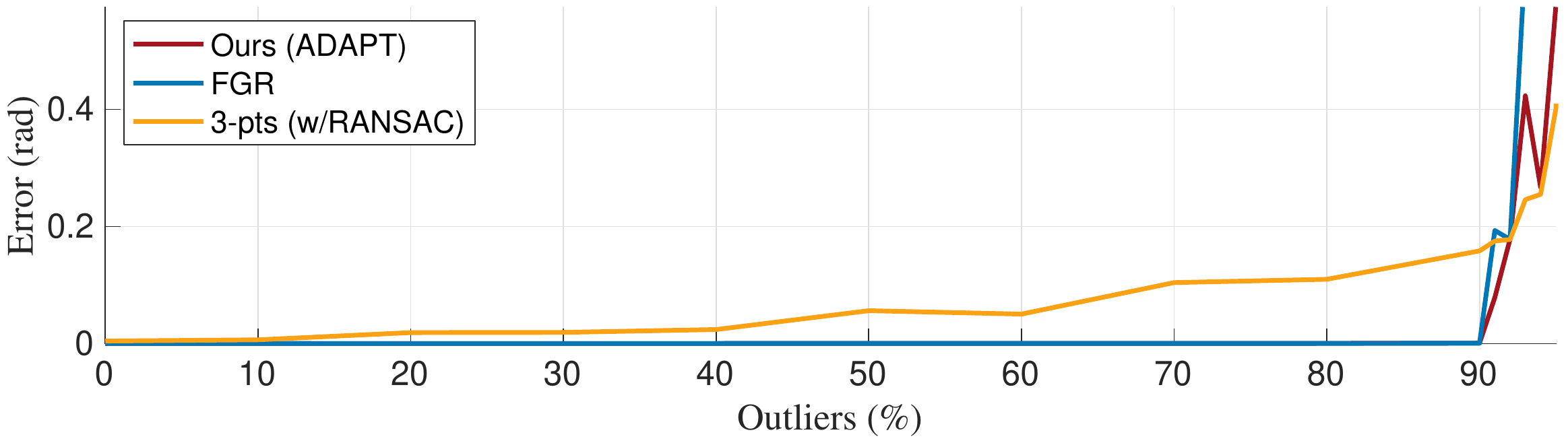} \\
			(a) Rotation Error
			\end{minipage}
\\
			\begin{minipage}{.5\columnwidth}%
			\centering%
			\includegraphics[width=1.0\columnwidth]{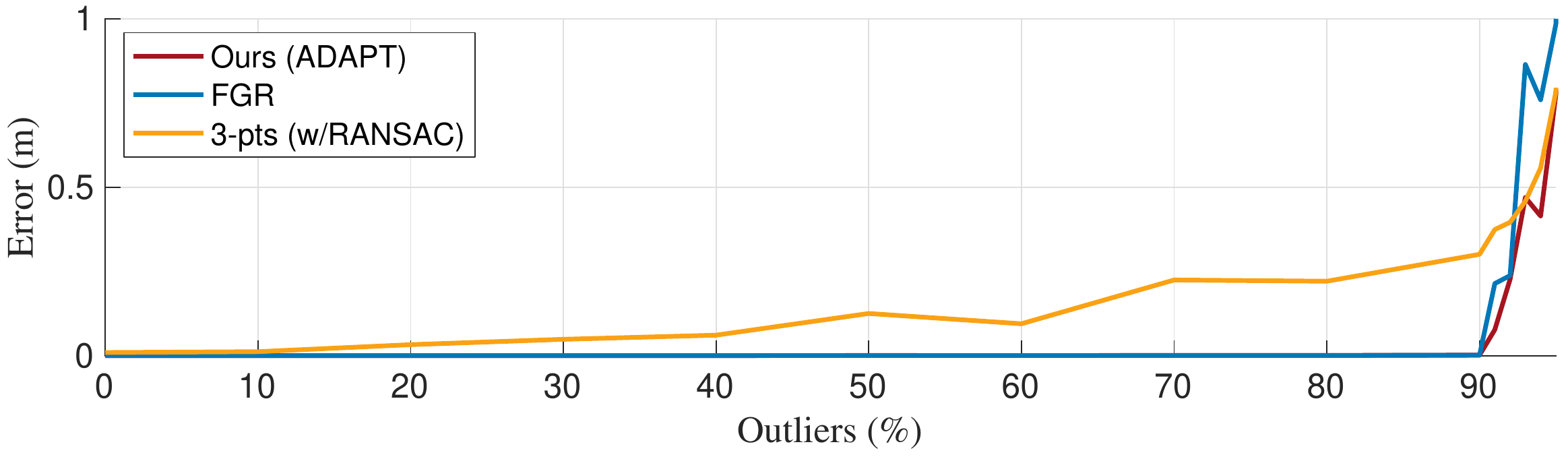} \\
			(b) Translation Error
			\end{minipage}
		\end{tabular}
	\end{minipage}
	\caption{3D registration: rotation and translation errors for \name, \FGR~\cite{Zhou16eccv-fastGlobalRegistration}, and \ransac on the Bunny dataset for increasing outlier percentages.}
	 \label{fig:regBunny}
	\end{center}
	\vspace{-5mm}
\end{figure}



\myParagraph{Experimental setup}
We test \name on two standard datasets for 3D registration: the Stanford Bunny and the ETH Hauptgebaude~\cite{pomerleau2012ethpc}. In both cases we downsample the point clouds obtaining 453 points for Bunny and 3617 points for ETH. 
For each point cloud $\Pone$ we generate a second point cloud $\Ptwo$ by applying a random rigid transformation and adding 
noise and outliers. The (inlier) noise standard deviation is set to  0.025\% and 0.05\% of the point cloud diameter respectively.
Outliers are generated by replacing a subset of the points in $\Ptwo$ with random points 
uniformly sampled in the bounding box containing $\Pone$.\omitted{The random sampled is than compared to the ground truth to make sure it far enough from the ground-truth measurement, i.e. $(x-x_{gt})\Sigma^{-1}(x-x_{gt})> F^{-1}_{\chi^2}(1-\epsilon, 3)$ where $F^{-1}_{\chi^2}(\cdot,\cdot)$ is inverse of the chi-square cumulative distribution function with probability $1-\epsilon$ and degree-of-freedom $3$.}
In each iteration, \name uses Horn's method~\cite{Horn87josa} as global solver. 
We benchmark \name against \emph{Fast Global Registration} (\FGR)~\cite{Zhou16eccv-fastGlobalRegistration} and the three-point \ransac. 
We set the maximum number of iterations in \ransac to 1000 and use default parameters for \FGR.
All methods are implemented in \MATLAB. 

\myParagraph{Results}  
Fig.~\ref{fig:regBunny} shows the (average) translation and rotation errors for the estimates computed by \name, \FGR, and \ransac on the Bunny dataset for increasing outlier percentages. 
\name performs on-pair with \FGR which is a specialized robust solver for 3D registration and 
they both achieve practically zero error for up to 90\% of outliers, after which they both break.  
\ransac starts performing distinctively worse early on and is dominated by the other techniques (after 90\% all techniques fail to provide a satisfactory estimate).
We obtain similar results on the ETH dataset hence for space reasons we report them in Appendix~\ref{sec:supp_exp}.

For both the Bunny and ETH datasets, we compute the sub-optimality bound for the result of \name, using Theorem~\ref{th:guarantees}. The plot of the bound is given in Appendix~\ref{sec:supp_exp}; the bound remains around $10^{-5}$, confirming that \name remains close to the optimal outlier selection.
 The runtime of \name is comparable to \FGR and is reported in Appendix~\ref{sec:supp_exp}.

  	\subsection{Robust Two-view Geometry}
\label{sec:exp-2view}

\renewcommand{\myhspace}{\hspace{-3mm}}
\renewcommand{\mpw}{4.5cm}


\begin{figure}[h!]
	\begin{center}
		\begin{minipage}{\textwidth}
			\hspace{-0.5cm}
			\begin{tabular}{cc}%
				\begin{minipage}{.5\columnwidth}%
					\centering%
					\includegraphics[width=1.0\columnwidth]{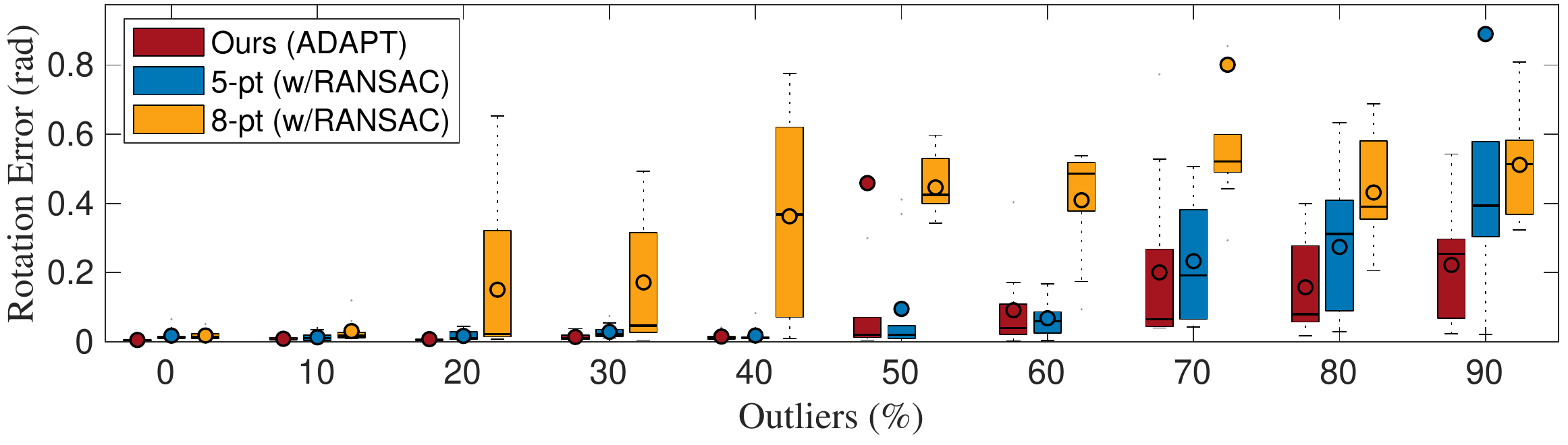}  \\
					(a) Rotation Error 
				\end{minipage}
				\\
				\begin{minipage}{.5\columnwidth}%
				\centering%
				\includegraphics[width=1.0\columnwidth]{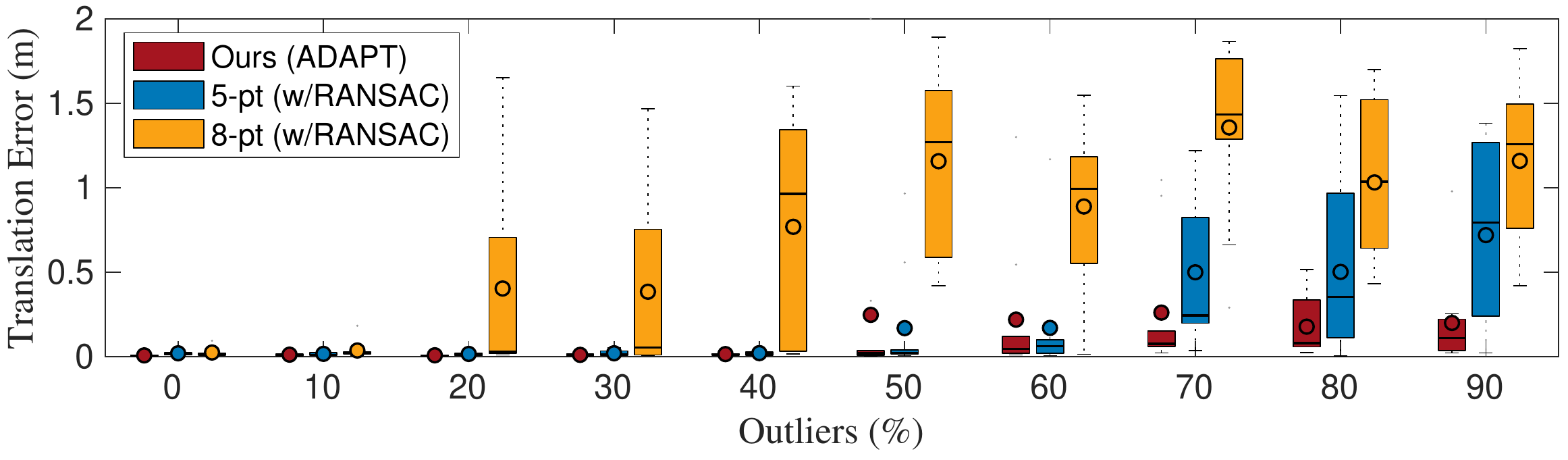} \\
				(b) Translation Error 
				\end{minipage}
			\end{tabular}
		\end{minipage}
		\caption{Two-view geometry: rotation and translation errors for \name, five- and eight-point \ransac, on a synthetic dataset for increasing outliers.}
		\label{fig:viewSynthetic}
	\end{center}
	\vspace{-5mm}
\end{figure}

\begin{figure}[h!]
	\begin{center}
		\begin{minipage}{\textwidth}
			\hspace{-0.5cm}
			\begin{tabular}{cc}%
				\begin{minipage}{.5\columnwidth}%
					\centering%
					\includegraphics[width=1.0\columnwidth]{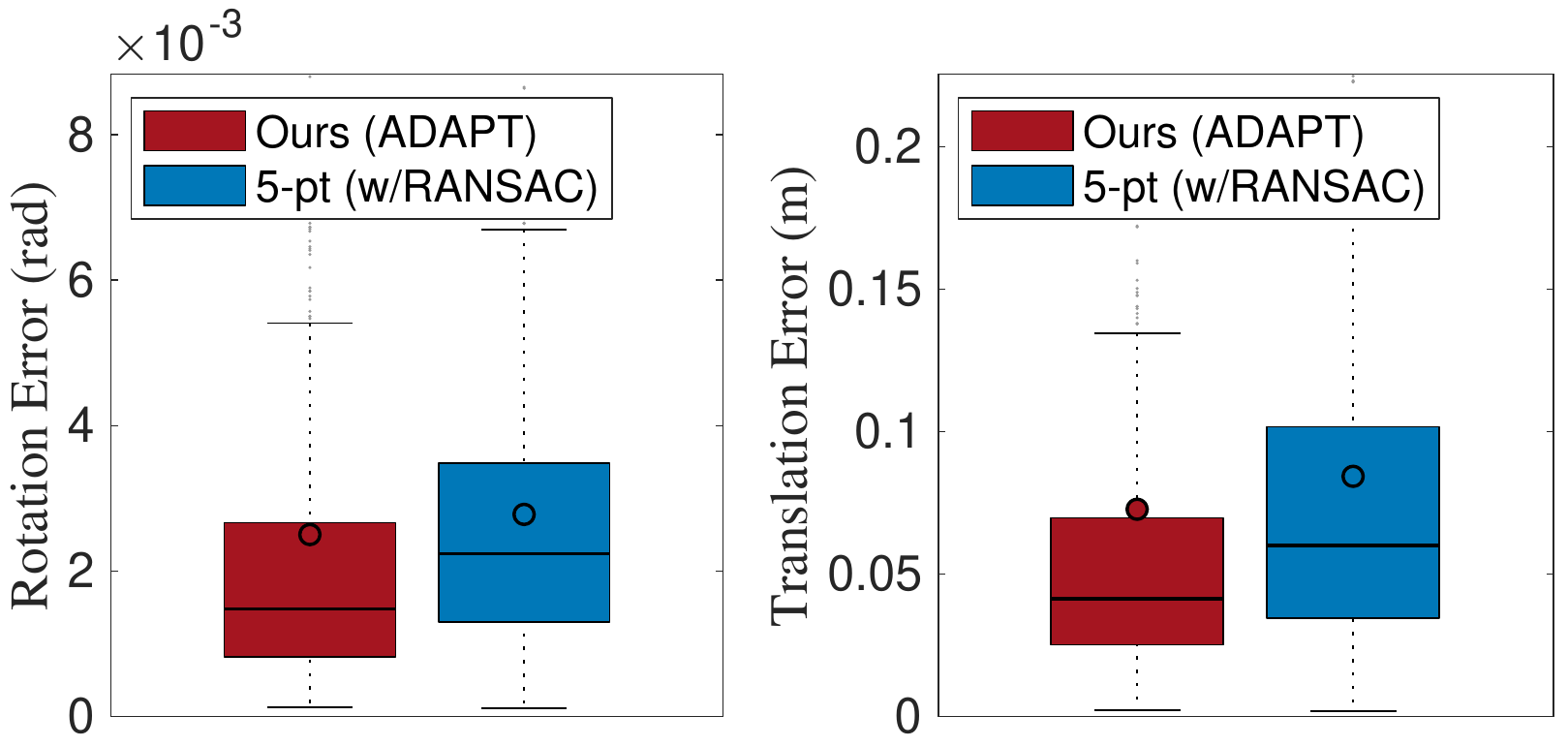} 
				\end{minipage}
			\end{tabular}
		\end{minipage}
		\caption{Two-view geometry: rotation and translation errors for \name, and five-point \ransac, on the sequence \text{MH_01} of the EuRoC dataset.}
		\label{fig:viewEuroc}
	\end{center}
		\vspace{-5mm}
\end{figure}


\myParagraph{Experimental setup}
We tested \name on both synthetic data and on the \text{MH_01} sequence of the \emph{EuRoC} dataset~\cite{burri2016euroc}. To generate the synthetic data 
we place the first camera at the origin (identity pose) and place the second camera randomly within a bounded region. Then, we generate a random point cloud within the field-of-view of both camera.
The points projected on the camera frame are finally corrupted with Gaussian noise, and outliers are added. 
For the EuRoC dataset, we extract Harris corners in each frame and track keypoints in consecutive frames using Lucas-Kanade feature tracking as in~\cite{Carlone18tro-attentionVIN}; the results are then averaged across all pairs of consecutive frames in the sequence.
In each iteration, \name uses Briales' QCQP relaxation~\cite{Briales18cvpr-global2view} as global solver.
We benchmarked \name against Nister's five-point~\cite{Nister04pami} and the eight-points algorithm~\cite{hartley1997eightpt} within \ransac.

\myParagraph{Results}  
Fig.~\ref{fig:viewSynthetic} shows the box-plot of translation and rotation errors for the estimates of 
\name, the five- and eight-point \ransac on the synthetic dataset. 
\name outperforms the other techniques across all the spectrum. 
\name and five-point perform on-pair till 40\% of outliers.  Beyond that point, the five-point method attains considerably higher errors than \name (50\% to 100\% more in rotation; and more than 300\% more in translation).  
The eight-point method results in higher errors than the five-point across the spectrum. 

Fig.~\ref{fig:viewEuroc} shows the results on the EuRoC dataset focusing on the comparison between \name and the five-point \ransac. \name achieves a mean rotation error of $2.5\cdot 10^{-3}$rad versus $2.8\cdot 10^{-3}$rad of the five-point.  Similarly for the translation error: 0.075m for \name versus 0.09m for the five-point.  For visualization purposes we cut the translation box-plot in Fig.~\ref{fig:viewEuroc} above 0.25m error: in reference to the rest of the plot, we report that \name exhibits translation errors larger than $1$ in 10\% of the frames, whereas only $1\%$ of the five-point estimates have translation errors larger than~$1$.

For the synthetic dataset, the typical value for the sub-optimality bound achieved by \name is $0.2$.  That is, \name makes a rejection that achieves an error that is at most 20\% away from the optimal, even in the presence of 90\% of outliers.  
The runtime of \name is reported in Appendix~\ref{sec:supp_exp}: our approach is one order of magnitude slower than the five-point method, mainly due to the  relatively high runtime of the global solver~\cite{Briales18cvpr-global2view}, which is called in each iteration.
  	\subsection{Robust SLAM}
\label{sec:exp-slam}

\renewcommand{\myhspace}{\hspace{-3mm}}
\renewcommand{\mpw}{4.5cm}

%

\begin{figure}[t]
	\begin{center}
	\begin{minipage}{\textwidth}
	\hspace{-0.5cm}
	\begin{tabular}{c}%
			\begin{minipage}{.5\columnwidth}%
			\centering%
			\includegraphics[width=1.0\columnwidth]{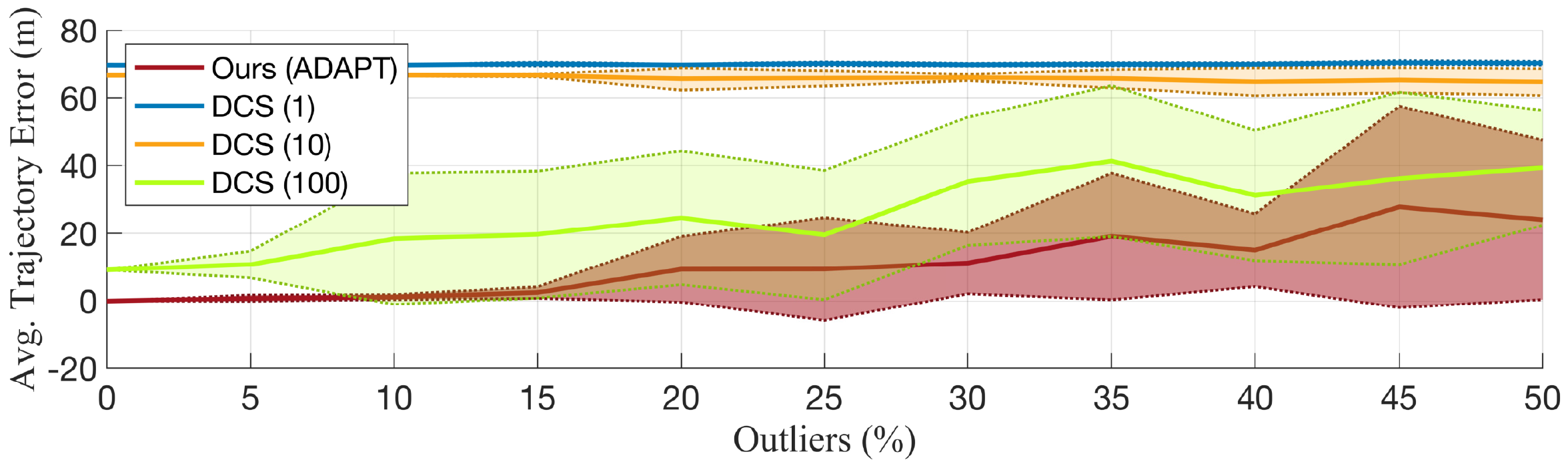} \\
      (a) MIT 
			\end{minipage}
\\
			\begin{minipage}{.5\columnwidth}%
			\centering%
			\includegraphics[width=1.0\columnwidth]{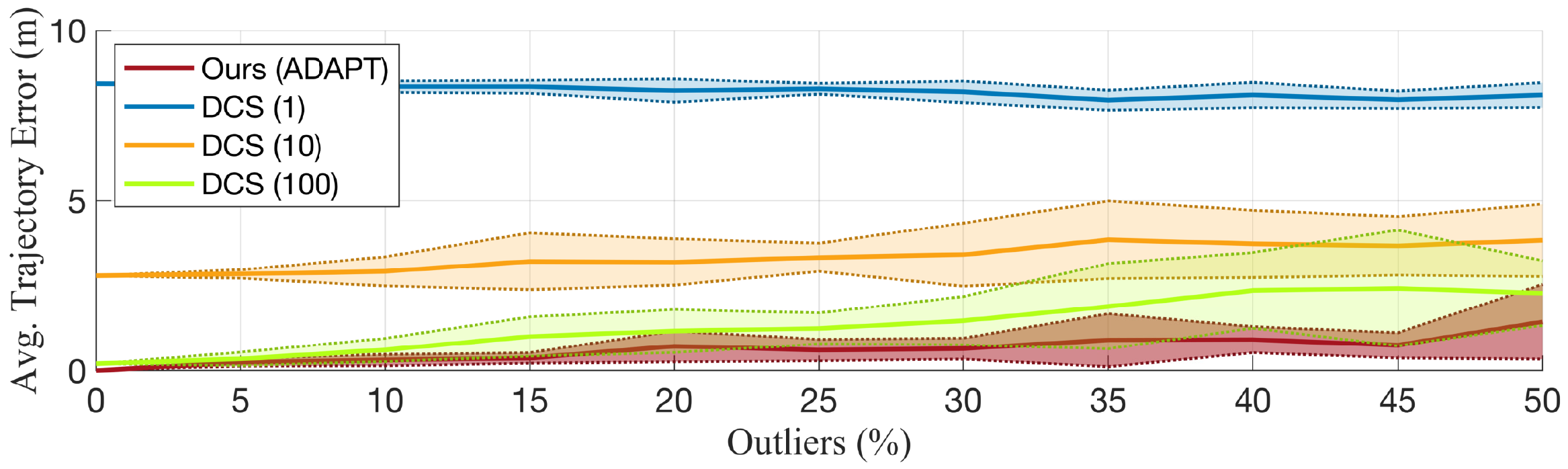} \\
			(b)  CSAIL
			\end{minipage}
		\end{tabular}
	\end{minipage}
	\caption{2D SLAM: Average Trajectory Error of \name and \DCS for increasing outliers in the MIT and CSAIL datasets.}
	 \label{fig:slam2D}
	\end{center}
	\vspace{-5mm}
\end{figure}

\begin{figure}[t]
	\begin{center}
	\begin{minipage}{\textwidth}
	\hspace{-0.5cm}
	\begin{tabular}{c}%
			\begin{minipage}{.5\columnwidth}%
			\centering%
			\includegraphics[width=1.0\columnwidth]{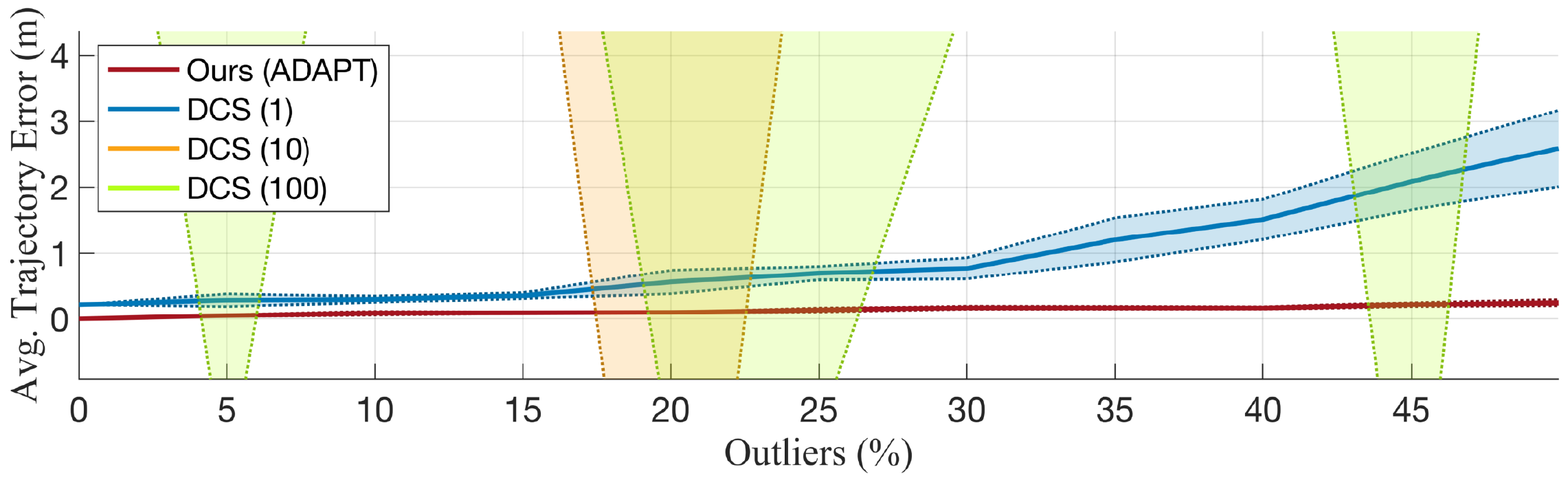} \\
			\end{minipage}
		\end{tabular}
	\end{minipage}
	\caption{3D SLAM: \emph{Detailed view }of the Average Trajectory Error of \name and \DCS on the Sphere 2500 dataset for increasing outliers.  \DCS\!(10) and \DCS\!(100) have error of at least 25 meters. Complete figure located in Appendix~\ref{sec:supp_exp}.}
	 \label{fig:slam3D}
	\end{center}
	\vspace{-5mm}
\end{figure}


\myParagraph{Experimental setup}
We test \name on standard 2D and 3D SLAM benchmarking datasets and report extra results on synthetic datasets in Appendix~\ref{sec:supp_exp}. We spoil existing datasets with spurious loop closures: we sample random pairs of nodes and we add an outlier 
relative pose measurement between them, where the relative translation is sampled in the ball of radius 5m and the rotation 
is sampled uniformly at random in \SOtwo or \SOthree.
The ground truth trajectory is generated by optimizing the problem with \emph{SE-Sync}~\cite{Rosen18ijrr-sesync} 
before adding outliers.
We also use \emph{SE-Sync} as the global solver for \name. 
We test the following datasets, described in~\cite{Carlone14tro-SO2,Rosen18ijrr-sesync}: \emph{MIT} (2D), \emph{Intel} (2D), \emph{CSAIL} (2D), and \emph{Sphere2500} (3D). 
We also test a simulated $5\times 5\times 5$ 3D grid dataset (results in Appendix~\ref{sec:supp_exp}).
We benchmark \name against \DCS~\cite{Agarwal13icra}; we report \DCS results for three choices of the robust kernel size: \{1,10,100\} (the default value is 1, see~\cite{Agarwal13icra}).

\myParagraph{Results}  
\name outperforms \DCS (independently on the choice of the kernel size) across all outlier percentages.  

\textit{2D SLAM}: In the \textit{MIT} dataset (Fig.~\ref{fig:slam2D}a), a particularly challenging dataset, 
\name is insensitive to up to 20\% of outliers. All variants of \DCS fail to produce an error smaller than 10 meters even in the absence of outliers: \dcs is an iterative solver, hence is may converge to local minima when bootstrapped from a bad initial guess (the odometric estimate is particularly noisy in MIT). \name leverages SE-sync, which is a global solver, hence is able to converge to the correct solution. 
 In the \textit{CSAIL} dataset (Fig.~\ref{fig:slam2D}b), \name also dominates \DCS while \DCS performance is acceptable 
when the kernel size is equal to 100. Unfortunately, the choice of a ``good'' kernel size is dataset-dependent and 
it can make the difference between an adequate performance in \DCS and a catastrophic failure.
\name also dominates \DCS in the INTEL dataset, whose statistics are reported in Appendix~\ref{sec:supp_exp}.

\textit{3D SLAM}: In the \textit{Sphere} dataset (Fig.~\ref{fig:slam3D}), \name achieves and error below 0.25m across all outlier percentages.  On the other hand, \DCS starts  with an error of 0.25 meters (0\% outliers) and ends up with at least 2.5 meters error (50\% outliers). Again, \name, a general-purpose approach for outlier rejection, outperforms specialized techniques for robust SLAM. Appendix~\ref{sec:supp_exp} also reports similar conclusions and results for the  \textit{3D grid} dataset.

For \textit{2D SLAM}, the typical value for the sub-optimality bound achieved by \name, per Theorem~\ref{th:guarantees}, is 0.1 (MIT) and 0.01 (Intel and CSAIL) across all spectrum of outlier percentages. For \textit{3D SLAM}, the typical value of the bound achieved by \name is 0.1 (3D grid) and 0.01 (sphere).

\name is one to two orders slower than \DCS. This is due to the repeated calls to SE-sync and is further aggravated in the 3D case by the fact that SE-sync tends to be slow in the presence of outliers (the Riemannian staircase method~\cite{Rosen18ijrr-sesync} requires multiple staircase iterations since the rank of the relaxation increases in the presence of outliers).

\section{Related Work}
\label{sec:relatedWork}

We extend the literature review in Section~\ref{sec:intro}, across robotics and computer vision (Section~\ref{sec:relatedWork_robotComp}), and statistics and control (Section~\ref{sec:relatedWork_statistics}), as well as, across sub-optimality guarantees in outlier-robust estimation (Section~\ref{subsec:relWork_guarantees}).

\subsection{Outlier-robust estimation in robotics and computer vision}\label{sec:relatedWork_robotComp}

Outlier-robust estimation is an active research area in robotics and computer vision for decades~\cite{Meer91ijcv-robustVision,Stewart99siam-robustVision,Bosse17fnt}. 
Traditionally, low-dimensional problems are solved using \ransac~\cite{Chen99pami-ransac} (e.g., 3D registration, and two-view geometry).
\ransac is efficient and effective for low outlier rates~\cite{Speciale17cvpr-consensusMaximization}.  However, the need to cope with higher outlier rates pushed research towards global optimization methods, such as \emph{branch-and-bound} (\BnB)~\cite{Bazin14eccv-robustRelRot,hartley2009ijcv-global,Zheng11cvpr-robustFitting,Li09cvpr-robustFitting,Speciale17cvpr-consensusMaximization}, and \emph{mixed-integer programming} (MIP)~\cite{Chin16cvpr-outlierRejection}.
Nevertheless, for high-dimensional problems (e.g., SLAM), \ransac, \BnB, and MIP are typically  slow to be practical: in the worst-case, they require exponential run-time.  Hence, research has also focused on  M-estimators, in conjunction with either non-convex optimization~\cite{Sunderhauf12iros,Agarwal13icra,Olson12rss}, or convex relaxations~\cite{Lajoie19ral-DCGM}, possibly also including decision variables for the  outlier rejection~\cite{black1996ijcv-unification,Lajoie19ral-DCGM,Sunderhauf12iros,Agarwal13icra}.
In more detail, representative outlier-robust methods for 3D registration, two-view geometry, and SLAM are the following. 

\myParagraph{Robust 3D registration} 
The goal is to find the transformation (translation and rotation) that aligns two point clouds.
In the presence of outliers, created by incorrect point correspondences, one typically 
resorts to \ransac~\cite{Fischler81,Chen99pami-ransac}, along with a 3-point minimal solver 
---the outlier-free problem admits well-known closed form solutions~\cite{Arun87pami,Horn87josa}.
But when the outliers' number is more than 50\%, \ransac tends to be slow, and brittle~\cite{Speciale17cvpr-consensusMaximization,Bustos18pami-GORE}.  Thereby, recent approaches adopt either robust cost functions~\cite{MacTavish15crv-robustEstimation,Bosse17fnt,Zhou16eccv-fastGlobalRegistration}, or \BnB~\cite{Yang16pami-goicp,Campbell17cvpr-globallyOptRobustTwoView,Bustos18pami-GORE}: Zhou\setal~\cite{Zhou16eccv-fastGlobalRegistration} propose \emph{fast global registration (FGR)}, which is based on the Geman-McClure robust cost function;
Yang\setal~\cite{Yang16pami-goicp} propose a  approach; Campbell\setal~\cite{Campbell17cvpr-globallyOptRobustTwoView} employ \BnB to search the space of camera poses, guaranteeing global optimality without requiring a pose prior;
and Bustos\setal~\cite{Bustos18pami-GORE} add a pre-processing step, that removes gross outliers before \ransac or \BnB.
Other approaches, that iteratively compute point correspondences, include \emph{iterative closest point} (ICP)~\cite{Besl92pami,Segal06rss-generalizedICP},
and \emph{trimmed iterative closest point algorithm}~\cite{Chetverikov05ivc-tICP}; all require an accurate initial guess~\cite{Zhou16eccv-fastGlobalRegistration}.

\myParagraph{Robust two-view geometry} 
The problem consists in estimating the relative pose (up to scale) between 
two images, given pixel correspondences. In robotics, \ransac is again the go-to approach~\cite{Scaramuzza11ram}, typically in conjunction with
Nister's 5-point method~\cite{Nister03cvpr}, a minimal solver ---other minimal solvers exist when one is given a reference 
direction~\cite{Naroditsky12pami-3point}, the relative rotation~\cite{Kneip11bmvc}, or motion constraints~\cite{Scaramuzza11ijcv}.
In computer vision, recent approaches are investigating the use of provably-robust techniques, typically based on \BnB:
Hartley\setal\cite{hartley2009ijcv-global} propose a \BnB approach for $\ell_\infty$ optimization in one-view and two-view geometry;
Li~\cite{Li09cvpr-robustFitting} uses \BnB and mixed-integer programming for two-view geometry;
Bazin\setal~\cite{Bazin14eccv-robustRelRot} use \BnB for rotation-only estimation;
Chin\setal~\cite{Chin16cvpr-outlierRejection} propose a method to remove outliers in conjunction with mixed-integer linear programming;
Zheng\setal~\cite{Zheng11cvpr-robustFitting} use \BnB to estimate the fundamental matrix;
and Speciale\setal~\cite{Speciale17cvpr-consensusMaximization} improve \BnB approaches by including linear matrix inequalities.
\BnB is typically slow~\cite{Li09cvpr-robustFitting}, \mbox{but it is able to tolerate high outlier rates.}

\myParagraph{Robust SLAM}
Outlier-robust SLAM has traditionally relied on M-estimators; e.g.,~\cite{Bosse17fnt}.  
Olson and Agarwal~\cite{Olson12rss} use a max-mixture distribution to approximate multi-modal
measurement noise. 
S\"{u}nderhauf and Protzel~\cite{Sunderhauf12iros,Sunderhauf12icra} augment the problem with latent binary variables responsible for 
deactivating outliers.  Tong and Barfoot~\cite{Tong11icra-robustSlam,Yong13astro-robustSLAM} propose algorithms to classify outliers via chi-square statistical tests that account for the effect of noise in the estimate.  Latif~\etal \cite{Latif12rss} propose \emph{realizing, reversing, and recovering} (\rrr), which performs loop-closure outlier rejection, by clustering measurements together and checking for consistency using the chi-square inverse test as an outlier-free bound.  A pair-wise consistency check for multi-robot SLAM, \emph{pair-wise consistency maximization} (\pcm), was proposed by Mangelson~\etal \cite{Mangelson18icra}.
Agarwal \etal~\cite{Agarwal13icra} propose \emph{dynamic covariance scaling} (\dcs), which adjusts the measurement covariances to reduce the influence of outliers.
Lee~\etal~\cite{Lee13iros} use expectation maximization. 
The papers above rely either on the availability of an initial guess for optimization, or on parameter tuning. 
Recent work also includes convex relaxations for outlier-robust SLAM~\cite{Wang13ima,Carlone18ral-robustPGO2D,Arrigoni18cviu,Lajoie19ral-DCGM}. 
Currently,  
only~\cite{Lajoie19ral-DCGM} provides so far 
sub-optimality guarantees, which however degrade with the quality of the relaxation.  Additionally,~\cite{Lajoie19ral-DCGM}  requires parameter tuning.

\subsection{Outlier-robust estimation in statistics and control}\label{sec:relatedWork_statistics}

Outlier-robust estimation has received long-time attention in statistics and control~\cite{Huber64ams,Kalman60}.  It has a fundamental applications, such as prediction and learning~\cite{Diakonikolas201focs-robustEstimation}, linear decoding~\cite{Candes05tit}, and secure state estimation for control~\cite{Pasqualetti13tac-attackDetection}.  

Outlier-robust estimation in statistics, in its simplest form, aims to learn the mean and covariance of an unknown distribution, given both a portion of \emph{noiseless} i.i.d.~samples, and a portion of arbitrarily corrupted samples (outliers); particularly, the number of outliers is assumed known.\footnote{In contrast, eq.~\eqref{eq:nls}'s framework (and, correspondingly, \MTS's)  considers the estimation of an unknown parameter given some measurements, which framework is equivalent to the estimation of the parameter given non-i.i.d.~samples (the samples depend on the unknown parameter).}
Then, researchers provide polynomial time near-optimal algorithms~\cite{Diakonikolas201focs-robustEstimation, Liu19arxiv-TrimmedHardThresholding}.
 
In scenarios where one aims to estimate an unknown parameter given noisy, and possibly outlying, measurements, then researchers focus on outlier-robust reformulations of eq.~\eqref{eq:nls}.  For example, Rousseeuw~\cite{Rousseeuw11dmkd} proposed a celebrated algorithm to solve a dual reformulation of \MTS, that aims to minimize the residual errors of the remaining measurements given a maximum number of measurement rejections.  The algorithm assumes the outliers' number known, and as such, requires parameter tuning. Similar celebrated algorithms, that also assume the outliers' number known,  are the forward greedy by Nemhauser \etal~\cite{nemhauser78analysis}, and forward-backward greedy by Zhang~\cite{Zhang11tit-ForwardBackGreedy} 
(notably, both algorithms have quadratic run-time, a typically prohibitive run-time for robotics and computer vision applications, such as 3D registration, two-view, and SLAM).
In contrast to~\cite{Rousseeuw11dmkd,nemhauser78analysis,Zhang11tit-ForwardBackGreedy}, the greedy-like algorithm proposed in~\cite{Liu2018tsp-GreedyRobustRegression} considers the outliers' number unknown.  However, it still requires parameter tuning, this time for an outlier-free bound parameter.

Outlier-robust estimation in control takes typically the form of secure state estimation in the presence of outliers (caused by sensor malfunctions, or measurement attacks)~\cite{Pasqualetti13tac-attackDetection,Mishra2017tac-secureStateEstimation,Aghapour2018cdc-outlierAccomodation,Luo19arxiv-SecureStateEst}.  In~\cite{Pasqualetti13tac-attackDetection,Mishra2017tac-secureStateEstimation,Aghapour2018cdc-outlierAccomodation,Luo19arxiv-SecureStateEst}, the researchers propose optimal algorithms, that achieve exact state estimation when the non-outlying measurements are noiseless.  However, the algorithms \mbox{have exponential run-time.}

\subsection{Sub-optimality guarantees in outlier-robust estimation}\label{subsec:relWork_guarantees}

Additionally to the aforementioned algorithms that offer sub-optimality gurantees (or certificates of optimality), researchers have also provided conditions for exact estimation when some of the measurements are outliers, while the rest are noiseless~\cite{Candes05tit,Zhang11tit-ForwardBackGreedy}. However, the conditions' evaluation is NP-hard.  Additionally, the conditions are restricted on a linear and convex framework, where the measurements are linear in the unknown parameter $x$, and $x \in \mathbb{R}^n$.\footnote{In contrast, Section~\ref{sec:guarantees} provides a posteriori sub-optimality bounds, computable in $O(1)$ run-time, and applicable to even non-convex and non-linear frameworks, such as for 3D registration, two-view geometry, and SLAM.}
\section{Conclusion} 
\label{sec:conclusion}

We proposed a \emph{minimally trimmed squares (\MTS)} formulation to estimate an unknown variable
 from measurements plagued with outliers. 
We proved that the resulting outlier rejection problem is inapproximable: one cannot compute even an approximate solution  in quasi-polynomial time. We derived theoretical performance bounds: while polynomial-time algorithms may perform poorly in the worst-case, the bounds allow assessing the algorithms' post-run performance on any given problem instances (which are typically more favorable than the worst-case). Finally, we proposed a linear-time, general-purpose algorithm for outlier rejection, and showed that it outperforms several specialized methods across three spatial perception problems (3D registration, two-view geometry, SLAM).    
This work paves the way for several research avenues. 
While we focused on a non-linear least squares cost function, many of our conclusions extend 
to other norms, and robust costs. 
 We also plan to explore applications of the proposed bounds to other algorithms, including \ransac, and to other perception problems.

\appendices
\section{Proof of Theorem~\ref{th:hardness}} 
\label{app:hardness}

Here, we show the inapproximability of \MTS by reducing it to the {\em variable selection} problem, which we define next. 

\begin{problem}[Variable Selection]\label{pr:selection}
Let $U \in \mathbb{R}^{m \times l}$, $z \in \mathbb{R}^m$, and let $\Delta$ be a non-negative number. The variable selection problem asks to pick $d \in \mathbb{R}^l$ that is an optimal solution to the following optimization problem:
\belowdisplayskip=-6pt\begin{equation*}
	\min_{d \in \mathbb{R}^l}\;\; \|d\|_0,\;\text{s.t.}\;\; \|Ud-z\|_{2}\leq \Delta.
\end{equation*}
\hfill $\lrcorner$
\end{problem}

Variable selection is inapproximable in quasi-polynomial time.  We summarise the result in Lemm~\ref{th:inapprox_selection} below.  To this end, we first review basic definitions from complexity theory.

\begin{definition}[Big O notation]\label{def:bigO} Let $\mathbb{N}_+$ be the set of non-negative natural numbers, and consider two functions $h:\mathbb{N}_+\mapsto \mathbb{R}$ and $g:\mathbb{N}_+\mapsto \mathbb{R}$. The \emph{big O notation} in the equality $h(n)=O(g(n))$ means there exists some constant $c>0$ such that for all large enough $n$, 
$h(n)\leq c g(n)$. \hfill $\lrcorner$
\end{definition}

That is, $O(g(n))$ denotes the collection of functions $h$ that are bounded asymptotically by $g$, up to a constant factor.

\begin{definition}[Big $\Omega$ notation]\label{def:bigOmega} Consider two functions $h:\mathbb{N}_+\mapsto \mathbb{R}$ and $g:\mathbb{N}_+\mapsto \mathbb{R}$. The \emph{big $\Omega$ notation} in the equality $h(n)=\Omega(g(n))$ means there exists some constant $c>0$ such that for all large enough $n$, $h(n)\geq c g(n)$. \hfill $\lrcorner$
\end{definition}

That is, $\Omega(g(n))$ denotes the collection of functions $h$ that are lower bounded asymptotically  by $g$, up to a constant.

\begin{lemma}[\hspace{-0.15mm}{\cite[{Proposition 6}]{Foster15colt-variableSelectionHard}}]\label{th:inapprox_selection}
	For each $\delta \in (0,1)$, unless it is ${\rm NP} {\in}{\rm BPTIME(\textit{l}^{\rm poly \log \textit{l}})}$,  there exist:
	\begin{itemize} \setlength\itemsep{0.09em}
		\item {a function $q_1(l)$ which is in {$2^{\Omega(\log^{1-\delta} l)}$}; }
		\item {a polynomial $p_1(l)$ which is in $O(l)$;}\footnote{{In this context, a function with a fractional exponent is considered to be a polynomial, e.g., $l^{1/5}$ is considered to be a polynomial in $l$.}}
		\item {a polynomial $\Delta(l)$;}
		\item {a polynomial $m(l)$,} \end{itemize} 
	{and a zero-one} $m(l) \times l$ matrix $U$ such that even if it is known that a solution to $ Ud=1_{m (l)}$ exists, no quasi-polynomial algorithm can distinguish between the next cases {for large~$l$}: 
	\begin{enumerate} \setlength\itemsep{0.09em}
		\item[$S_1$)] There exists a vector $d \in {\mathbb{R}^l}$ such that $Ud = \mathbf{1}_{m(l)}$ and $||d||_{0} \leq p_1(l)$. 
		\item[$S_2$)] For any vector $d \in \mathbb{R}^l$ such that $||Ud - \mathbf{1}_{m(l)}||_2^2 \leq \Delta(l)$, we have $||d||_0 \geq p_1(l) q_1(l)$.
	\end{enumerate}
\end{lemma}  

Unless ${\rm NP} {\in}{\rm BPTIME(\textit{l}^{\rm poly \log \textit{l}})}$, Theorem~2 says that variable selection is inapproximable even in quasi-polynomial time. This is in the sense that  for large $l$ there is no quasi-polynomial algorithm that can distinguish between the two mutually exclusive statements $S_1$ and $S_2$. These statements are indeed mutually exclusive for large $l$, since then $q_1(l) > 1$, since it is $q_1(l) = 2^{ \Omega \left( \log^{1-\delta} l \right)}$.

From the inapproximability of variable selection, we can now infer the inapproximability for the following problem:
\begin{equation}\label{pr:aux_1}
\min_{d \in \mathbb{R}^l}\;\; \|d\|_0,\;\text{s.t.}\;\; Ud=1_{m(l)}.
\end{equation} 

\textit{Proof that problem in eq.~\eqref{pr:aux_1} is inapproximable}
Indeed, it suffices to set $\Delta=0$ in the definition of the variable selection problem, and then consider Lemma~\ref{th:inapprox_selection}.
\hfill $\blacksquare$

Next, from the inapproximability of the problem in eq.~\eqref{pr:aux_1}, we next infer the inapproximability for the problem below:
\begin{equation}\label{pr:aux_2}
\min_{d \in \mathbb{R}^l\!,~x\in \mathbb{R}^n}\;\; \|d\|_0,\;\text{s.t.}\;\; y=Ax+d.
\end{equation} 
To this end, consider the instance of Lemma~\ref{th:inapprox_selection}, and let: $\Delta'(l)=m^2(l)\Delta(l)$; $y$ be any solution to $Uy=1_{m(l)}$ (per Lemma~\ref{th:inapprox_selection} we know there exists a solution to this equation); and $A$ be a matrix in $\mathbb{R}^{l\times n}$, where $n=l-\text{rank}(U)$,\footnote{By this construction, t is $l>n$. That is, $A$ is a tall matrix with more rows than columns.} such that the columns of $A$ span the null space of $U$ (hence, $A$ is such that $UA=0$).  This instance of the problem in eq.~\eqref{pr:aux_2} is constructed in polynomial time in $l$, since solving a system of equations (as well as finding eigenvectors that span the null space of a matrix) happens in polynomial time.

Given the above instance of the problem in eq.~\eqref{pr:aux_2}, we next prove that the following two statements are indistinguishable to prove that the problem is inapproximable:
\begin{enumerate}\setlength\itemsep{0.09em}
	\item[$S_1'$)] There exist vectors $d \in {\mathbb{R}^l}$ and $x \in {\mathbb{R}^n}$ such that $y=Ax+d$ and $||d||_{0} \leq p_1(l)$. 
	\item[$S_2'$)] For any vectors $d \in {\mathbb{R}^l}$ and $x \in {\mathbb{R}^n}$ such that $||y-Ax-d||^2 \leq \Delta'(l)$,\footnote{The norm in $S_2'$ (namely, the $||y-Ax-d||^2$) can be any norm that is polynomially close (in $l$) to $\ell_2$-norm, such as the $\ell_1$-norm.} we have $||d||_0 \geq p_1(l) q_1(l)$.
\end{enumerate}

\textit{Proof that $S_1'$ and $S_2'$ are indistinguishable:}
	We prove that whenever statements $S_1$ and $S_2$ in Theorem~\ref{th:inapprox_selection} are true, then also statements $S_1'$ and $S_2'$ above are true, respectively. That is, all true instances of $S_1$ and $S_2$ are mapped to true instances of $S_1'$ and $S_2'$. Then, since also the mapping is done in polynomial time, this implies that no algorithm can solve the problem in eq.~\eqref{pr:aux_2} in quasi-polynomial time and distinguish the cases $S_1'$ and $S_2'$, because that would contradict that $S_1$ and $S_2$ are indistinguishable.

	\paragraph{Proof that when $S_1$ is true then also $S_1'$ also is}  Since $Uy=UAx+Ud$ implies $1_{m(l)}=Ud$, when $S_1$ is true, then $S_1'$ is also true with $x$ being the unique solution of $Ax=y-d$ (it is unique since $A$ is full column rank).
	
	\paragraph{Proof that when $S_2$ is true then also $S_2'$ also is} By contradiction: Assume that there are vectors $d \in {\mathbb{R}^l}$ and $x \in {\mathbb{R}^n}$ such that $||y-Ax-d||^2 \leq \Delta'(l)$ and $||d||_0 < p_1(l) q_1(l)$. Without loss of generality, assume $\ell_1$ norm for $||y-Ax-d||^2$. Then, $||y-Ax-d||^2_1 \leq \Delta'(l)$ implies $||U||^2_1||y-Ax-d||^2_1 \leq ||U||^2_1\Delta'(l)$, which implies $||U(y-Ax-d)||^2_1 \leq ||U||^2_1\Delta'(l)$, or $||1_{m(l)}-Ud||^2 \leq ||U||^2\Delta'(l)$, or $||1_{m(l)}-Ud||^2 \leq m^2(l)\Delta'(l)$, where the last holds true because $U $ is a zero-one matrix. Finally, because by definition of $\Delta'(l)$ it is $m^2(l)\Delta'(l)=\Delta(l)$, we have  $||1_{m(l)}-Ud||^2 \leq \Delta(l)$.  Overall, we just proved that there exist $d$ such that $||1_{m(l)}-Ud||^2 \leq \Delta(l)$ and $||d||_0 < p_1(l) q_1(l)$, which contradicts $S_2$.
\hfill $\blacksquare$

The final step for the proof of Theorem~\ref{th:hardness} is to reduce \MTS to the problem in eq.~\eqref{pr:aux_2}.  To this end, per the statement of Theorem~\ref{th:hardness} we focus on the linear framework of Example~\ref{ex:linear_dec}.  Also, we use the following notation:
\begin{itemize}
	\item Let $y_{\grSet\setminus \selSet}$ be the vector-stack of measurements $y_i$ such that $i\in \grSet\setminus\selSet$, given a measurement rejection set $\selSet$.
	\item Similarly, let  $d_{\grSet\setminus \selSet}$ be the vector-stack of noises $d_i$ such that $i\in \grSet\setminus\selSet$.
	\item And similarly, let $A_{\selSet}$ be the matrix-stack of measurement row-vectors $a_i^\top$.
\end{itemize}

Given the above notation, we can now write \MTS per the framework of Example~\ref{ex:linear_dec} as follows:
\begin{equation}\label{pr:aux_3}
\min_{\selSet\in\grSet\!,~x\in \mathbb{R}^n}\;\; |\selSet|,\;\text{s.t.}\;\; \|y_{{\grSet}\setminus {\selSet}}-A_{{\grSet}\setminus {\selSet}}x\|^2\leq \epsilon.
\
\end{equation}

To prove now the inapproximability of the \MTS problem in eq.~\eqref{pr:aux_3}, consider an instance of Lemma~\ref{th:inapprox_selection}, along with a corresponding instance considered above for the inapproximability of the problem in~\eqref{pr:aux_2}.  Then, set $\epsilon=\Delta'(l)$ in eq.~\eqref{pr:aux_3}.
We prove that the following two statements are indistinguishable:
\begin{enumerate}\setlength\itemsep{0.09em}
	\item[$S_1''$)] There exist  ${\selSet} \in \{1,\ldots,l\}$ and $x \in {\mathbb{R}^n}$ such that $y_{{\grSet}\setminus {\selSet}}=A_{{\grSet}\setminus {\selSet}}x$ and $|{\selSet}| \leq p_1(l)$. 
	\item[$S_2''$)] For any ${\selSet} \in \{1,\ldots,l\}$ and $x \in {\mathbb{R}^n}$ such that $||y_{{\grSet}\setminus {\selSet}}-A_{{\grSet}\setminus {\selSet}}x||^2 \leq \Delta'(l)$,\footnote{The norm in $S_2''$ can be any norm that is polynomially close (in $l$) to $\ell_2$-norm, such as the $\ell_1$-norm.} we have $|{\selSet}| \geq p_1(l) q_1(l)$.
\end{enumerate}

\textit{Proof that $S_1''$ and $S_2''$ are indistinguishable}
	We prove that whenever statements $S_1'$ and $S_2'$ above are true, then also statements $S_1''$ and $S_2''$ above are true, respectively.
	\setcounter{paragraph}{0}
	\paragraph{Proof that when $S_1'$ is true then also $S_1''$ also is} Assume $S_1$ is true. Let ${\selSet}=\{\text{all } i \text{ such that } d_i\neq 0, i\in \{1,\ldots,l\}\}$. Then, $y_{{\grSet}\setminus {\selSet}}=A_{{\grSet}\setminus {\selSet}}x$ since $d_{{\grSet}\setminus {\selSet}}=0$, and $|{\selSet}|=||d||_0\leq p_1(l)$.
	
	\paragraph{Proof that when $S_2'$ is true then also $S_2''$ also is} By contradiction: Assume there are ${\selSet} \in \{1,\ldots,l\}$ and $x \in {\mathbb{R}^n}$ such that $||y_{{\grSet}\setminus {\selSet}}-A_{{\grSet}\setminus {\selSet}}x||^2 \leq \Delta'(l)$ and $|{\selSet}| < p_1(l) q_1(l)$. Let $d_{{\grSet}\setminus {\selSet}}=0$, and $d_{\selSet}=y_{\selSet}-A_{\selSet}x$.  Then, $||d||_0=|S|< p_1(l) q_1(l)$ and $||y-Ax-d||^2=||y_{{\grSet}\setminus {\selSet}}-A_{{\grSet}\setminus {\selSet}}x||^2\leq \Delta'(l)$, which contradicts $S_2'$.
\hfill $\blacksquare$

Overall, we found instances for \MTS, per Example~\ref{ex:linear_dec}, and for a number $l$ of  measurements, where the statements $S_1''$ and $S_2''$ are indistinguishable even in quasi-polynomial time, and as a result, the proof of Theorem~\ref{th:hardness} is now complete.
\section{Proof of Theorem~\ref{th:guarantees}}
\label{sec:supp_guarantees}

First observe that $\res^\star_{|\selSet|}\leq \res(\selSet)$.  This holds true due to the definition of $\res^\star_{|\selSet|}$ as the smallest value of $\res(\cdot)$ among all sets with cardinality $|\selSet|$. Next, define the quantities:
\begin{itemize}
	\item $f(\selSet)\doteq \res(\emptyset)-\res(\selSet)$;
	\item $f^\star\doteq \res(\emptyset)-\res^\star_{|\selSet|}$.
\end{itemize}
We observe that:
\begin{align*}
f^\star&= \res(\emptyset)-\res^\star_{|\selSet|}\\
&\leq \res(\emptyset)\\
&= \res(\emptyset)+\res(\selSet)-\res(\selSet)\\
&=f(\selSet)+\res(\selSet).
\end{align*}
The above now imples:
\begin{align*}
\frac{f(\selSet)}{f^\star}&\geq 1- \frac{\res(\selSet)}{f^\star} \\
&\geq 1- \frac{\res(\selSet)}{f(\selSet)},
\end{align*}
where the latter holds since $f(\selSet)\leq f^\star$, which in turns holds because $\res(\selSet)\geq \res^\star$.  Finally, the above is:
\begin{align*}
\frac{f(\selSet)}{f^\star}&\geq 1- \apost_\selSet,
\end{align*}
which gives:
\begin{align*}
\res(\emptyset)-\res(\selSet)&\geq (1- \apost_\selSet)\res(\emptyset)- (1- \apost_\selSet)\res^\star_{|\selSet|} \Rightarrow\\
\apost_\selSet(\res(\emptyset)-\res^\star_{|\selSet|})&\geq \res(\selSet)-\res^\star_{|\selSet|}
\end{align*}
which gives eq.~\eqref{eq:guarantees}.  

To prove eq.~\eqref{eq:bound_refined}, we first observe that $\res^\star\geq \res^\star_{|\selSet|}$, since $|\selSet| \geq |\selSet^\star|$.  Now, it can be verified that:
\begin{equation}\label{app:eq:aux_guarantees}
\frac{\res(\selSet)-\res^\star}{\res(\emptyset)-\res^\star}\leq \frac{\res(\selSet)-\res^\star_{|\selSet|}}{\res(\emptyset)-\res^\star_{|\selSet|}},
\end{equation}
since it also is $\res(\emptyset)\geq \res(\selSet)$.
Substituting eq.~\eqref{app:eq:aux_guarantees} in eq.~\eqref{eq:guarantees}, the proof of the theorem is now complete.
\section{Supplemental for Experiments and Applications}\label{sec:supp_exp}

We first provide \name's selected input parameters for each experiment, along with the methodology we applied to select them.  Then, we list all missing plots from the experimental section of the paper (Experiments and Applications).

For our experiments, we selected the input parameters for \name by testing the algorithm on each dataset on scenarios where we added 20\% of outliers to the dataset.  We did not add outliers to the EuRoc dataset.  instead, we used the parameters selected for the synthetic two-view dataset.   In particular, the parameters $g$ and $\gamma$ were chosen to make the algorithm terminate with a few iterations (e.g., around 20 iterations for SLAM) ---we recall that $g$ (outlier group size) and $\gamma$ (outlier-threshold discount factor) control the maximum number of rejected measurements at each iteration.  The parameter $\delta$ (convergence threshold) was chosen so the algorithm rejects an accurate number of measurements  upon termination.  Specifically, 20\% of measurements, since 20\% was the percentage of the added outliers by us to the datatset.  We kept $T$ (number of iterations to decide convergence) fixed to its default value 2 for all the experiments.  Overall, the selected parameters for each experiment are shown in Table~\ref{tab:parameters}. Finally, \name requires the minimum number of measurements required by global solver.  This number for registration is 3; for two-view is 5; and for SLAM is 0.

\begin{table}[h!]
\centering
\caption{Parameters for \name.}
\begin{tabular}{|l|cccc|}
\hline
{\bf Experiment} & $T$ & $\delta$ & $g$   & $\gamma$  \\
\hline \\[-1em]
Bunny       & 2 & \num{1e-4}  & 10  & 0.99   \\
ETH         & 2 & \num{1e-2}  & 10  & 0.99   \\
\hline \\[-1em]
MIT         & 2 & 10    & 20  & 0.99   \\
Intel       & 2 & 60    & 20  & 0.5    \\
CSAIL       & 2 & 60    & 20  & 0.5    \\
Grid        & 2 & 20    & 20  & 0.99   \\
Sphere 2500 & 2 & 10    & 200 & 0.9    \\
\hline \\[-1em]
Two-view (synthetic) & 2 & \num{1.5e-4} & 20 & 0.99  \\
EuRoC &  2 & \num{1.5e-4} & 20 & 0.99 \\
\hline
\end{tabular}
\label{tab:parameters}
\end{table}

We list all missing plots from the experimental section of the paper in the following pages. 

\newpage
\clearpage
\subsection{Supplemental for  Robust Registration}
\label{sec:supp_registration}

\begin{figure}[h!]
	\begin{center}
	\begin{minipage}{\textwidth}
	\hspace{-0.5cm}
	\begin{tabular}{c}%
			\begin{minipage}{.5\columnwidth}%
			\centering%
			\includegraphics[width=1.0\columnwidth]{bunny-Error-Rotation.pdf} \\
      (a) Rotation Error 
			\end{minipage}
\\
			\begin{minipage}{.5\columnwidth}%
			\centering%
			\includegraphics[width=1.0\columnwidth]{bunny-Error-Translation.pdf} \\
      (b) Translation Error 
			\end{minipage}
\\
			\begin{minipage}{.5\columnwidth}%
			\centering%
			\includegraphics[width=1.0\columnwidth]{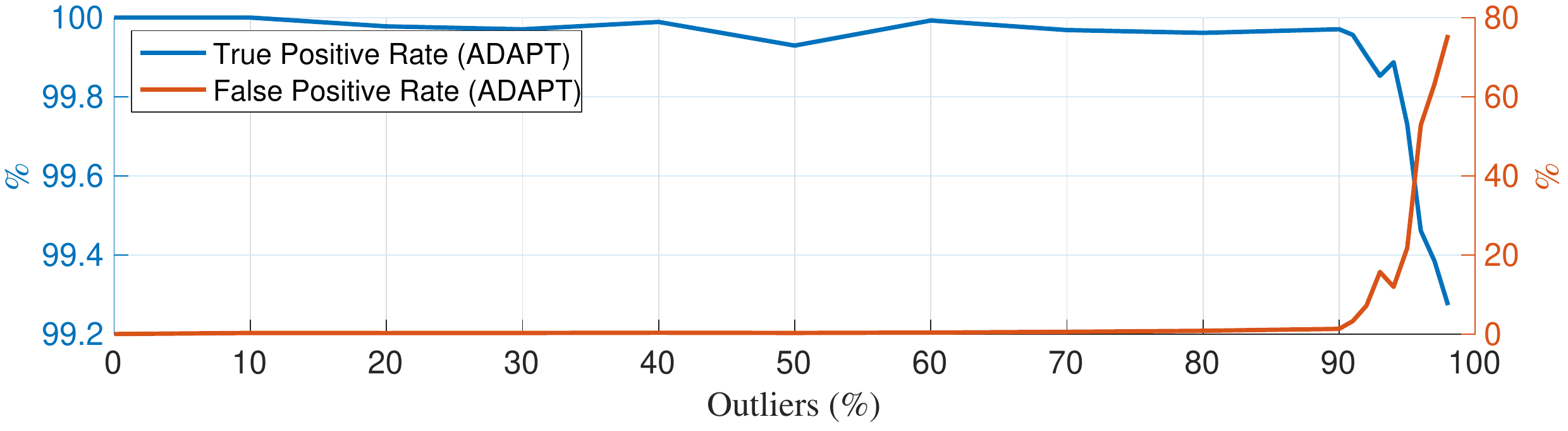} \\
      (c) True/False Positive Rate
			\end{minipage}
\\
			\begin{minipage}{.5\columnwidth}%
			\centering%
			\includegraphics[width=1.0\columnwidth]{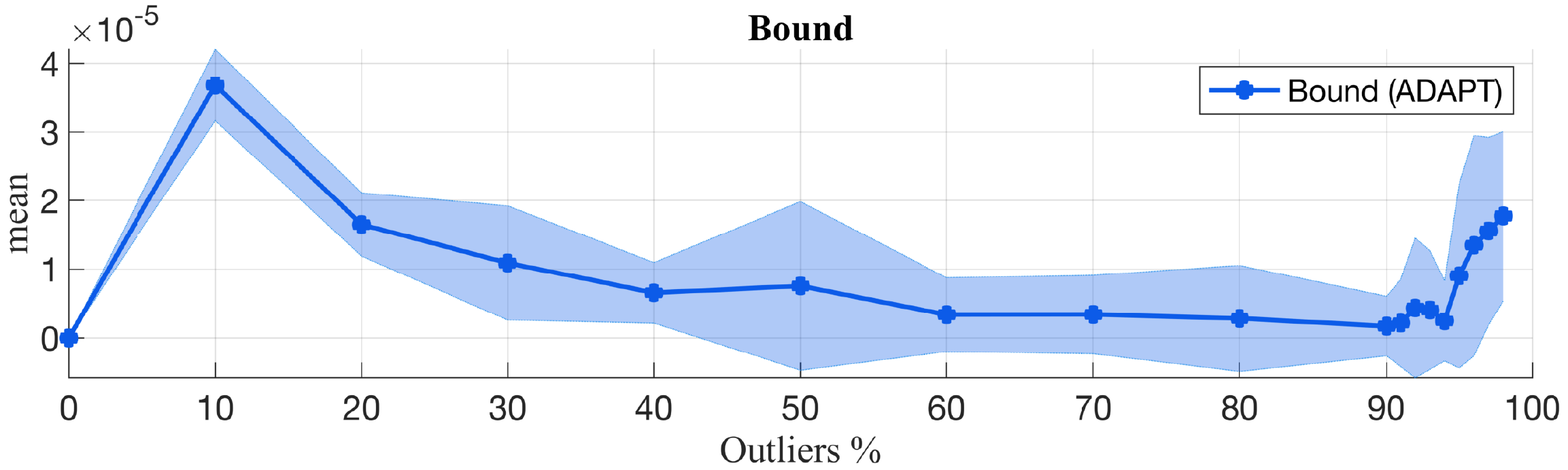} \\
			(d)  Bound
			\end{minipage}
\\
			\begin{minipage}{.5\columnwidth}%
			\centering%
			\includegraphics[width=1.0\columnwidth]{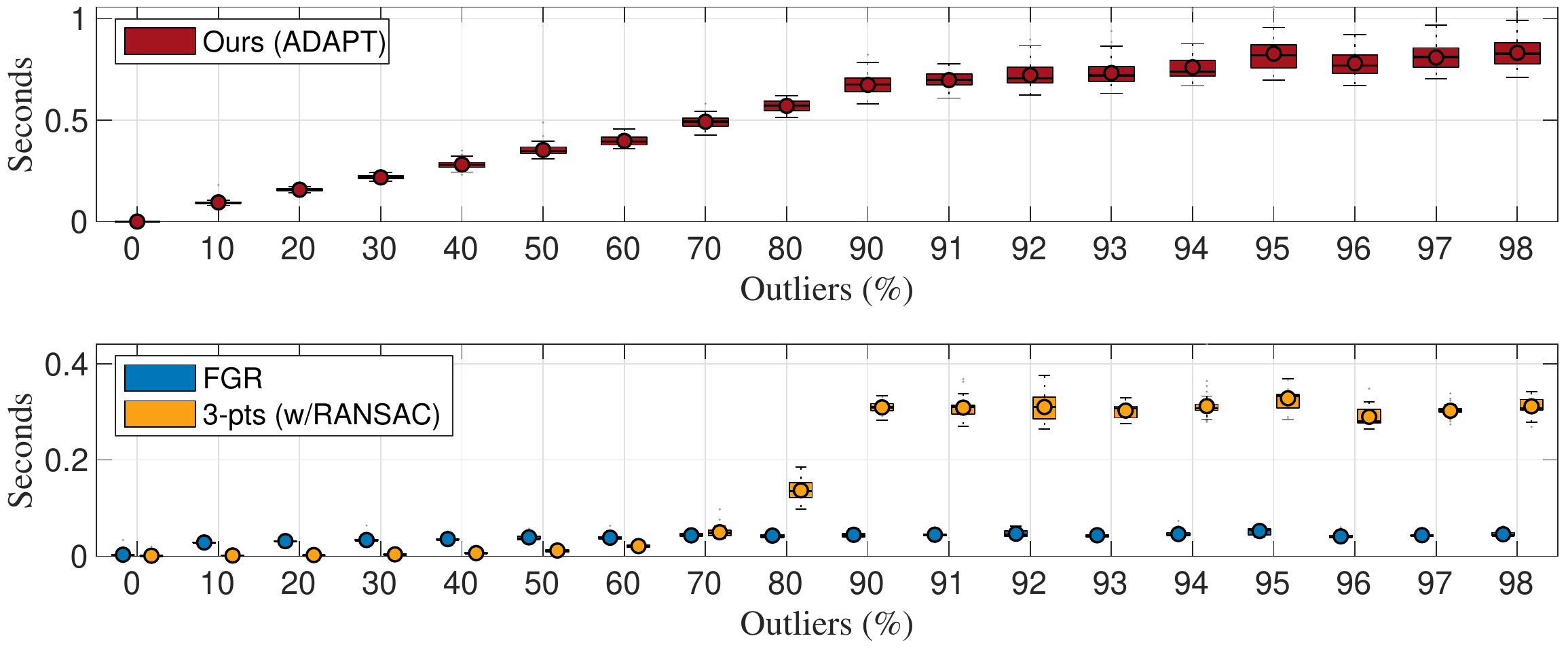} \\
			(e)  Running Time
			\end{minipage}
		\end{tabular}
	\end{minipage}
	\caption{\name  over Bunny dataset.}
	 \label{fig:supp_regBunny}
	\end{center}
\end{figure}

\begin{figure}[h!]
	\begin{center}
	\begin{minipage}{\textwidth}
  \vspace{-5cm}
	\hspace{-0.5cm}
	\begin{tabular}{c}%
			\begin{minipage}{.5\columnwidth}%
			\centering%
			\includegraphics[width=1.0\columnwidth]{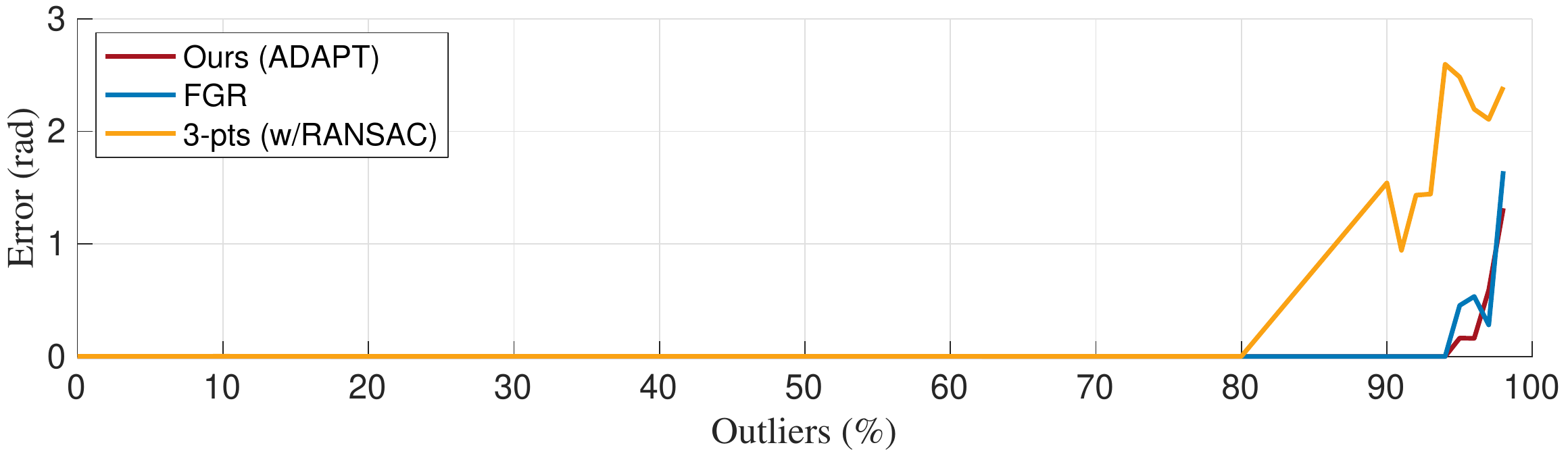} \\
      (a) Rotation Error 
			\end{minipage}
\\
			\begin{minipage}{.5\columnwidth}%
			\centering%
			\includegraphics[width=1.0\columnwidth]{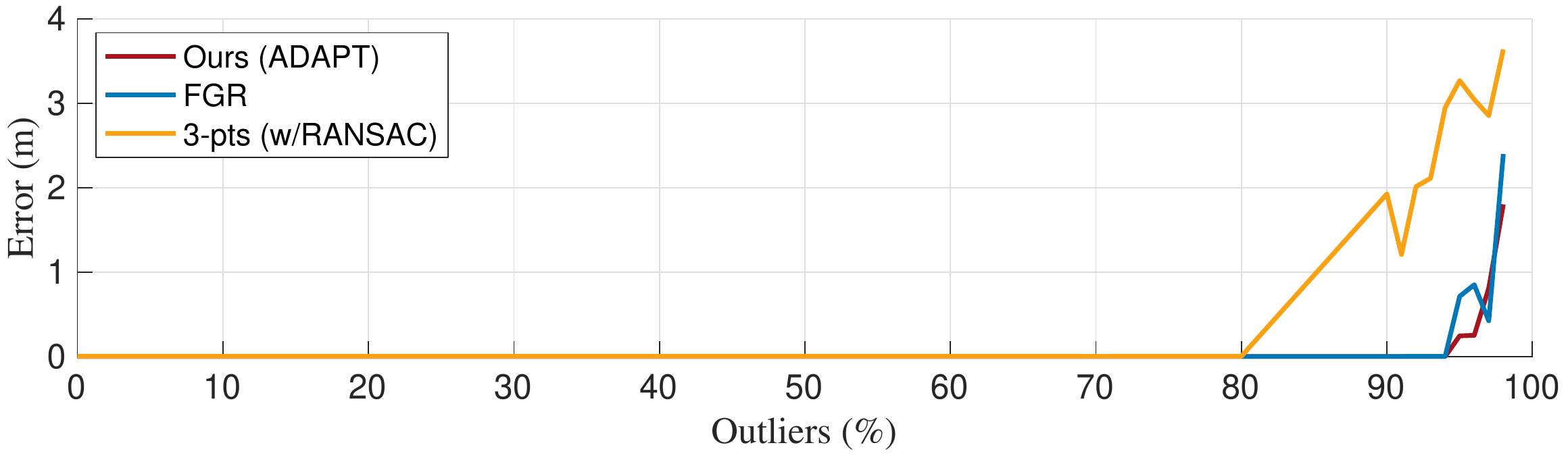} \\
      (a) Translation Error 
			\end{minipage}
\\
			\begin{minipage}{.5\columnwidth}%
			\centering%
			\includegraphics[width=1.0\columnwidth]{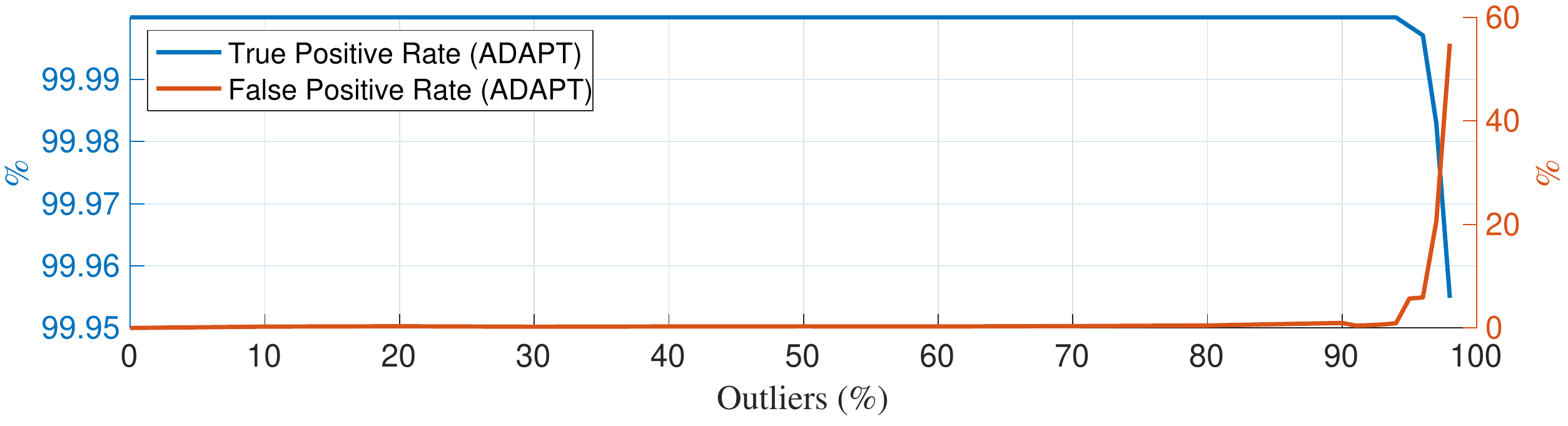} \\
      (a) True/False Positive Rate
			\end{minipage}
\\
			\begin{minipage}{.5\columnwidth}%
			\centering%
			\includegraphics[width=1.0\columnwidth]{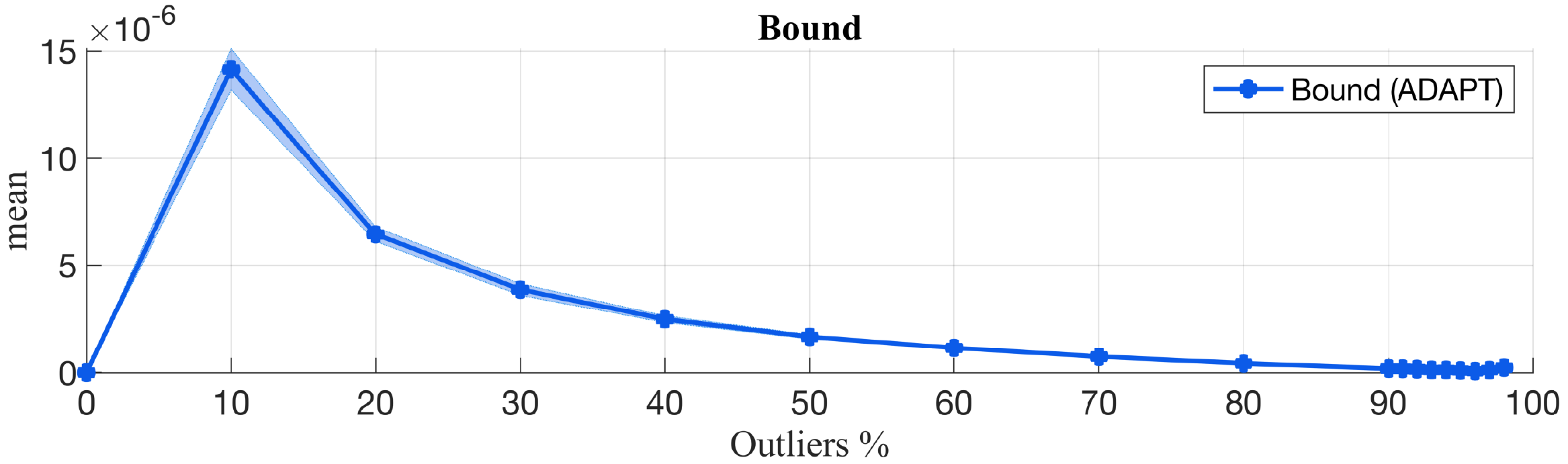} \\
			(b)  Bound
			\end{minipage}
\\
			\begin{minipage}{.5\columnwidth}%
			\centering%
			\includegraphics[width=1.0\columnwidth]{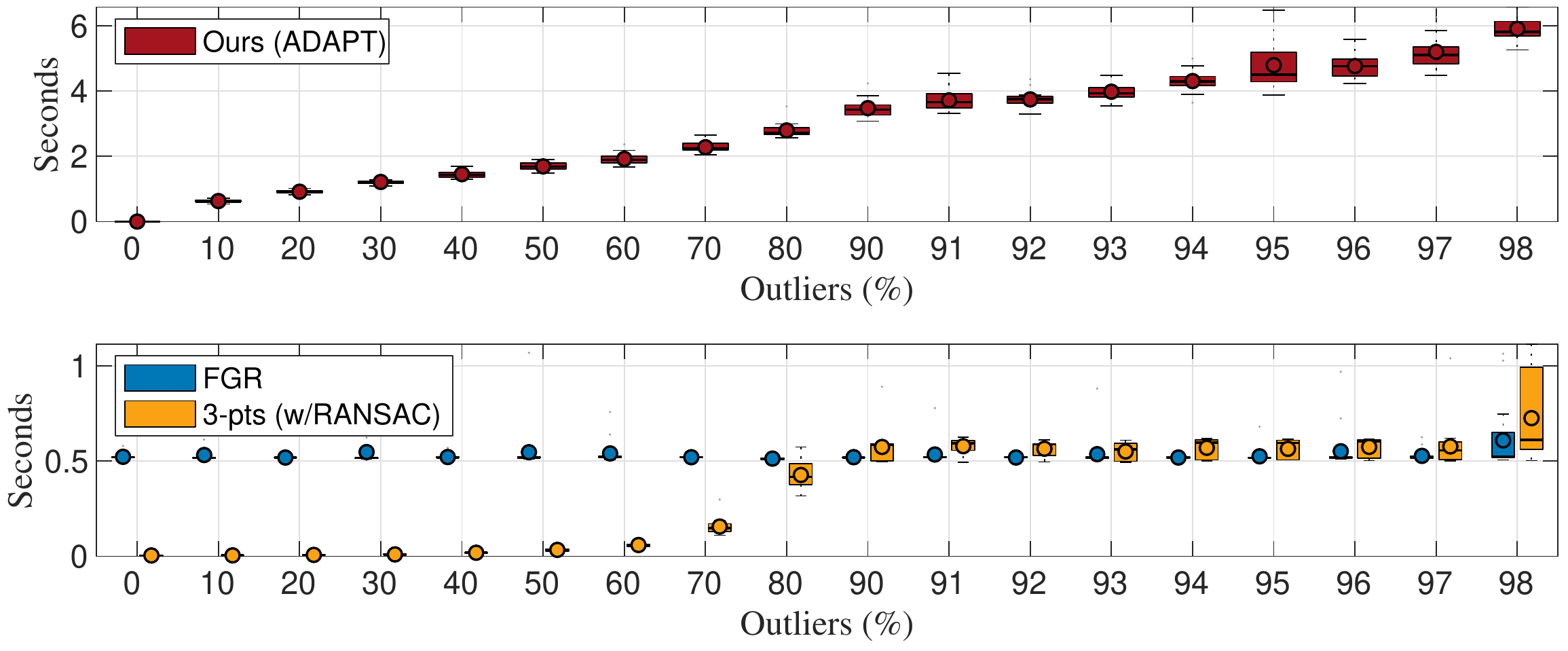} \\
			(c)  Running Time
			\end{minipage}
		\end{tabular}
	\end{minipage}
	\caption{\name  over ETH dataset.}
	 \label{fig:supp_regETH}
	\end{center}
\end{figure}

\newpage
\clearpage
\subsection{Supplemental for  Robust Two-view Geometry}
\label{sec:supp_view}

\begin{figure}[h!]
	\begin{center}
  \begin{minipage}{\textwidth}
	\hspace{-0.5cm}
	\begin{tabular}{c}%
      \begin{minipage}{.5\columnwidth}%
      \centering%
      \includegraphics[width=1.0\columnwidth]{two-view-Error-Rotation.pdf}  \\
      (a) Rotation Error 
      \end{minipage}
\\
      \begin{minipage}{.5\columnwidth}%
      \centering%
      \includegraphics[width=1.0\columnwidth]{two-view-Error-Translation.pdf} \\
      (b) Translation Error 
      \end{minipage}
\\
      \begin{minipage}{.5\columnwidth}%
			\centering%
			\includegraphics[width=1.0\columnwidth]{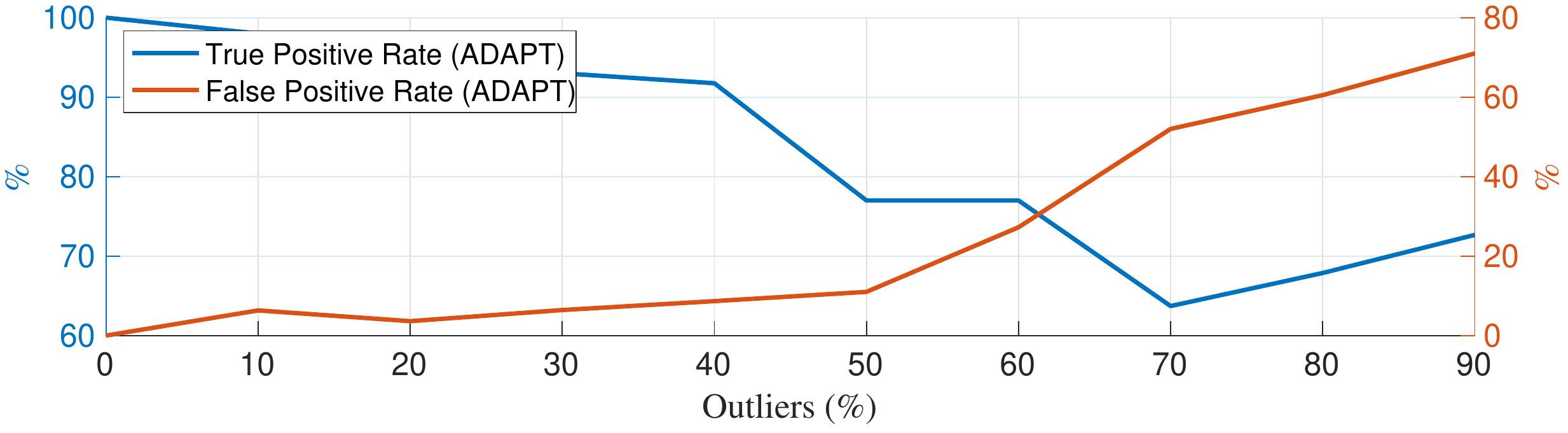} \\
      (c) True/False Positive Rate
			\end{minipage}
\\
			\begin{minipage}{.5\columnwidth}%
			\centering%
			\includegraphics[width=1.0\columnwidth]{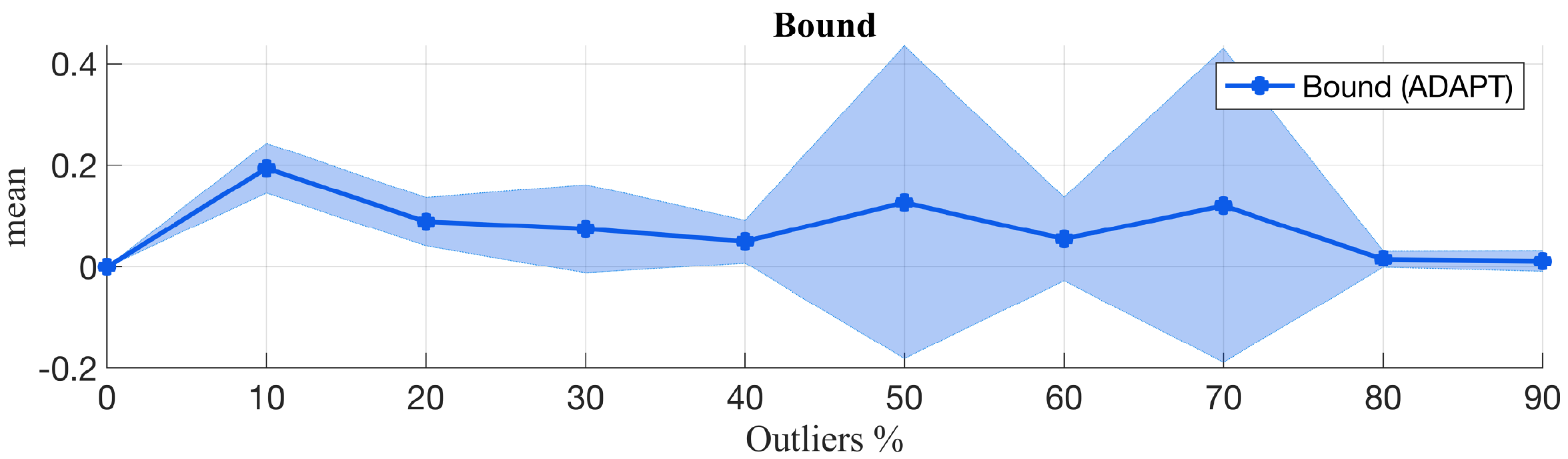} \\
			(d)  Bound
			\end{minipage}
\\
			\begin{minipage}{.5\columnwidth}%
			\centering%
			\includegraphics[width=1.0\columnwidth]{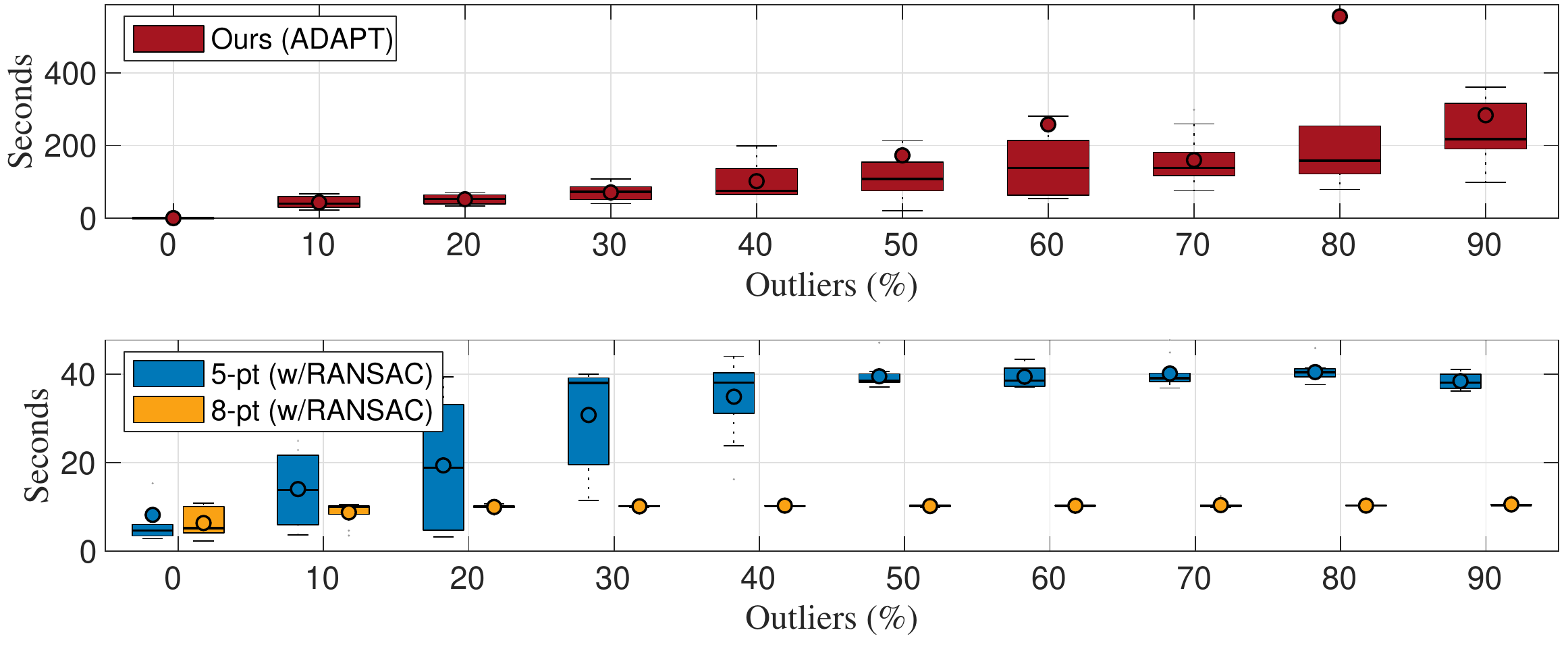} \\
			(e)  Running Time
			\end{minipage}
\\
			\begin{minipage}{.5\columnwidth}%
			\centering%
			\includegraphics[width=1.0\columnwidth]{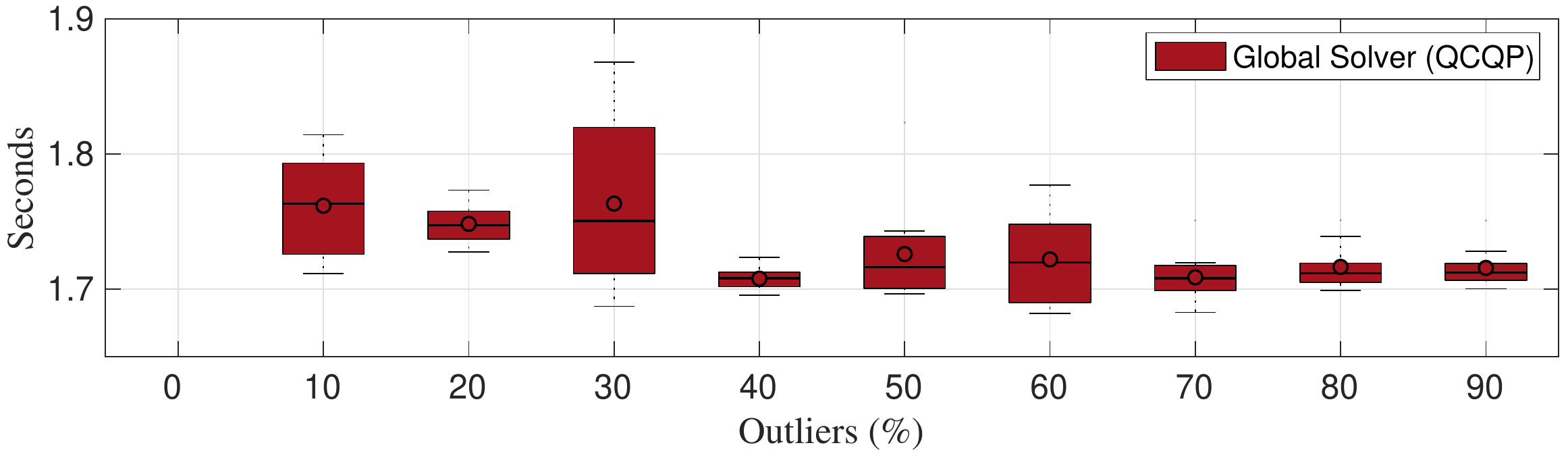} \\
			(f)  Global Solver's Running Time
			\end{minipage}
		\end{tabular}
	\end{minipage}
	\caption{\name  over 2-view synthetic dataset.}
	 \label{fig:supp_2-view_synthetic}
	\end{center}
\end{figure}

\newpage

\begin{figure}[h!]
	\begin{center}
    \begin{minipage}{\textwidth}
      \vspace{0.5cm}
			\hspace{-0.5cm}
			\begin{tabular}{cc}%
				\begin{minipage}{.5\columnwidth}%
					\centering%
					\includegraphics[width=1.0\columnwidth]{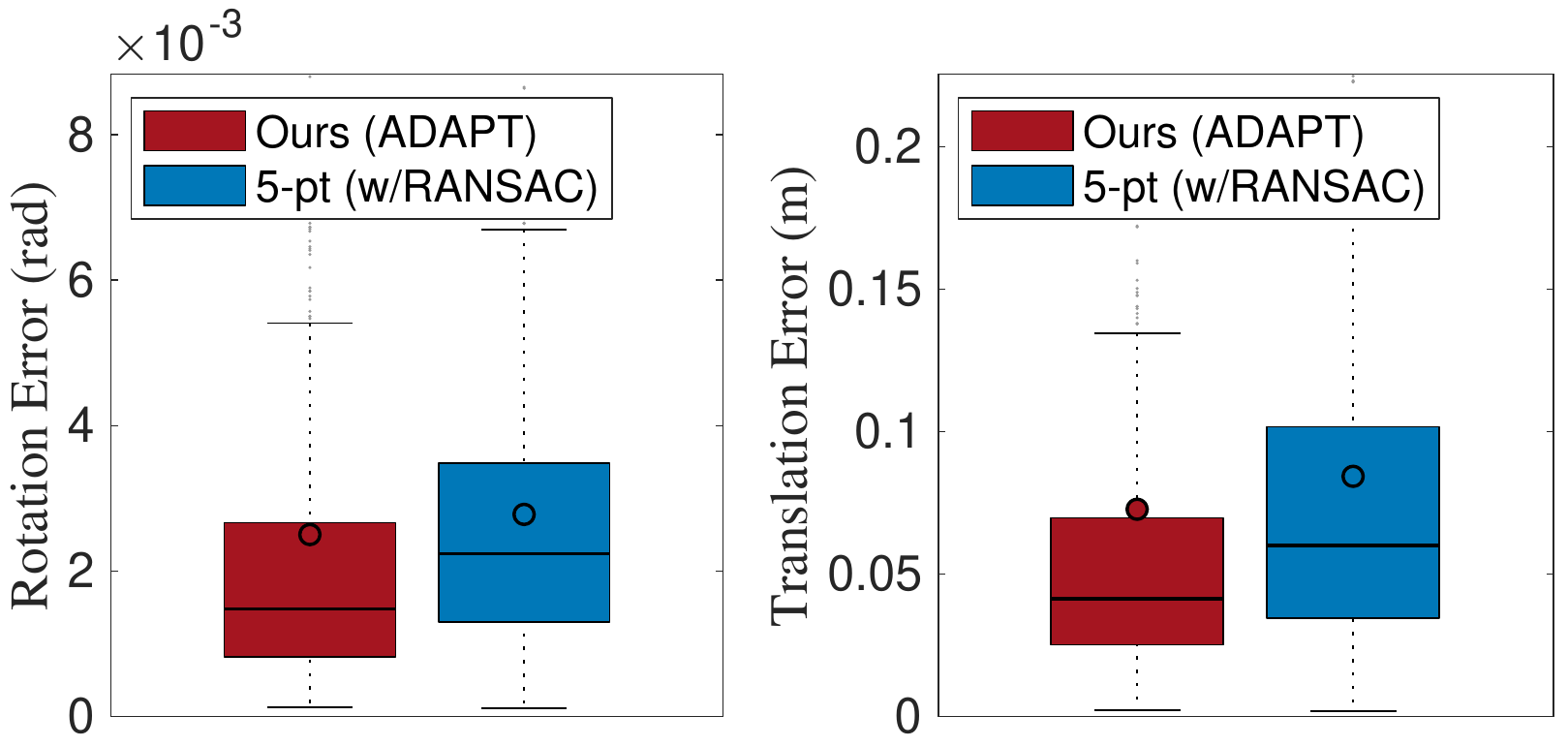} 
				\end{minipage}
			\end{tabular}
		\end{minipage}
		\caption{\nameFast rotation and translation error over EuRoC dataset.}
		\label{fig:supp_viewEuroc}
	\end{center}
\end{figure}

\newpage
\clearpage
\subsection{Supplemental for Robust SLAM}
\label{sec:supp_slam}


\begin{figure}[h!]
	\begin{center}
		\begin{minipage}{\textwidth}
			\hspace{-0.5cm}
			\begin{tabular}{c}%
				\begin{minipage}{.5\columnwidth}%
					\centering%
					\includegraphics[width=1.0\columnwidth]{mit-ATE.pdf}  \\
					(a) Average Trajectory Error
				\end{minipage}
				\\
				\begin{minipage}{.5\columnwidth}%
					\centering%
					\includegraphics[width=1.0\columnwidth]{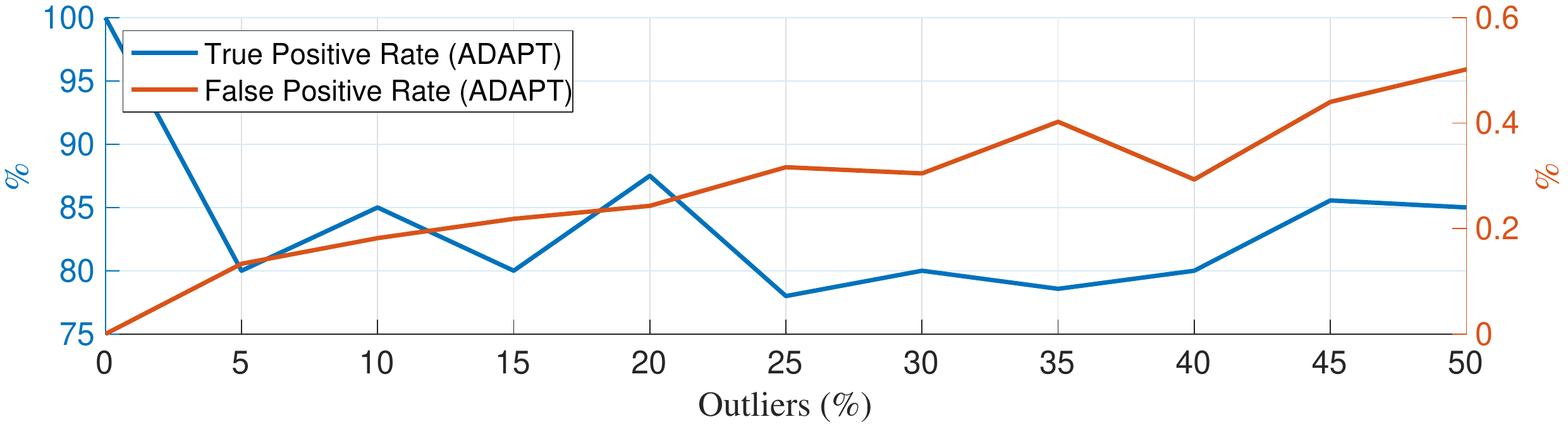}  \\
					(b) True/False Positive Rate
				\end{minipage}
				\\
				\begin{minipage}{.5\columnwidth}%
          \centering%
          \includegraphics[width=1.0\columnwidth]{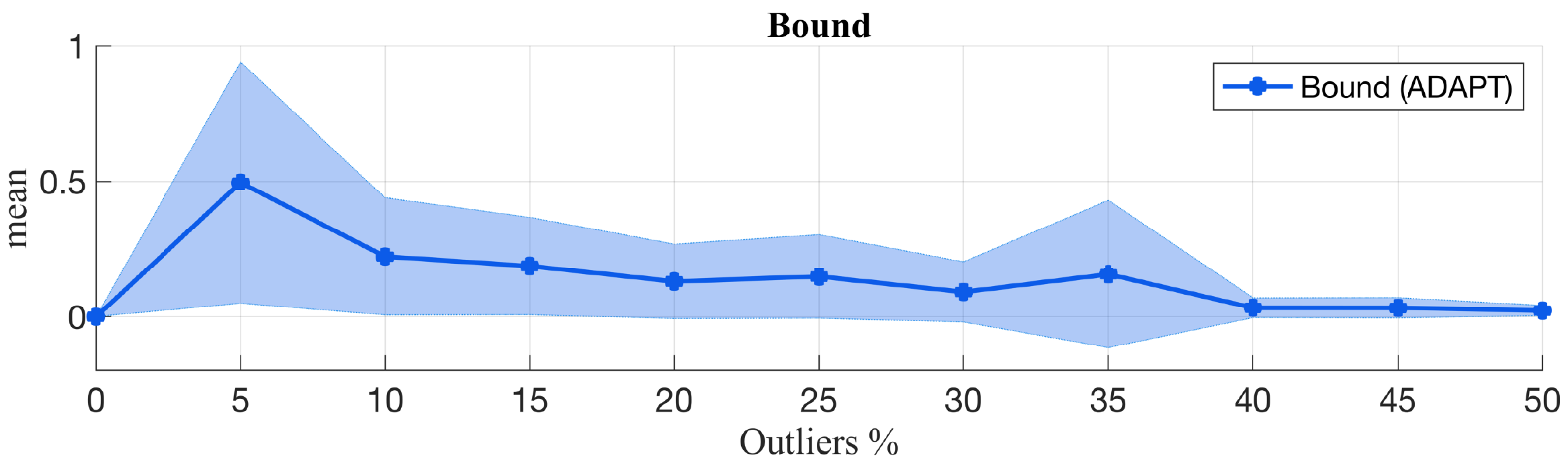} \\
          (c) Bound
        \end{minipage}
        \\
        \begin{minipage}{.5\columnwidth}%
          \centering%
          \includegraphics[width=1.0\columnwidth]{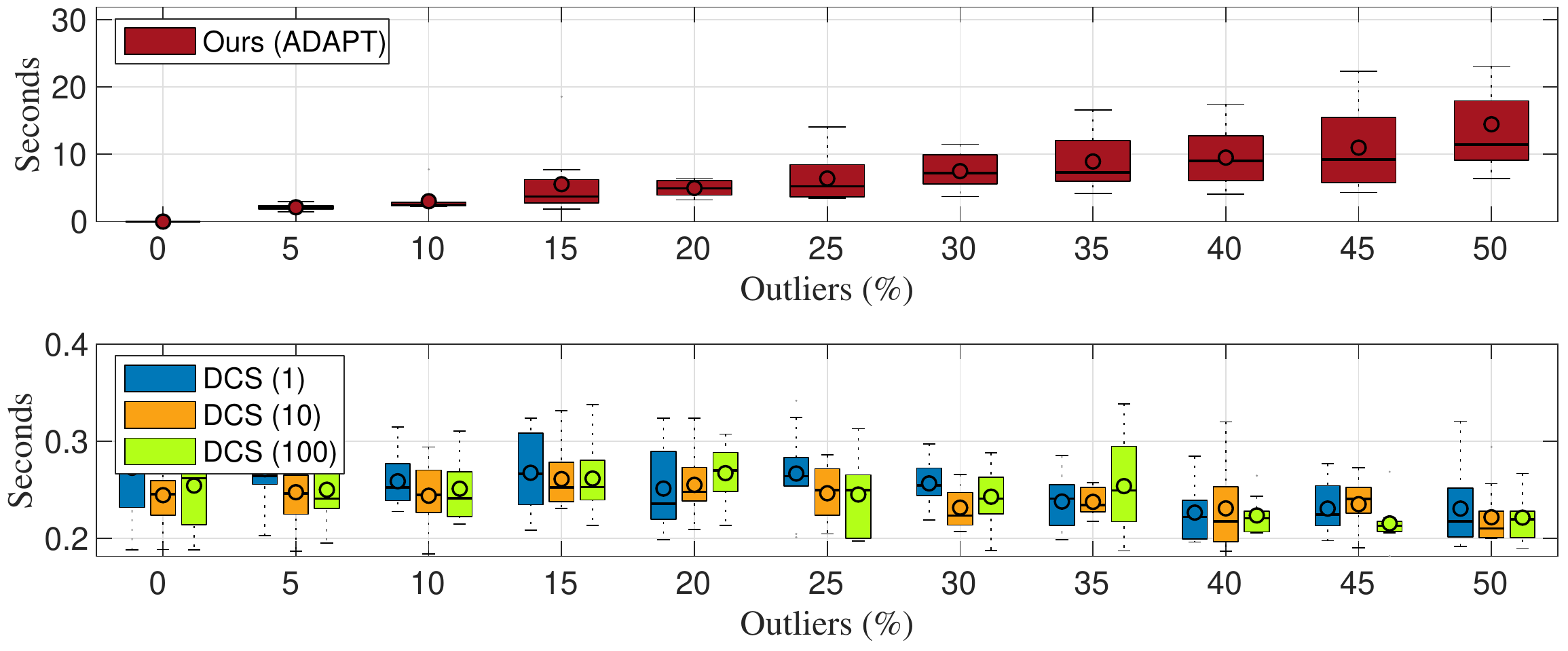} \\
          (d) Running time
        \end{minipage}
        \\
				\begin{minipage}{.5\columnwidth}%
          \centering%
          \includegraphics[width=1.0\columnwidth]{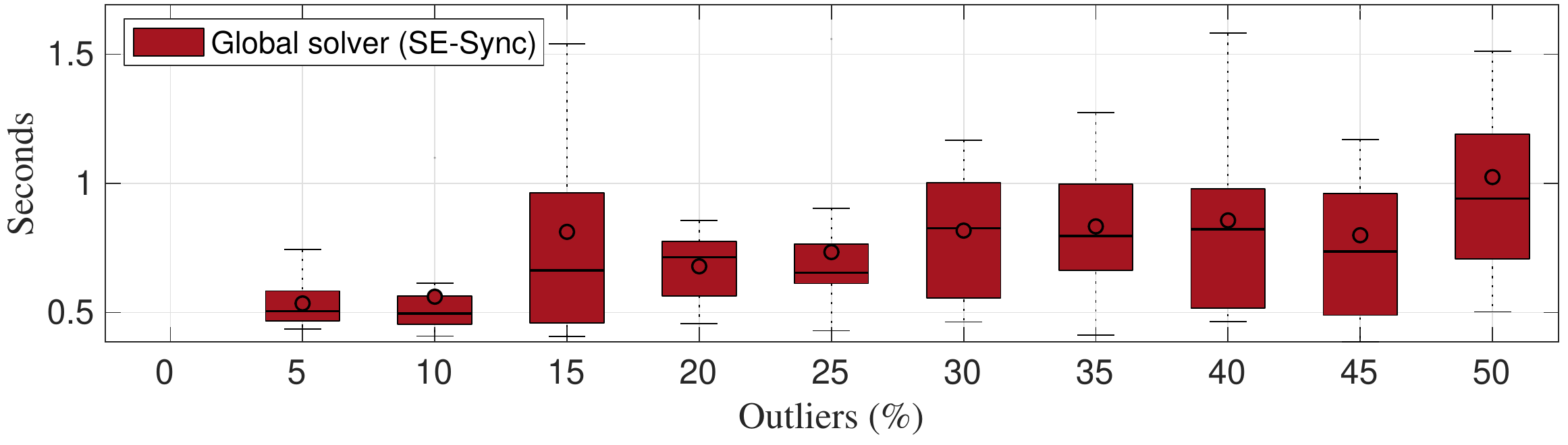} \\
          (e) Global Solver's Running Time
				\end{minipage}
			\end{tabular}
		\end{minipage}
		\caption{\name over MIT dataset.}
		\label{fig:supp_slamMIT}
	\end{center}
\end{figure}

\begin{figure}[h!]
	\begin{center}
    \begin{minipage}{\textwidth}
      \vspace{-4.9cm}
			\hspace{-0.5cm}
			\begin{tabular}{c}%
				\begin{minipage}{.5\columnwidth}%
					\centering%
					\includegraphics[width=1.0\columnwidth]{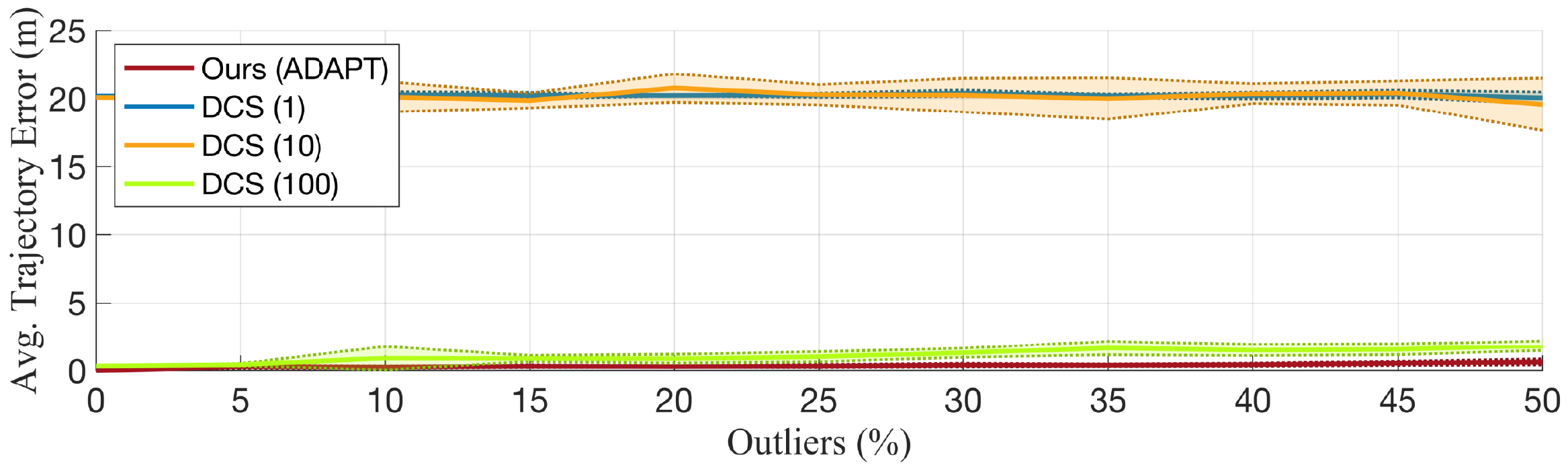}  \\
					(a) Average Trajectory Error
				\end{minipage}
				\\
				\begin{minipage}{.5\columnwidth}%
					\centering%
					\includegraphics[width=1.0\columnwidth]{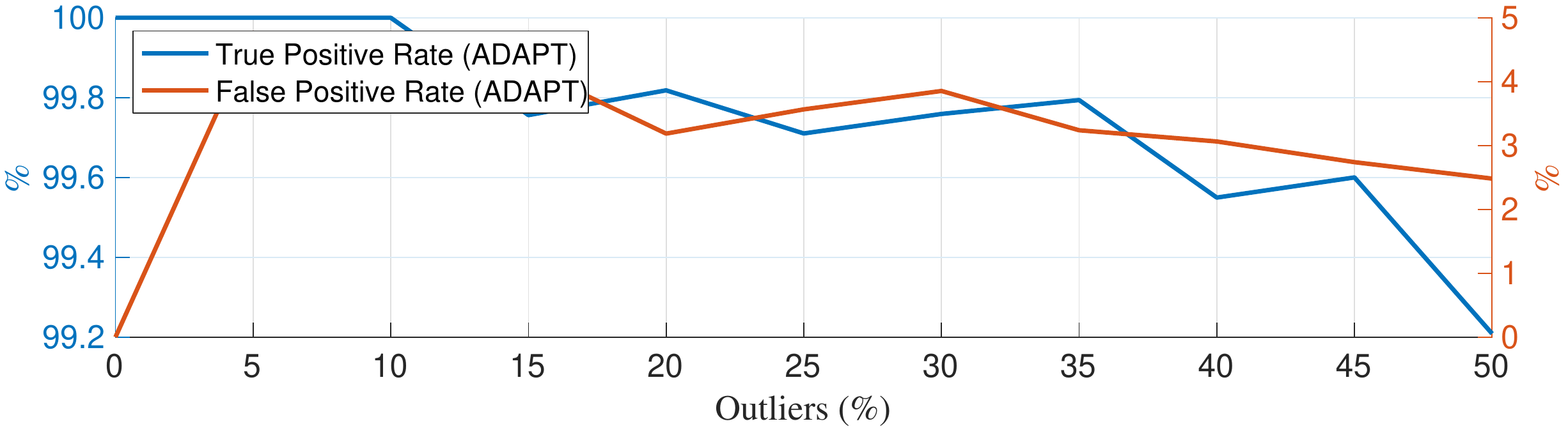}  \\
					(b) True/False Positive Rate
				\end{minipage}
				\\
				\begin{minipage}{.5\columnwidth}%
          \centering%
          \includegraphics[width=1.0\columnwidth]{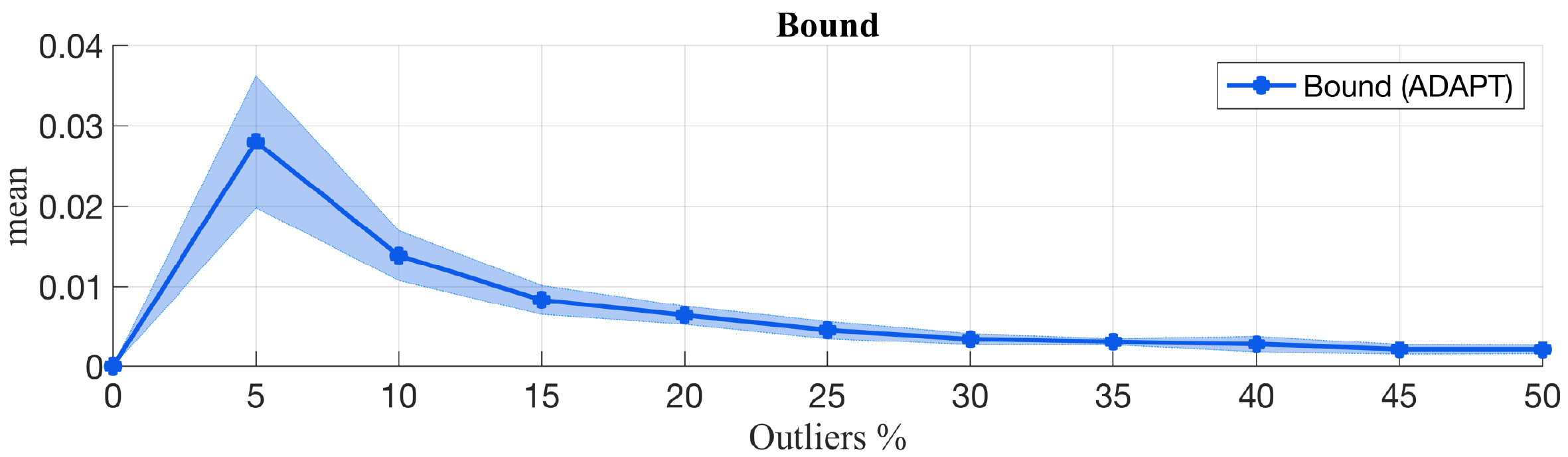} \\
          (c) Bound
        \end{minipage}
        \\
        \begin{minipage}{.5\columnwidth}%
          \centering%
          \includegraphics[width=1.0\columnwidth]{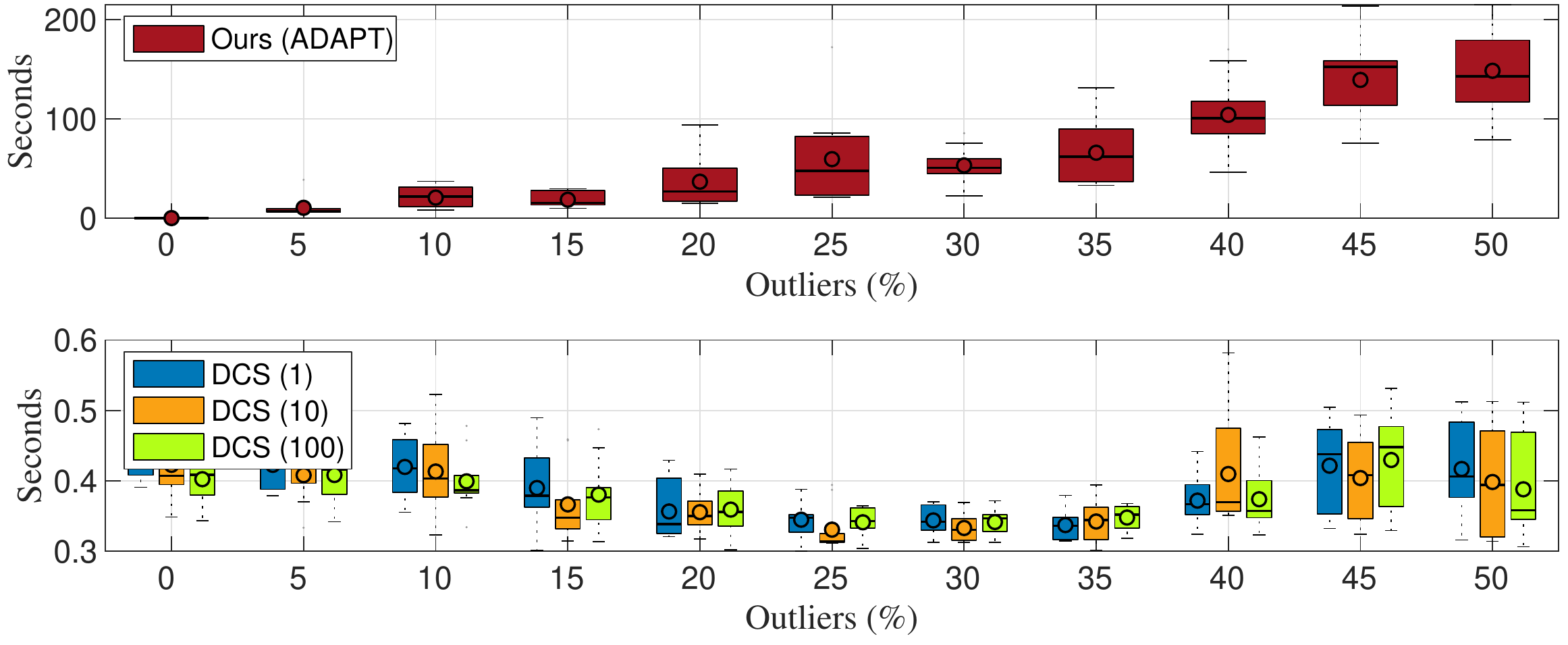} \\
          (d) Running time
        \end{minipage}
        \\
				\begin{minipage}{.5\columnwidth}%
          \centering%
          \includegraphics[width=1.0\columnwidth]{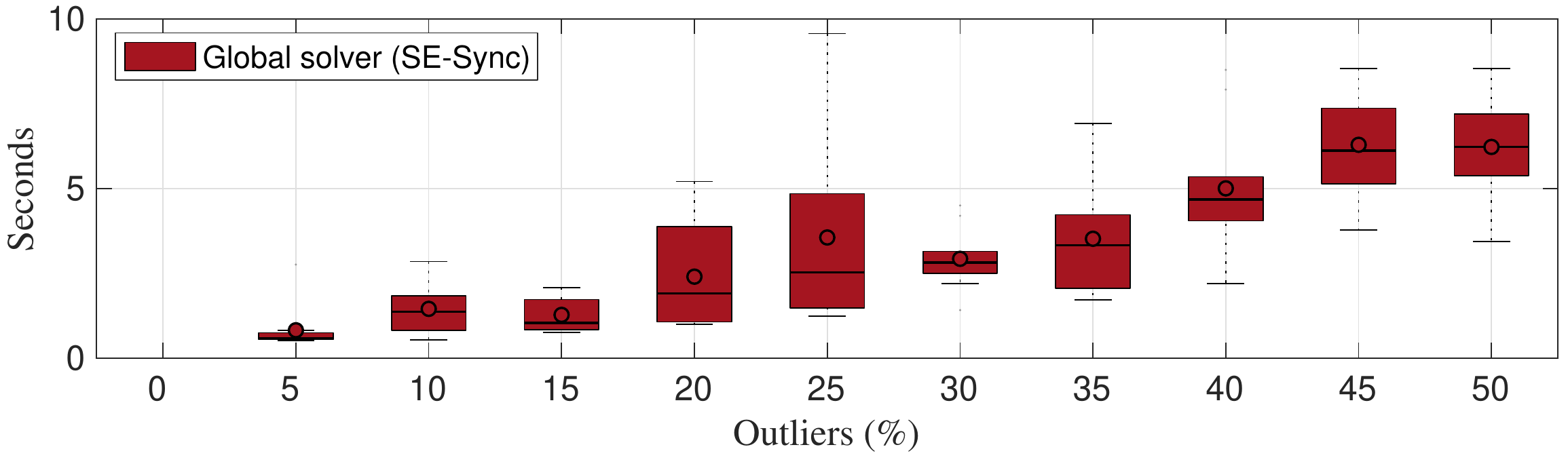} \\
          (e) Global Solver's Running Time
				\end{minipage}
			\end{tabular}
		\end{minipage}
		\caption{\name over Intel dataset.}
		\label{fig:supp_slamIntel}
	\end{center}
\end{figure}

\newpage
\clearpage

\begin{figure}[h!]
	\begin{center}
		\begin{minipage}{\textwidth}
			\hspace{-0.5cm}
			\begin{tabular}{c}%
				\begin{minipage}{.5\columnwidth}%
					\centering%
					\includegraphics[width=1.0\columnwidth]{csail-ATE.pdf}  \\
					(a) Average Trajectory Error
				\end{minipage}
				\\
				\begin{minipage}{.5\columnwidth}%
					\centering%
					\includegraphics[width=1.0\columnwidth]{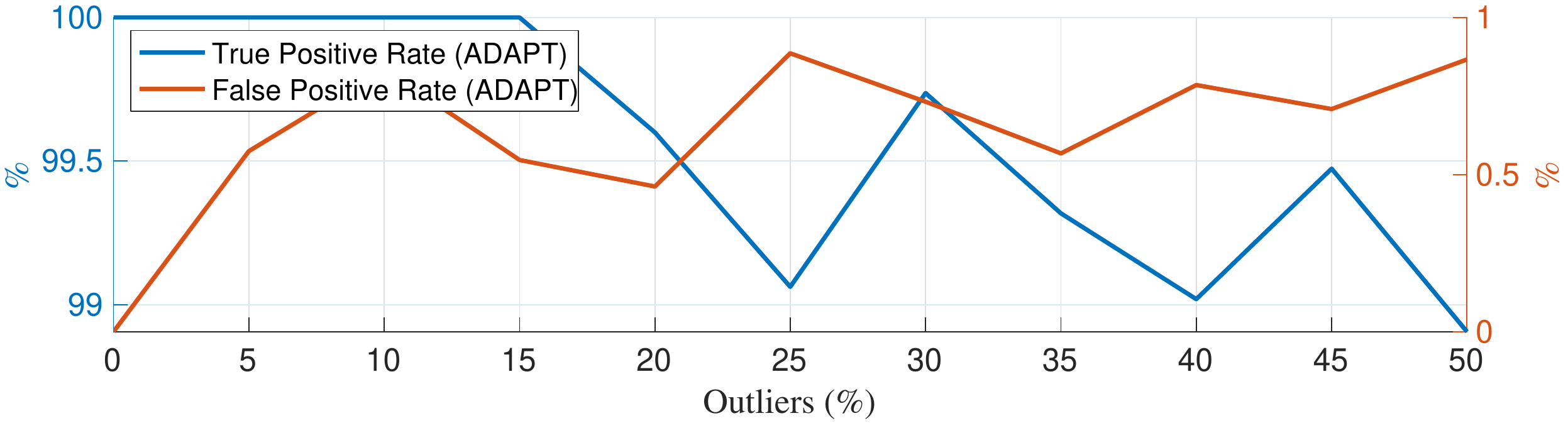}  \\
					(b) True/False Positive Rate
				\end{minipage}
				\\
				\begin{minipage}{.5\columnwidth}%
          \centering%
          \includegraphics[width=1.0\columnwidth]{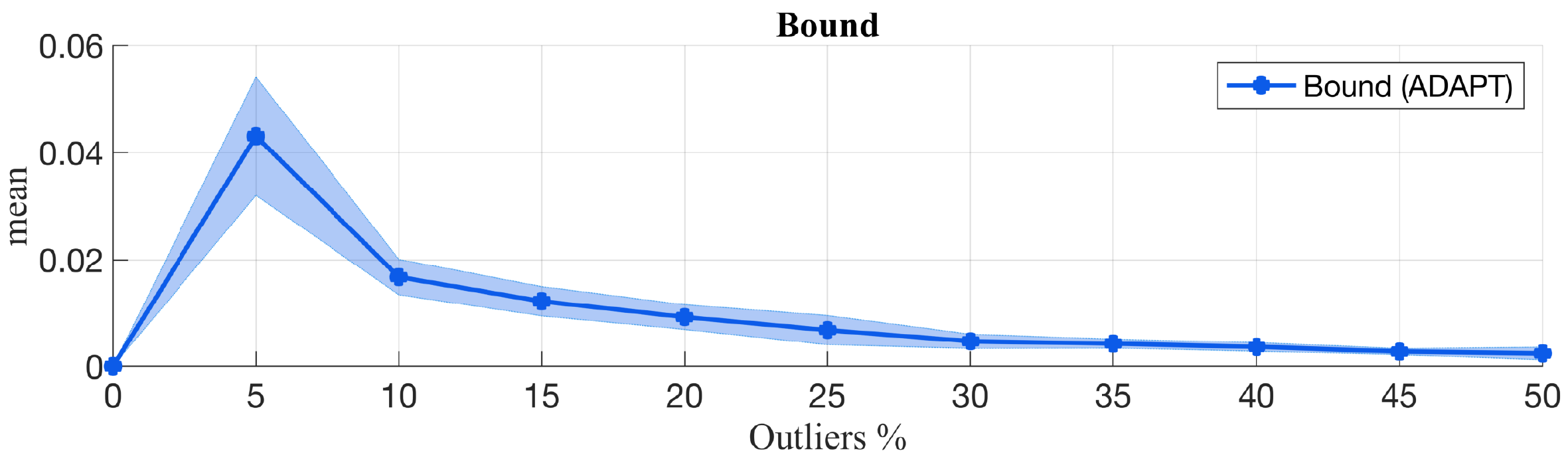} \\
          (c) Bound
        \end{minipage}
        \\
        \begin{minipage}{.5\columnwidth}%
          \centering%
          \includegraphics[width=1.0\columnwidth]{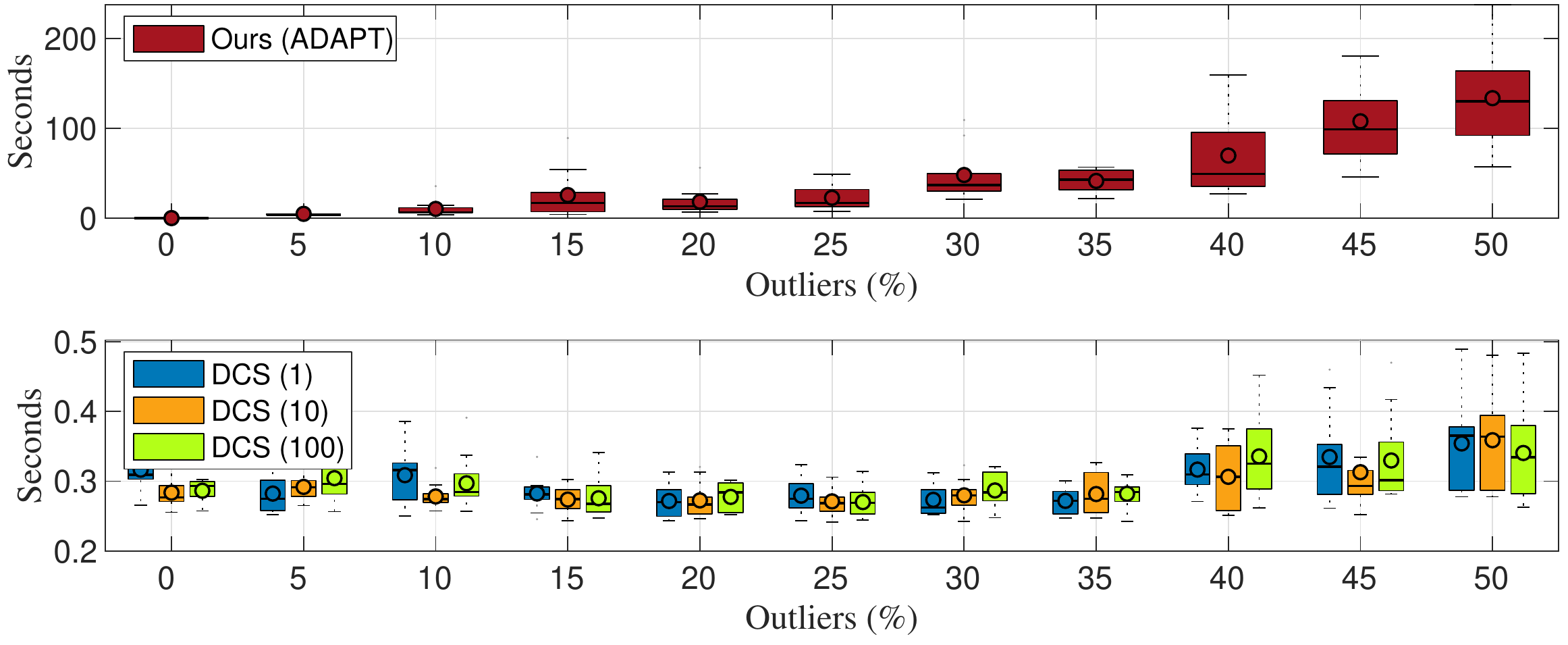} \\
          (d) Running time
        \end{minipage}
        \\
				\begin{minipage}{.5\columnwidth}%
          \centering%
          \includegraphics[width=1.0\columnwidth]{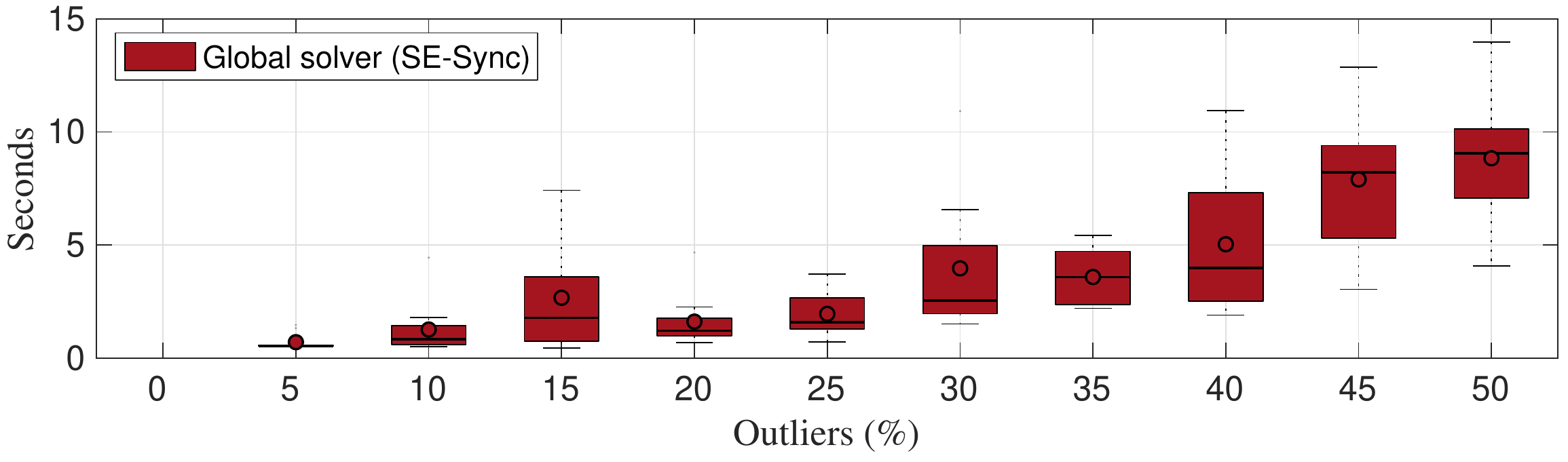} \\
          (e) Global Solver's Running Time
				\end{minipage}
			\end{tabular}
		\end{minipage}
		\caption{\name over CSAIL dataset.}
		\label{fig:supp_slamCSAIL}
	\end{center}
\end{figure}

\begin{figure}[h!]
	\begin{center}
		\begin{minipage}{\textwidth}
      \vspace{-6.8cm}
			\hspace{-0.5cm}
			\begin{tabular}{c}%
				\begin{minipage}{.5\columnwidth}%
					\centering%
					\includegraphics[width=1.0\columnwidth]{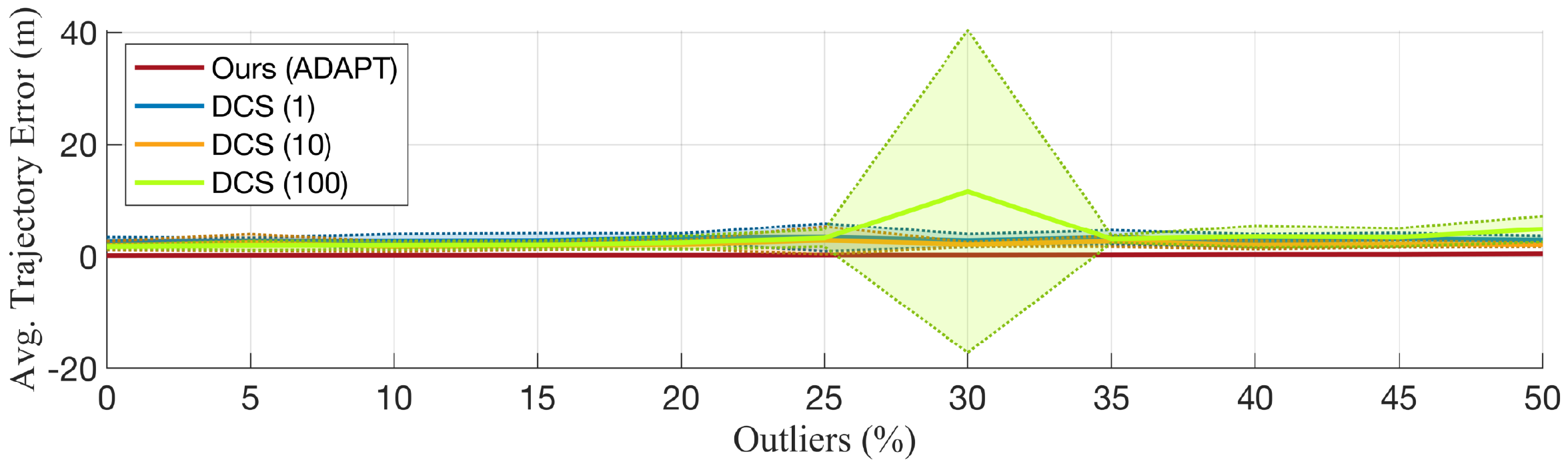}  \\
					(a) Average Trajectory Error
				\end{minipage}
				\\
				\begin{minipage}{.5\columnwidth}%
					\centering%
					\includegraphics[width=1.0\columnwidth]{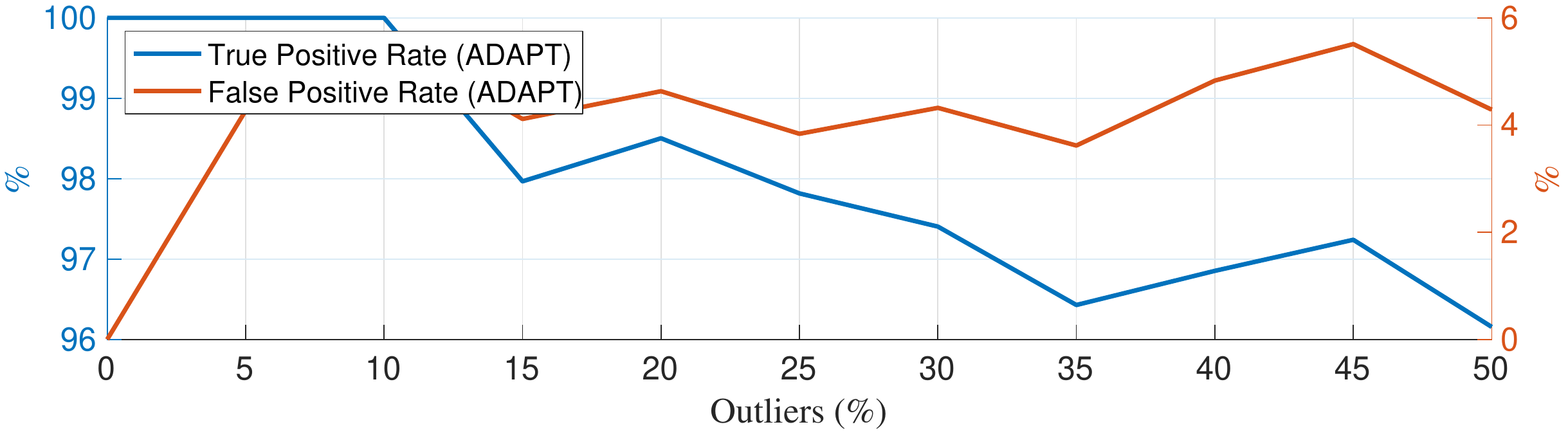}  \\
					(b) True/False Positive Rate
				\end{minipage}
				\\
				\begin{minipage}{.5\columnwidth}%
          \centering%
          \includegraphics[width=1.0\columnwidth]{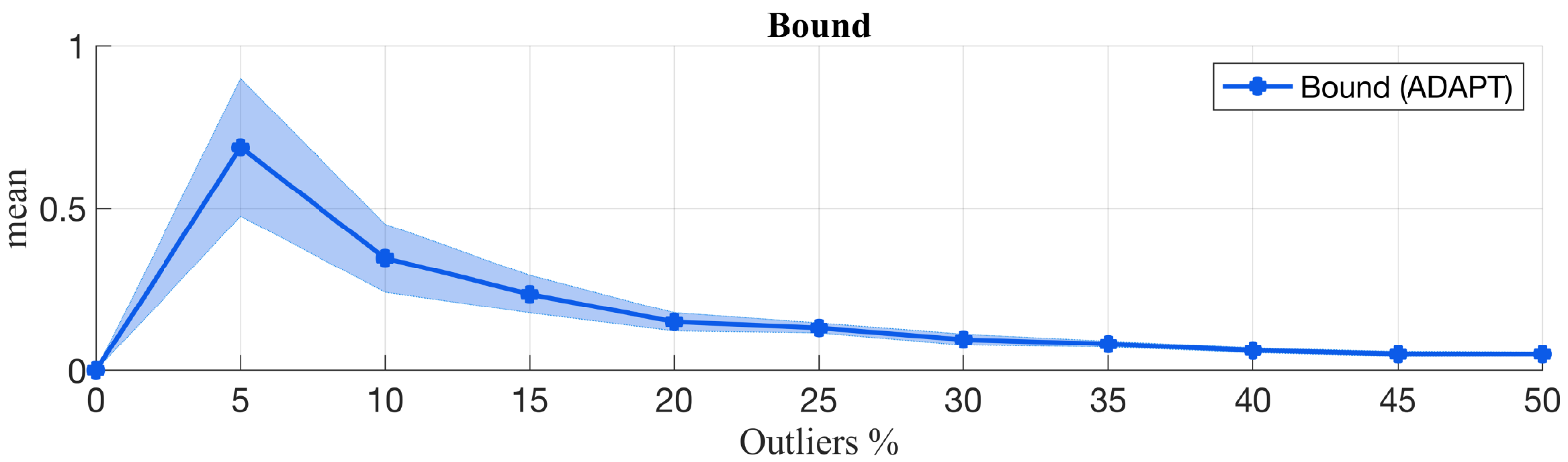} \\
          (c) Bound
        \end{minipage}
        \\
        \begin{minipage}{.5\columnwidth}%
          \centering%
          \includegraphics[width=1.0\columnwidth]{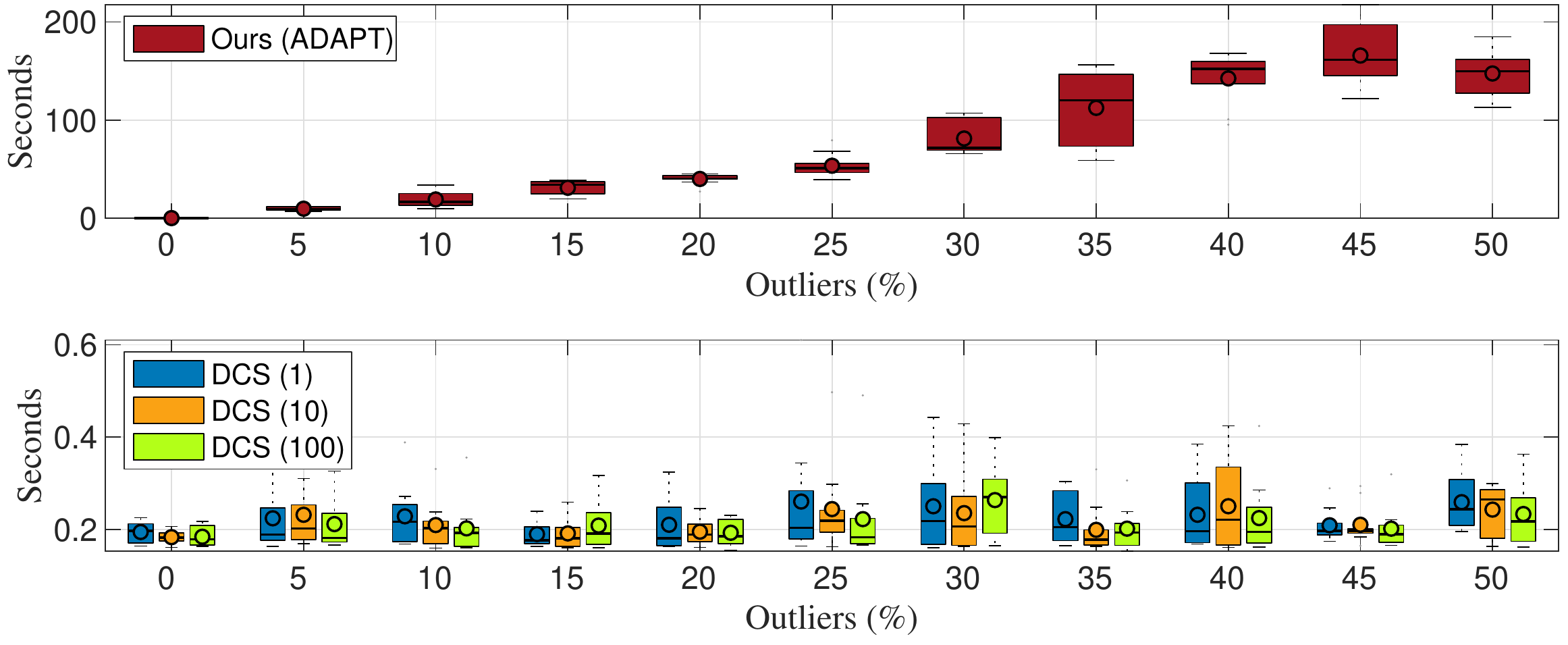} \\
          (d) Running time
        \end{minipage}
        \\
				\begin{minipage}{.5\columnwidth}%
          \centering%
          \includegraphics[width=1.0\columnwidth]{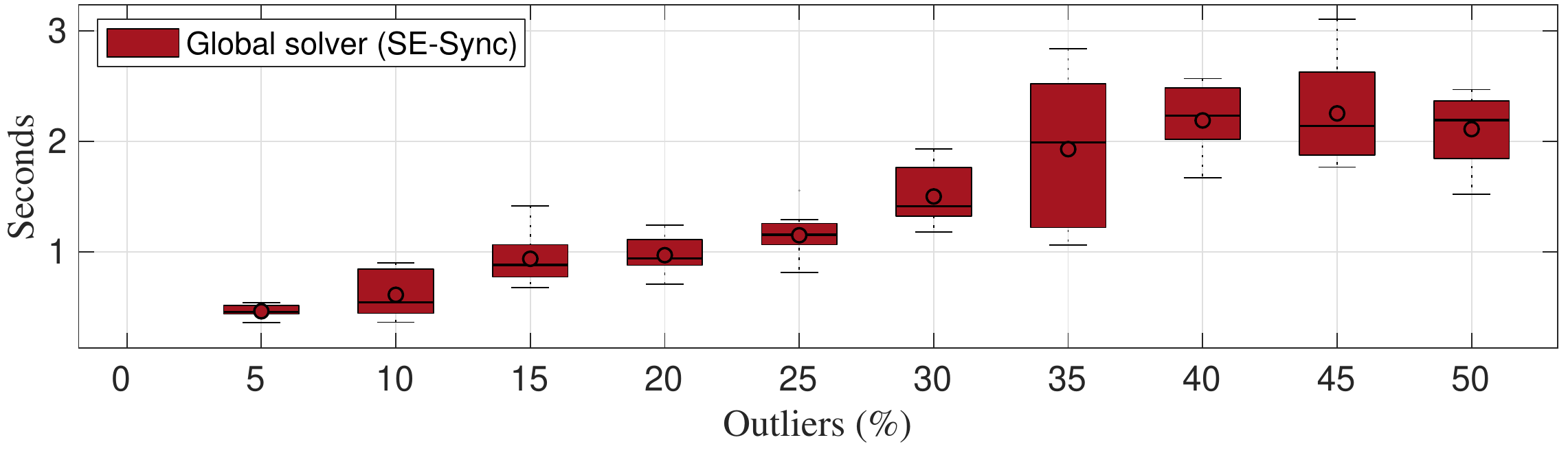} \\
          (e) Global Solver's Running Time
				\end{minipage}
			\end{tabular}
		\end{minipage}
		\caption{\name over 3D Grid dataset.}
		\label{fig:supp_slamGrid}
	\end{center}
\end{figure}

\newpage
\clearpage
\begin{figure}[h!]
	\begin{center}
		\begin{minipage}{\textwidth}
			\hspace{-0.5cm}
			\begin{tabular}{c}%
				\begin{minipage}{.5\columnwidth}%
					\centering%
					\includegraphics[width=1.0\columnwidth]{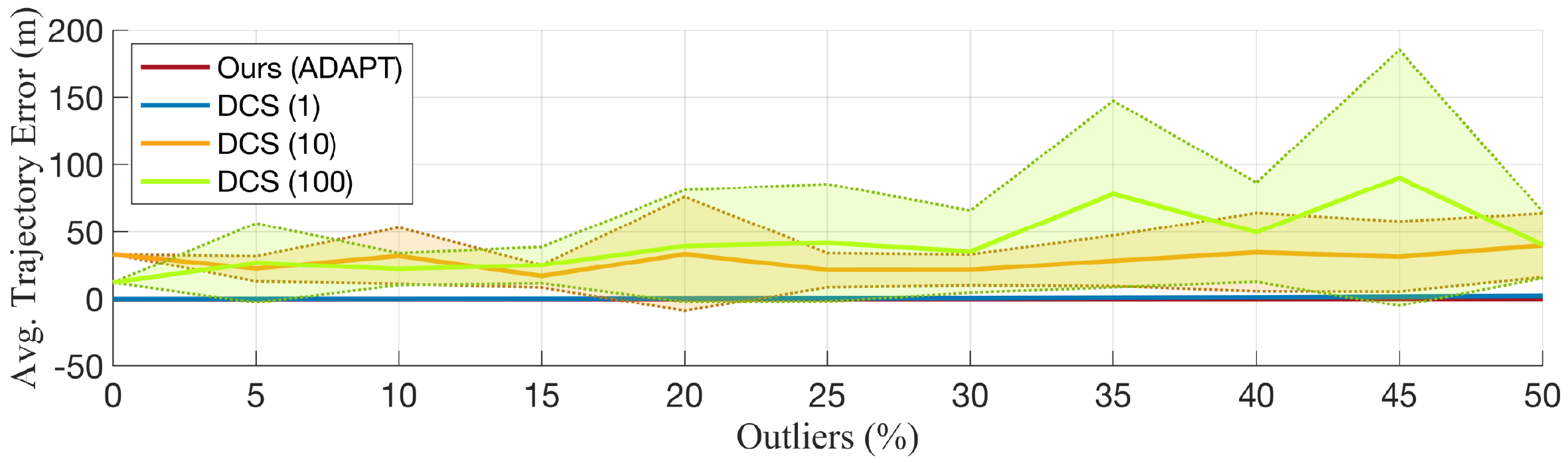}  \\
					(a) Average Trajectory Error
				\end{minipage}
				\\
				\begin{minipage}{.5\columnwidth}%
					\centering%
					\includegraphics[width=1.0\columnwidth]{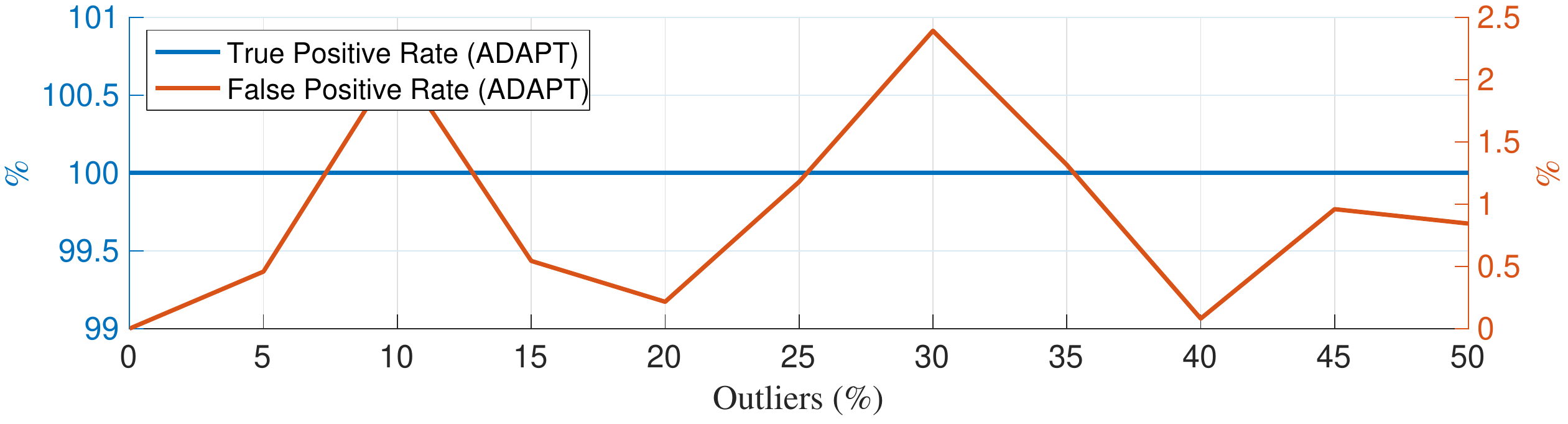}  \\
					(b) True/False Positive Rate
				\end{minipage}
				\\
				\begin{minipage}{.5\columnwidth}%
          \centering%
          \includegraphics[width=1.0\columnwidth]{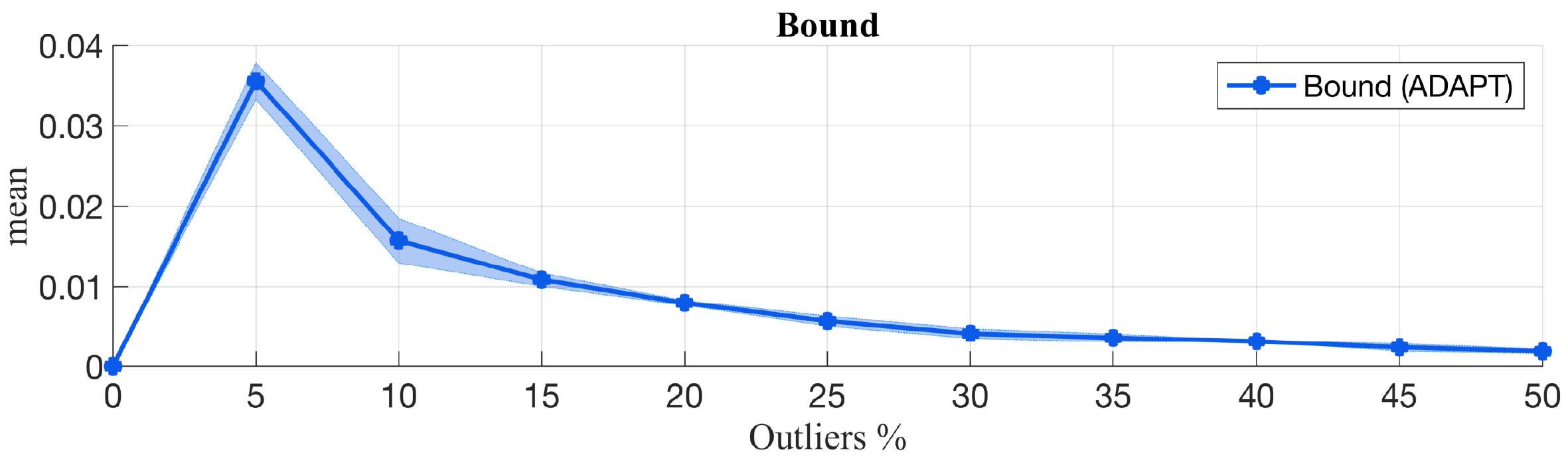} \\
          (c) Bound
        \end{minipage}
        \\
        \begin{minipage}{.5\columnwidth}%
          \centering%
          \includegraphics[width=1.0\columnwidth]{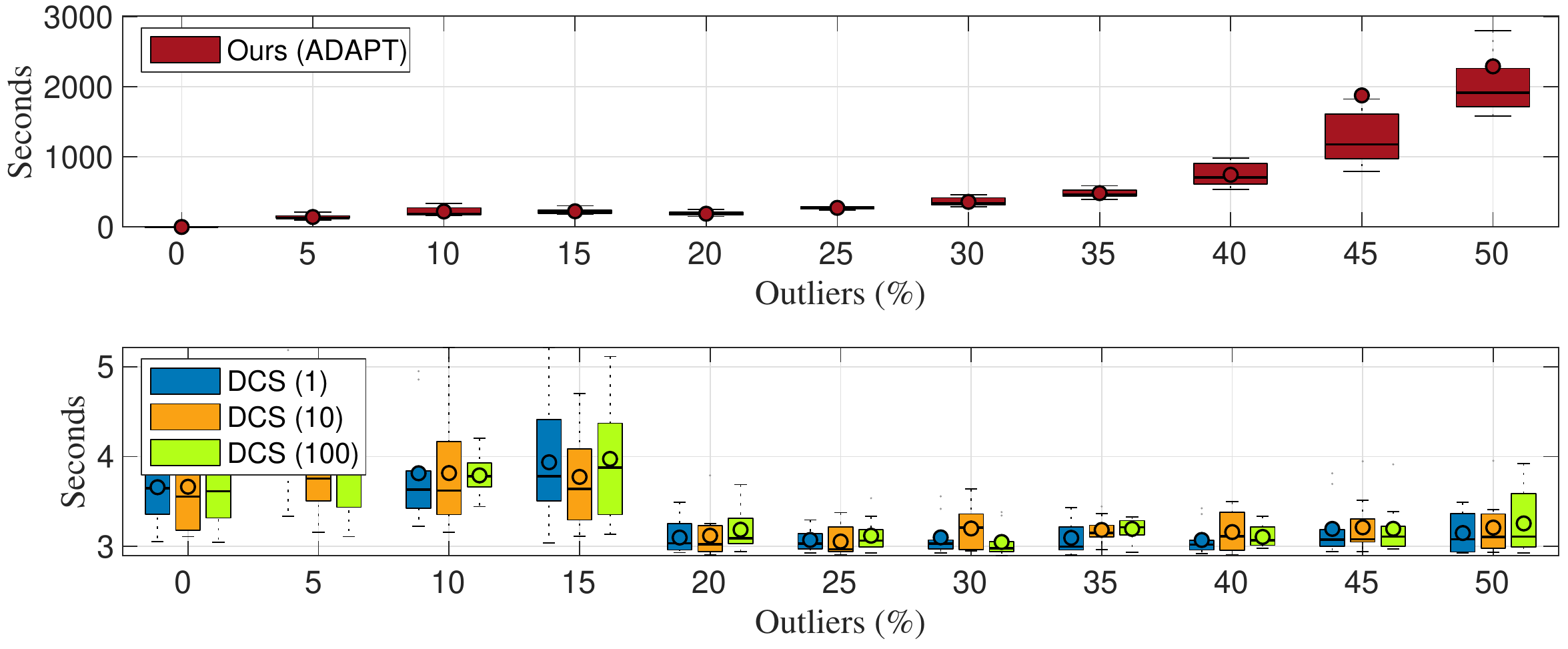} \\
          (d) Running time
        \end{minipage}
        \\
				\begin{minipage}{.5\columnwidth}%
          \centering%
          \includegraphics[width=1.0\columnwidth]{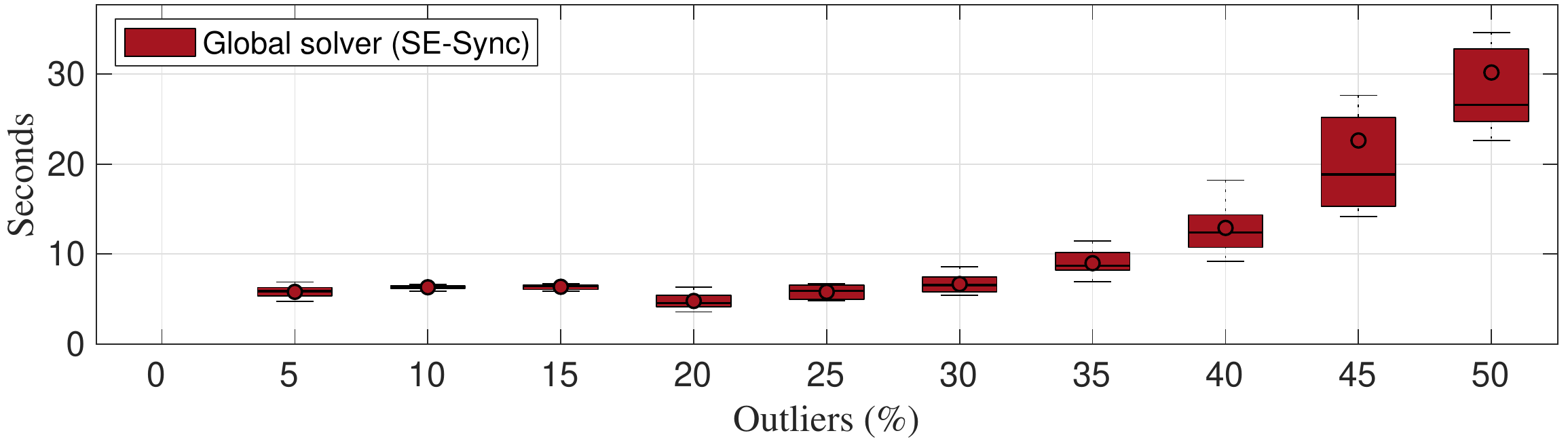} \\
          (e) Global Solver's Running Time
				\end{minipage}
			\end{tabular}
		\end{minipage}
		\caption{\name over Sphere 2500 dataset.}
		\label{fig:supp_slamSphere}
	\end{center}
\end{figure}

\bibliographystyle{IEEEtran}
\bibliography{refs.bib,myRefs.bib,refsRobustPerception.bib}

\begin{thebibliography}{10}
\providecommand{\url}[1]{#1}
\csname url@samestyle\endcsname
\providecommand{\newblock}{\relax}
\providecommand{\bibinfo}[2]{#2}
\providecommand{\BIBentrySTDinterwordspacing}{\spaceskip=0pt\relax}
\providecommand{\BIBentryALTinterwordstretchfactor}{4}
\providecommand{\BIBentryALTinterwordspacing}{\spaceskip=\fontdimen2\font plus
\BIBentryALTinterwordstretchfactor\fontdimen3\font minus
  \fontdimen4\font\relax}
\providecommand{\BIBforeignlanguage}[2]{{%
\expandafter\ifx\csname l@#1\endcsname\relax
\typeout{** WARNING: IEEEtran.bst: No hyphenation pattern has been}%
\typeout{** loaded for the language `#1'. Using the pattern for}%
\typeout{** the default language instead.}%
\else
\language=\csname l@#1\endcsname
\fi
#2}}
\providecommand{\BIBdecl}{\relax}
\BIBdecl

\bibitem{Scaramuzza11ram}
D.~Scaramuzza and F.~Fraundorfer, ``Visual odometry: Part {I} the first 30
  years and fundamentals,'' 2011.

\bibitem{Tam13tvcg-registrationSurvey}
G.~K.~L. Tam, Z.~Q. Cheng, Y.~K. Lai, F.~C. Langbein, Y.~Liu, D.~Marshall,
  R.~R. Martin, X.~F. Sun, and P.~L. Rosin, ``Registration of 3d point clouds
  and meshes: a survey from rigid to nonrigid.'' \emph{IEEE Trans. Vis. Comput.
  Graph.}, vol.~19, no.~7, pp. 1199--1217, 2013.

\bibitem{Roumeliotis02tra}
S.~Roumeliotis and G.~Bekey, ``Distributed multi-robot localization,''
  \emph{{IEEE} Trans. Robot. Automat.}, vol.~18, no.~5, pp. 781--795, October
  2002.

\bibitem{Whelan15rss}
T.~Whelan, S.~Leutenegger, R.~Salas-Moreno, B.~Glocker, and A.~Davison,
  ``{ElasticFusion}: Dense {SLAM} without a pose graph,'' in \emph{Robotics:
  Science and Systems (RSS)}, 2015.

\bibitem{Cadena16tro-SLAMsurvey}
C.~Cadena, L.~Carlone, H.~Carrillo, Y.~Latif, D.~Scaramuzza, J.~Neira, I.~Reid,
  and J.~J. Leonard, ``Past, present, and future of simultaneous localization
  and mapping: Toward the robust-perception age,'' \emph{{IEEE} Trans.
  Robotics}, vol.~32, no.~6, pp. 1309--1332, 2016.

\bibitem{Rosen18ijrr-sesync}
D.~Rosen, L.~Carlone, A.~Bandeira, and J.~Leonard, ``{SE-Sync}: a certifiably
  correct algorithm for synchronization over the {Special Euclidean} group,''
  \emph{Intl. J. of Robotics Research}, 2018.

\bibitem{Briales18cvpr-global2view}
J.~Briales, L.~Kneip, and J.~Gonzalez-Jimenez, ``A certifiably globally optimal
  solution to the non-minimal relative pose problem,'' in \emph{IEEE Conf. on
  Computer Vision and Pattern Recognition (CVPR)}, June 2018.

\bibitem{Horn87josa}
B.~K.~P. Horn, ``Closed-form solution of absolute orientation using unit
  quaternions,'' \emph{J. Opt. Soc. Amer.}, vol.~4, no.~4, pp. 629--642, Apr
  1987.

\bibitem{Bosse17fnt}
M.~Bosse, G.~Agamennoni, and I.~Gilitschenski, ``Robust estimation and
  applications in robotics,'' \emph{Foundations and Trends in Robotics},
  vol.~4, no.~4, pp. 225--269, 2016.

\bibitem{Speciale17cvpr-consensusMaximization}
P.~Speciale, D.~P. Paudel, M.~R. Oswald, T.~Kroeger, L.~V. Gool, and
  M.~Pollefeys, ``Consensus maximization with linear matrix inequality
  constraints,'' in \emph{IEEE Conf. on Computer Vision and Pattern Recognition
  (CVPR)}, July 2017, pp. 5048--5056.

\bibitem{Chen99pami-ransac}
C.~S. Chen, Y.~P. Hung, and J.~B. Cheng, ``{RANSAC}-based {DARCES}: A new
  approach to fast automatic registration of partially overlapping range
  images,'' \emph{{IEEE} Trans. Pattern Anal. Machine Intell.}, vol.~21,
  no.~11, pp. 1229--1234, 1999.

\bibitem{Latif12rss}
Y.~Latif, C.~D.~C. Lerma, and J.~Neira, ``Robust loop closing over time.'' in
  \emph{Robotics: Science and Systems (RSS)}, 2012.

\bibitem{Mangelson18icra}
J.~G. Mangelson, D.~Dominic, R.~M. Eustice, and R.~Vasudevan, ``Pairwise
  consistent measurement set maximization for robust multi-robot map merging,''
  in \emph{IEEE Intl. Conf. on Robotics and Automation (ICRA)}, 2018.

\bibitem{Bazin14eccv-robustRelRot}
J.~C. Bazin, Y.~Seo, R.~I. Hartley, and M.~Pollefeys, ``Globally optimal inlier
  set maximization with unknown rotation and focal length,'' in \emph{European
  Conf. on Computer Vision (ECCV)}, 2014, pp. 803--817.

\bibitem{hartley2009ijcv-global}
R.~I. Hartley and F.~Kahl, ``Global optimization through rotation space
  search,'' \emph{Intl. J. of Computer Vision}, vol.~82, no.~1, pp. 64--79,
  2009.

\bibitem{Zheng11cvpr-robustFitting}
Y.~Zheng, S.~Sugimoto, and M.~Okutomi, ``Deterministically maximizing feasible
  subsystem for robust model fitting with unit norm constraint,'' in \emph{IEEE
  Conf. on Computer Vision and Pattern Recognition (CVPR)}, 2011, pp.
  1825--1832.

\bibitem{Li09cvpr-robustFitting}
H.~Li, ``Consensus set maximization with guaranteed global optimality for
  robust geometry estimation,'' in \emph{Intl. Conf. on Computer Vision
  (ICCV)}, 2009, pp. 1074--1080.

\bibitem{Bustos18pami-GORE}
{\'A}.~P. Bustos and T.~J. Chin, ``Guaranteed outlier removal for point cloud
  registration with correspondences,'' \emph{{IEEE} Trans. Pattern Anal.
  Machine Intell.}, vol.~40, no.~12, pp. 2868--2882, 2018.

\bibitem{Chin18eccv-robustFitting}
T.-J. Chin, Z.~Cai, and F.~Neumann, ``Robust fitting in computer vision: Easy
  or hard?'' in \emph{European Conf. on Computer Vision (ECCV)}, 2018.

\bibitem{Rousseeuw87book}
P.~J. Rousseeuw and A.~M. Leroy, \emph{Robust Regression and Outlier
  Detection}.\hskip 1em plus 0.5em minus 0.4em\relax John Wiley \& Sons, New
  York, NY, 1987.

\bibitem{Huber64ams}
P.~J. Huber, ``Robust estimation of a location parameter,'' \emph{The Annals of
  Mathematical Statistics}, vol.~35, no.~1, pp. 73--101, 1964.

\bibitem{Dellaert17fnt-factorGraph}
F.~Dellaert and M.~Kaess, ``Factor graphs for robot perception,''
  \emph{Foundations and Trends in Robotics}, vol.~6, no. 1-2, pp. 1--139, 2017.

\bibitem{Rousseeuw11dmkd}
P.~J. Rousseeuw and M.~Hubert, ``Robust statistics for outlier detection,''
  \emph{Wiley Interdisciplinary Reviews: Data Mining and Knowledge Discovery},
  vol.~1, no.~1, pp. 73--79, 2011.

\bibitem{Yang16pami-goicp}
J.~Yang, H.~Li, D.~Campbell, and Y.~Jia, ``{Go-ICP}: A globally optimal
  solution to {3D ICP} point-set registration,'' \emph{{IEEE} Trans. Pattern
  Anal. Machine Intell.}, vol.~38, no.~11, pp. 2241--2254, Nov. 2016.

\bibitem{Zhou16eccv-fastGlobalRegistration}
Q.~Y. Zhou, J.~Park, and V.~Koltun, ``Fast global registration,'' in
  \emph{European Conf. on Computer Vision (ECCV)}.\hskip 1em plus 0.5em minus
  0.4em\relax Springer, 2016, pp. 766--782.

\bibitem{Nister04pami}
D.~Nist\'er, ``An efficient solution to the five-point relative pose problem,''
  \emph{{IEEE} Trans. Pattern Anal. Machine Intell.}, vol.~26, no.~6, pp.
  756--770, 2004.

\bibitem{Hartley04book}
R.~I. Hartley and A.~Zisserman, \emph{Multiple View Geometry in Computer
  Vision}, 2nd~ed.\hskip 1em plus 0.5em minus 0.4em\relax Cambridge University
  Press, 2004.

\bibitem{Arora09book-complexity}
S.~Arora and B.~Barak, \emph{Computational complexity: A modern
  approach}.\hskip 1em plus 0.5em minus 0.4em\relax Cambridge University Press,
  2009.

\bibitem{nemhauser78analysis}
G.~Nemhauser, L.~Wolsey, and M.~Fisher, ``An analysis of approximations for
  maximizing submodular set functions -- {I},'' \emph{Mathematical
  Programming}, vol.~14, no.~1, pp. 265--294, 1978.

\bibitem{CPLEXwebsite}
\BIBentryALTinterwordspacing
IBM, ``{CPLEX: IBM ILOG CPLEX Optimization Studio}.'' [Online]. Available:
  \url{https://www.ibm.com/products/ilog-cplex-optimization-studio}
\BIBentrySTDinterwordspacing

\bibitem{pomerleau2012ethpc}
F.~Pomerleau, M.~Liu, F.~Colas, and R.~Siegwart, ``{Challenging data sets for
  point cloud registration algorithms},'' \emph{The International Journal of
  Robotics Research}, vol.~31, no.~14, pp. 1705--1711, Dec. 2012.

\bibitem{burri2016euroc}
\BIBentryALTinterwordspacing
M.~Burri, J.~Nikolic, P.~Gohl, T.~Schneider, J.~Rehder, S.~Omari, M.~W.
  Achtelik, and R.~Siegwart, ``The euroc micro aerial vehicle datasets,''
  \emph{The International Journal of Robotics Research}, 2016. [Online].
  Available:
  \url{http://ijr.sagepub.com/content/early/2016/01/21/0278364915620033.abstract}
\BIBentrySTDinterwordspacing

\bibitem{Carlone18tro-attentionVIN}
L.~Carlone and S.~Karaman, ``Attention and anticipation in fast visual-inertial
  navigation,'' \emph{{IEEE} Trans. Robotics}, 2018.

\bibitem{hartley1997eightpt}
R.~I. {Hartley}, ``In defense of the eight-point algorithm,'' \emph{IEEE
  Transactions on Pattern Analysis and Machine Intelligence}, vol.~19, no.~6,
  pp. 580--593, June 1997.

\bibitem{Carlone14tro-SO2}
L.~Carlone and A.~Censi, ``From angular manifolds to the integer lattice:
  Guaranteed orientation estimation with application to pose graph
  optimization,'' \emph{{IEEE} Trans. Robotics}, vol.~30, no.~2, pp. 475--492,
  2014.

\bibitem{Agarwal13icra}
P.~Agarwal, G.~D. Tipaldi, L.~Spinello, C.~Stachniss, and W.~Burgard, ``Robust
  map optimization using dynamic covariance scaling,'' in \emph{IEEE Intl.
  Conf. on Robotics and Automation (ICRA)}, 2013.

\bibitem{Meer91ijcv-robustVision}
P.~Meer, D.~Mintz, A.~Rosenfeld, and D.~Y. Kim, ``Robust regression methods for
  computer vision: A review,'' \emph{Intl. J. of Computer Vision}, vol.~6,
  no.~1, pp. 59--70, Apr 1991.

\bibitem{Stewart99siam-robustVision}
\BIBentryALTinterwordspacing
C.~Stewart, ``Robust parameter estimation in computer vision,'' \emph{SIAM
  Review}, vol.~41, no.~3, pp. 513--537, 1999. [Online]. Available:
  \url{https://doi.org/10.1137/S0036144598345802}
\BIBentrySTDinterwordspacing

\bibitem{Chin16cvpr-outlierRejection}
T.~Chin, Y.~H. Kee, A.~Eriksson, and F.~Neumann, ``Guaranteed outlier removal
  with mixed integer linear programs,'' in \emph{IEEE Conf. on Computer Vision
  and Pattern Recognition (CVPR)}, June 2016, pp. 5858--5866.

\bibitem{Sunderhauf12iros}
N.~S\"{u}nderhauf and P.~Protzel, ``Switchable constraints for robust pose
  graph {SLAM},'' in \emph{IEEE/RSJ Intl. Conf. on Intelligent Robots and
  Systems (IROS)}, 2012.

\bibitem{Olson12rss}
E.~Olson and P.~Agarwal, ``Inference on networks of mixtures for robust robot
  mapping,'' in \emph{Robotics: Science and Systems (RSS)}, July 2012.

\bibitem{Lajoie19ral-DCGM}
P.~Lajoie, S.~Hu, G.~Beltrame, and L.~Carlone, ``Modeling perceptual aliasing
  in {SLAM} via discrete-continuous graphical models,'' \emph{{IEEE} Robotics
  and Automation Letters ({RA-L})}, 2019.

\bibitem{black1996ijcv-unification}
M.~J. Black and A.~Rangarajan, ``On the unification of line processes, outlier
  rejection, and robust statistics with applications in early vision,''
  \emph{Intl. J. of Computer Vision}, vol.~19, no.~1, pp. 57--91, 1996.

\bibitem{Fischler81}
M.~Fischler and R.~Bolles, ``Random sample consensus: a paradigm for model
  fitting with application to image analysis and automated cartography,''
  \emph{Commun. ACM}, vol.~24, pp. 381--395, 1981.

\bibitem{Arun87pami}
K.~S. Arun, T.~S. Huang, and S.~D. Blostein, ``Least-squares fitting of two
  3-{D} point sets,'' \emph{{IEEE} Trans. Pattern Anal. Machine Intell.},
  vol.~9, no.~5, pp. 698 --700, sept. 1987.

\bibitem{MacTavish15crv-robustEstimation}
K.~M. Tavish and T.~D. Barfoot, ``At all costs: A comparison of robust cost
  functions for camera correspondence outliers,'' in \emph{Computer and Robot
  Vision (CRV), 2015 12th Conference on}.\hskip 1em plus 0.5em minus
  0.4em\relax IEEE, 2015, pp. 62--69.

\bibitem{Campbell17cvpr-globallyOptRobustTwoView}
D.~Campbell, L.~Petersson, L.~Kneip, and H.~Li, ``Globally-optimal inlier set
  maximisation for simultaneous camera pose and feature correspondence,'' in
  \emph{IEEE Conf. on Computer Vision and Pattern Recognition (CVPR)}, 2017,
  pp. 1--10.

\bibitem{Besl92pami}
P.~J. Besl and N.~D. McKay, ``A method for registration of {3-D} shapes,''
  \emph{{IEEE} Trans. Pattern Anal. Machine Intell.}, vol.~14, no.~2, 1992.

\bibitem{Segal06rss-generalizedICP}
A.~Segal, D.~H\"{a}hnel, and S.~Thrun, ``Generalized-{ICP},'' 06 2009.

\bibitem{Chetverikov05ivc-tICP}
\BIBentryALTinterwordspacing
D.~Chetverikov, D.~Stepanov, and P.~Krsek, ``Robust euclidean alignment of {3D}
  point sets: the trimmed iterative closest point algorithm,'' \emph{Image and
  Vision Computing}, vol.~23, no.~3, pp. 299 -- 309, 2005. [Online]. Available:
  \url{http://www.sciencedirect.com/science/article/pii/S0262885604001179}
\BIBentrySTDinterwordspacing

\bibitem{Nister03cvpr}
D.~Nist\'er, ``An efficient solution to the five-point relative pose problem,''
  in \emph{IEEE Conf. on Computer Vision and Pattern Recognition (CVPR)}, 2003.

\bibitem{Naroditsky12pami-3point}
O.~Naroditsky, X.~S. Zhou, J.~Gallier, S.~I. Roumeliotis, and K.~Daniilidis,
  ``Two efficient solutions for visual odometry using directional
  correspondence,'' \emph{{IEEE} Trans. Pattern Anal. Machine Intell.},
  vol.~34, no.~4, pp. 818--824, April 2012.

\bibitem{Kneip11bmvc}
L.~Kneip, M.~Chli, and R.~Siegwart, ``Robust real-time visual odometry with a
  single camera and an {IMU},'' in \emph{British Machine Vision Conf. (BMVC)},
  2011, pp. 16.1--16.11.

\bibitem{Scaramuzza11ijcv}
D.~Scaramuzza, ``1-point-ransac structure from motion for vehicle-mounted
  cameras by exploiting non-holonomic constraints,'' \emph{Intl. J. of Computer
  Vision}, pp. 1--12, 2011.

\bibitem{Sunderhauf12icra}
N.~Sunderhauf and P.~Protzel, ``Towards a robust back-end for pose graph
  {SLAM},'' in \emph{IEEE Intl. Conf. on Robotics and Automation (ICRA)}, 2012,
  pp. 1254--1261.

\bibitem{Tong11icra-robustSlam}
C.~H. Tong and T.~D. Barfoot, ``Batch heterogeneous outlier rejection for
  feature-poor slam,'' in \emph{IEEE Intl. Conf. on Robotics and Automation
  (ICRA)}, 2011, pp. 2630--2637.

\bibitem{Yong13astro-robustSLAM}
------, ``Evaluation of heterogeneous measurement outlier rejection schemes for
  robotic planetary surface mapping,'' \emph{Acta Astronautica}, vol.~88, pp.
  146--162, 2013.

\bibitem{Lee13iros}
G.~H. Lee, F.~Fraundorfer, and M.~Pollefeys, ``Robust pose-graph loop-closures
  with expectation-maximization,'' in \emph{IEEE/RSJ Intl. Conf. on Intelligent
  Robots and Systems (IROS)}, 2013.

\bibitem{Wang13ima}
L.~Wang and A.~Singer, ``Exact and stable recovery of rotations for robust
  synchronization,'' \emph{Information and Inference: A Journal of the IMA},
  vol.~30, 2013.

\bibitem{Carlone18ral-robustPGO2D}
L.~Carlone and G.~Calafiore, ``Convex relaxations for pose graph optimization
  with outliers,'' \emph{{IEEE} Robotics and Automation Letters ({RA-L})},
  vol.~3, no.~2, pp. 1160--1167, 2018.

\bibitem{Arrigoni18cviu}
F.~Arrigoni, B.~Rossi, P.~Fragneto, and A.~Fusiello, ``Robust synchronization
  in {SO(3)} and {SE(3)} via low-rank and sparse matrix decomposition,''
  \emph{Comput. Vis. Image Underst.}, 2018.

\bibitem{Kalman60}
R.~E. Kalman, ``A new approach to linear filtering and prediction problems,''
  \emph{Trans. ASME, Journal of Basic Engineering}, vol.~82, pp. 35--45, 1960.

\bibitem{Diakonikolas201focs-robustEstimation}
I.~Diakonikolas, G.~Kamath, D.~M. Kane, J.~Li, A.~Moitra, and A.~Stewart,
  ``Robust estimators in high dimensions without the computational
  intractability,'' in \emph{IEEE 57th Annual Symposium on Foundations of
  Computer Science}.\hskip 1em plus 0.5em minus 0.4em\relax IEEE, 2016, pp.
  655--664.

\bibitem{Candes05tit}
E.~J. Candes and T.~Tao, ``Decoding by linear programming,'' \emph{IEEE
  Transactions on Information Theory}, vol.~51, no.~12, pp. 4203--4215, 2005.

\bibitem{Pasqualetti13tac-attackDetection}
F.~Pasqualetti, F.~D{\"o}rfler, and F.~Bullo, ``Attack detection and
  identification in cyber-physical systems,'' \emph{IEEE Transactions on
  Automatic Control}, vol.~58, no.~11, pp. 2715--2729, 2013.

\bibitem{Liu19arxiv-TrimmedHardThresholding}
L.~Liu, T.~Li, and C.~Caramanis, ``High dimensional robust estimation of sparse
  models via trimmed hard thresholding,'' \emph{arXiv preprint: 1901.08237},
  2019.

\bibitem{Zhang11tit-ForwardBackGreedy}
T.~Zhang, ``Adaptive forward-backward greedy algorithm for learning sparse
  representations,'' \emph{IEEE Transactions on Information Theory}, vol.~57,
  no.~7, pp. 4689--4708, 2011.

\bibitem{Liu2018tsp-GreedyRobustRegression}
J.~Liu, P.~C. Cosman, and B.~D. Rao, ``Robust linear regression via $\ell\_0$
  regularization,'' \emph{IEEE Transactions on Signal Processing}, vol.~66,
  no.~3, pp. 698--713.

\bibitem{Mishra2017tac-secureStateEstimation}
S.~Mishra, Y.~Shoukry, N.~Karamchandani, S.~N. Diggavi, and P.~Tabuada,
  ``Secure state estimation against sensor attacks in the presence of noise,''
  \emph{IEEE Transactions on Control of Network Systems}, vol.~4, no.~1, pp.
  49--59, 2017.

\bibitem{Aghapour2018cdc-outlierAccomodation}
E.~Aghapour, F.~Rahman, and J.~A. Farrell, ``Outlier accommodation by
  risk-averse performance-specified linear state estimation,'' in \emph{2018
  IEEE Conference on Decision and Control}.\hskip 1em plus 0.5em minus
  0.4em\relax IEEE, 2018, pp. 2310--2315.

\bibitem{Luo19arxiv-SecureStateEst}
X.~Luo, M.~Pajic, and M.~M. Zavlanos, ``A scalable and optimal graph-search
  method for secure state estimation,'' \emph{arXiv preprint arXiv:1903.10620},
  2019.

\bibitem{Foster15colt-variableSelectionHard}
D.~Foster, H.~Karloff, and J.~Thaler, ``Variable selection is hard,'' in
  \emph{Conference on Learning Theory}, 2015, pp. 696--709.

\end{thebibliography}

\end{document}